\documentclass[12pt]{article}

\usepackage{amsmath}
\usepackage{amssymb}
\usepackage{natbib}
\usepackage{graphicx}
\usepackage{url}
\usepackage{algorithm2e}

\usepackage[margin=1in]{geometry}
\usepackage{authblk}

\usepackage[colorlinks=true, allcolors=blue]{hyperref}
\usepackage{amsfonts}
\usepackage{xcolor}
\usepackage{times}
\newcommand{\eps}{\varepsilon}
\newcommand{\acts}{\mathcal{A}}
\newcommand{\obs}{\mathcal{O}}
\newcommand{\hypo}{\mathcal{H}}
\newcommand{\expec}{\mathbb{E}}
\newcommand{\estim}{\widehat{M}}
\newcommand{\guess}{\widehat{M}_{t}}
\newcommand{\guesspi}{M_{t,\pi}}
\newcommand{\poli}{\tilde{\pi}}
\newcommand{\hells}{D^{2}_{H}}
\newcommand{\imdp}{\mathbb{M}}
\newcommand{\gen}{\overline{M}}
\newcommand{\states}{\mathcal{S}}

\newcommand{\tsum}{\sum_{t=1}^{T}}
\newcommand{\bonval}{\overline{V}}
\newcommand{\hellsc}{D^{2}_{H_{\{0,1\}}}}
\newcommand{\Set}[2]{\left\{#1 \ \middle|\ #2\right\}}
\newcommand{\bel}[2]{\underset{#2}{#1}}
\newtheorem{defi}{Definition}
\newtheorem{prop}{Proposition}
\newtheorem{coro}{Corollary}
\newtheorem{theo}{Theorem}
\newtheorem{lemm}{Lemma}
\newtheorem{exam}{Example}

\title{Regret Bounds for Robust Online Decision Making}

\author[1]{Alexander Appel}
\affil[1]{Computational Rational Agents Laboratory}
\author[2]{Vanessa Kosoy}
\affil[2]{Faculty of Mathematics, Technion - Israel Institute of Technology}

\begin{document}

\maketitle

\begin{abstract}%
  We propose a framework which generalizes "decision making with structured observations" from \cite{FosterGH23} by allowing \emph{robust} (i.e. multivalued) models. In this framework, each model associates each decision with a \emph{convex set} of probability distributions over outcomes. Nature can choose distributions out of this set in an arbitrary (adversarial) manner, that can be non-oblivious and depend on past history. The resulting framework offers much greater generality than classical bandits and reinforcement learning, since the realizability assumption becomes much weaker and more realistic. We then derive a theory of regret bounds for this framework, which extends the "decision-estimation coefficients" of \cite{FosterGH23}. Although our lower and upper bounds are not tight, they are sufficient to fully characterize power-law learnability. We demonstrate this theory in two special cases: robust linear bandits (previously studied in \cite{Kosoy2024}) and tabular robust online reinforcement learning (previously studied in \cite{Tian0YS21}). In both cases, we derive regret bounds that improve state-of-the-art (except that we do not address computational efficiency).
\end{abstract}

\section{Introduction}

Traditional approaches to formal guarantees and statistical complexity analysis for interactive decision-making (e.g. multi-armed bandits, reinforcement learning theory) often need strong assumptions about the environment. Typical of assumptions include

\begin{itemize}
    \item Assuming that the outcomes of any particular decision are IID samples from a fixed distribution (stochastic multi-armed bandits, see e.g. \cite{Lattimore2020}).
    \item Assuming that the state of environment changes according to a constant Markovian law (see e.g. \cite{AzarOM17}).
    \item Assuming that the distribution of outcomes belongs to a known parametric family of distributions (see e.g. \cite{Russo13,Osband14,Jin20}).
    \item Assuming that the MDP value function or some related quantity belongs to some known parametric family of functions (model-free reinforcement learning, see e.g. \cite{Wang20,Du21,Jin21}).
\end{itemize}

Clearly, such realizability assumptions are often unrealistic, especially for agents operating in the physical world (as opposed to a game or a simulation). Any exact Markovian description of the real world would have an astronomically large state space, and any family of distributions that includes an exact description of physical law would be intractably large, at least because the computational complexity of the physical world is far beyond that of the agent itself.

There are several approaches for relaxing these assumptions, but each has major limitations.

The first approach is allowing a linear term in the regret bound, whose coefficient scales with some measure of distance between the real environment and the algorithm's class of models (see e.g. \cite{Zanette20}). However, most work in this approach is limited to the model-free setting. Moreover, this typically requires the model to well-approximate the environment \emph{uniformly} over the state space, whereas realistic approximations often break down outside of particular regions.

The second approach is allowing the environment to adversarially change its behavior between episodes/trials: adversarial multi-armed bandits (see e.g. \cite{Lattimore2020}) and adversarial reinforcement learning (see e.g. \cite{Liu22,Foster22}). However, the latter is plagued by statistical intractability results. Even in the limited special cases when adversarial learning is tractable, it still requires the environment to have a simple unambiguous description within each trial: e.g. in linear adversarial bandits the reward has to be a linear function of the arm.

The third approach, which is the basis for our own, is using \emph{robust} Markov Decision Processes (RMDP) (see e.g. \cite{Suilen25}). As opposed to an ordinary MDP, an RMDP has a \emph{multivalued} transition kernel, and nature can adversarially choose the distribution upon each state transition. However, most work on reinforcement learning with RMDP's (see e.g. \cite{Wang21,Dong22,Panaganti22}) has two assumptions which we dispense with:

\begin{itemize}
    \item The set of distributions assigned by the transition kernel to a state-action pair has to be a ball relative to some metric or divergence.
    \item The training date for the algorithm consists of interaction with the MDP defined by taking the \emph{centers} of these balls.
\end{itemize}

The typical motivation is training a reinforcement learning agent in a simulation (corresponding to the center MDP) and deploying it in a physical environment (corresponding to the RMDP). This requires that a sufficiently accurate simulation is available, and that the realizability assumption is satisfied for the simulation. On the other hand, we want to study robust \emph{online} reinforcement learning, where the agent has to learn directly from the real world (or the simulation is so complex that the realizability assumption fails even there). One work which \emph{does} consider learning from interacting with the actual RMDP is \cite{Lim13}. Their setting is a very narrow special case of our framework, where the multivalued kernel is always either a singleton or equal to a \emph{known} set.

We propose to study an agent that makes a sequence of decisions. On each round, the agent chooses an action $a$ from a set $\acts$, receives an observation $o$ from a set $\obs$, and calculates its reward from $a,o$. A class $\hypo$ of models is given, where a model is a \emph{multivalued} function from $\acts$ to distributions over $\obs$. $\hypo$ acts as a set of possible constraints which nature may fulfill. In rich environments, models of this form may be much easier to specify and learn than a full model of the environment. Given a model $M$, the set $M(a)$ may be interpreted as a menu of choices available to an adversarial environment\footnote{This can be regarded as decision-making with uncertainty expressed as \emph{imprecise probability}, see e.g. \cite{Augustin2014}.}. Each model $M$ has an associated maximin value, and we define the \emph{regret} relatively to that value\footnote{The typical choice in adversarial reinforcement learning is to measure the regret against the best action in hindsight. However, Theorem 1 of \cite{Tian0YS21} gives a family of robust MDP's with a lower bound on best-action regret which is exponential in the horizon. Maximin regret avoids this impossibility result.} assuming that nature is constrained to be consistent with $M$\footnote{Importantly, we don't require nature to choose a \emph{fixed} mapping from $\acts$ to distributions on $\obs$. Nature's policy can be non-stationary and non-oblivious.}.

Notice that even though this setting has the appearance of a multi-armed bandit, it allows episodic reinforcement learning as a special case: in this case, $\acts$ consists of within-episode policies and $\obs$ is the set of possible trajectories within an episode.

Here are some examples of possible applications.

\begin{exam}
A robot learns to complete tasks (e.g. moving objects, cleaning, cooking) in an environment that has inanimate objects and people. The robot's own movement and the interaction with objects can be described fairly well by fixed probability distributions. However, the behavior of people can only be described as a set of probability distributions. Moreover, their behavior might be affected by previous interactions with the robot in hard-to-predict ways.
\end{exam}

\begin{exam}
A self-driving car needs to bring passengers to their destination, while taking into account speed, convenience, fuel consumption and safety. The movement of cars can be partially predicted based on physical laws and common driver habits, but partially it depends on hard-to-predict idiosyncrasies of individual drivers. Moreover, the car's sensors only give it information about a certain neighborhood around the car. Previously unknown objects can enter this area, and the probability distribution of such new objects can change between days or geographical areas in complex ways.
\end{exam}

\begin{exam}
An AI system manages an investment portfolio on the stock market. Not only that some price movements are hard to predict (e.g. caused by a technological breakthrough or a political event), but some are the result of adversarial actors trying to profit on expense of the AI's portfolio that learning from its previous behavior.
\end{exam}

In \cite{FosterGH23}, regret bounds for online decision-making were studied for conventional (single-valued) models. They proved that a certain function called the \emph{decision-estimation coefficient} enabled an almost tight characterization of the regret bound for any model class. Here, we define an analogue of the decision-estimation coefficient for the robust setting, and show a similar upper bound on regret. We also show a lower bound on regret, which is weaker than the non-robust analogue, but still sufficient to characterize model classes which admit a sublinear power-law regret bound.

In \cite{FosterGH23}, the algorithm that implements that upper bound requires access to an online distribution learning oracle, and the bound scales with the statistical complexity of the latter. This statistical complexity has bounds in terms of model class cardinality or covering number, which are derived in \cite{Foster21} using the techniques of \cite{Vovk95}. In our setting, we require a similar oracle for \emph{robust} online distribution learning. Again, we derive similar bounds on statistical complexity, but this requires different techniques: our algorithm imitates a \emph{prediction market}, an idea inspired by \cite{Shafer05,Garrabrant17,Kosoy17}.

We apply our methods to derive upper bounds on regret in two special cases. One case is tabular episodic robust online reinforcement learning, which was studied in \cite{Tian0YS21}. The motivation of \cite{Tian0YS21} was studying multi-agent learning, but since an RMDP is equivalent to a zero-sum two-player stochastic game\footnote{One player is choosing the RMDP action, the other player is choosing the distribution out of the multivalued transition kernel.}, their framework is a special case of our own. There, we get a regret bound of $\tilde{\mathcal{O}}(H \sqrt{S^3 A T})$, where $H$ is the episode length, $S$ is the number of states, $A$ is the number of actions, and $T$ is the number of episodes. For $T\gg 0$, this is an improvement on \cite{Tian0YS21}'s bound of $\tilde{\mathcal{O}}(H^2 \sqrt[3]{S A T^2})$. However, our algorithm is not computationally efficient. It remains an open question whether there exists a polynomial-time algorithm with regret of $\tilde{\mathcal{O}}(\mathrm{poly}(H,S,A)\sqrt{T})$\footnote{In \cite{Xie20}, a polynomial-time algorithm for learning a zero-sum game is presented, which has regret $\tilde{\mathcal{O}}(\sqrt{H^3 S^3 A_1^3 A_2^3 T})$, where $A_1$ and $A_2$ are the cardinalities of the action sets of the two players. However, they assume that the player can observe the opponent's action, which with our motivation is an unnatural assumption.}.

The other special case is robust linear bandits, previously studied in \cite{Kosoy2024}. There, we get a similar $\tilde{\mathcal{O}}(\sqrt{T})$ regret bound, except that the dependence on the dimension $Z$ of the model class is improved from $Z^2$ to $Z$.
\subsection{Notation and General Framework}
Given a space $X$, $\Delta X$ is the space of probability distributions $\mu,\nu$ on $X$. An \emph{imprecise belief} on $X$ is a nonempty closed convex subset of $\Delta X$. $\Box X$ is the space of all imprecise beliefs $\Psi,\Phi$ on $X$. Expectations of functions with respect to imprecise beliefs are defined to be the worst-case expectation.
$$\expec_{\Psi}[f]:=\bel{\min}{\mu\in\Psi}\ \expec_{\mu}[f]$$
Our general framework for robust reinforcement learning is as follows. There are spaces $\acts,\obs$ of actions $a$ and observations $o$, and a known reward function $r:\acts\times\obs\to[0,1]$. $T$ is the number of timesteps. A stochastic algorithm $\pi:(\acts\times\obs)^{<T}\to\Delta\acts$ repeatedly interacts with an environment $\theta:(\acts\times\obs)^{<T}\to(\acts\to\Delta\obs)$ to produce a history. $\theta\bowtie\pi:\Delta\left((\acts\times\obs)^{T}\right)$ is the distribution on histories produced by $\theta$ and $\pi$ interacting. 

A model $M$ is of type $\acts\to\Box\obs$ and $\hypo\subseteq\acts\to\Box\obs$ is a hypothesis class. To link environments $\theta$ and models $M$, we say that $\theta$ is \emph{consistent} with $M$ if, for all $h,a$, $\theta(h)(a)\in M(a)$. This is written as $\theta\models M$. If $\theta$ is the true environment, and $\theta\models M$, we say that $M$ is a true model\footnote{In particular, there may be \emph{multiple} true models.} The notation $M^{*}$ denotes an arbitrary true model.

The (worst-case) expected reward of a model $M$ for an action $a$ is denoted as $f^{M}(a)$, which abbreviates $\bel{\min}{\mu\in M(a)}\expec_{o\sim\mu}[r(a,o)]$. $\max(f^{M})$ abbreviates $\bel{\max}{a\in\acts}f^{M}(a)$, the maximin expected reward for $M$. We define the regret of an algorithm $\pi$ against a model $M$ for $T$ timesteps as
$$\textbf{REG}(\pi,M,T):=\max_{\theta:\theta\models M}\left(T\cdot\max\left(f^{M}\right)-\expec_{\theta\bowtie\pi}\left[\tsum r(a_{t},o_{t})\right]\right)$$

Something to note is that any algorithm $\pi$ with low regret on all $M\in\hypo$ will exploit non-adversarial environments $\theta$, in the sense that the average reward against $\theta$ will be comparable to or exceed the most optimistic maximin value $\max_{M:\theta\models M}\max(f^{M})$.
\section{Generalizing the Decision-Estimation Coefficient}
The decision-estimation coefficient (DEC) was introduced in \cite{Foster21} to quantify the difficulty of learning to behave optimally in various classical reinforcement learning problems, and provides nearly matching upper and lower bounds on minimax regret.

The definition of the DEC and its variants depends on the \emph{Hellinger distance} between probability distributions, which is defined as
$$D_{H}(\mu,\nu):=\sqrt{\frac{1}{2}\int\left(\sqrt{\frac{d\mu}{d\xi}}-\sqrt{\frac{d\nu}{d\xi}}\right)^{2}d\xi}$$
where $\mu$ and $\nu$ are absolutely continuous with respect to $\xi$. The choice of $\xi$ doesn't affect the Hellinger distance. $\hells(\mu,\nu)$ is the square of the Hellinger distance.
\\
\\
There are many variants of the decision-estimation coefficient, and we introduce the constrained DEC from \cite{FosterGH23} as an illustrative example. This quantity depends on the hypothesis space $\hypo$, a belief $\gen:\acts\to\Delta\obs$, and a parameter $\eps>0$, and is defined in the classical case as
$$\text{dec}^{c}_{\eps}\left(\hypo,\gen\right):=\min_{p\in\Delta\acts}\max_{M\in\hypo}\Set{\max\left(f^{M}\right)-\bel{\expec}{a\sim p}\left[f^{M}(a)\right]}{\bel{\expec}{a\sim p}\left[\hells\left(\gen(a),M(a)\right)\right]\leq\eps^{2}}$$
This quantity is relevant to decision-making because a good algorithm should act to either attain low regret, or acquire information to distinguish models. If actions are selected from $p$, the $M$ where $\bel{\expec}{a\sim p}[\hells(\gen(a),M(a))]\leq\eps^{2}$ nearly mimic the belief $\gen$ and the information gained doesn't effectively distinguish among this cluster of models. Therefore, the regret should be low among this cluster, which justifies minimizing the maximum regret among near-mimic models.

To generalize the DEC to our setting, we must first change our notion of what counts as a "near-mimic" of the beliefs $\gen$. If $\Psi$ is our belief state, and $\Psi\subseteq\Phi$, then any distribution from $\Psi$ could have been produced by $\Phi$, so $\Phi$ can mimic $\Psi$. This motivates the following definition. 
\begin{defi}[Asymmetric Distance]
Given a distance metric $D$ on $\Delta X$, we define the asymmetric distance from the set $\Psi$ to $\Phi$ as
\[D(\Psi\to\Phi):=\bel{\max}{\mu\in\Psi}\ \bel{\min}{\nu\in\Phi}D(\mu,\nu)\]
\end{defi}

$D(\Psi\to\Phi)$ is low when $\Psi$ is almost a subset of $\Phi$. For $D^{2}(\Psi\to\Phi)$, it doesn't matter whether the squaring happens inside or outside the maximin. Accordingly, the models $M$ where $\bel{\expec}{a\sim p}[\hells(\gen(a)\to M(a))]\leq\eps^{2}$ are considered to be the "near-mimics" of $\gen$.

Our second change is to the regret term $\max(f^{M})-\expec_{a\sim p}[f^{M}(a)]$. Our chosen notion of regret is the gap between the maximin reward and the expected reward against the true environment $\theta$, but $\expec_{a\sim p}[f^{M}(a)]$ is the expected \emph{worst-case} reward. We do not know $\theta$, so we substitute with the expected reward against the \emph{beliefs} $\gen$ so the regret term becomes $\max(f^{M})-\expec_{a\sim p}[f^{\gen}(a)]$. The expected reward (according to $\gen$) still needs to be linked to the true reward, but this is a matter of prediction, not decision-making.

We now proceed to our third change to the definition of the DEC. The E2D+ algorithm from \cite{FosterGH23}, with a regret bound that scaled with the constrained DEC, did not generalize well to our setting. The primary obstacle was that no regret guarantee could be shown for models $M$ which were too far from $\gen$ to count as a near-mimic. To remedy this matter, we introduce the "fuzzy DEC", which is a relaxation of the constrained DEC. To present the definition, we use sub-probability distributions, which are measures $\mu$ with $\sum_{x}\mu(x)\in[0,1]$. The space of these measures is notated as $\Delta^{s}X$. Given some $\mu:\Delta^{s}X$ and $\nu:\Delta Y$, the notation $\expec_{x,y\sim\mu,\nu}[f(x,y)]$ abbreviates $\sum_{x,y}\mu(x)\nu(y)f(x,y)$.
\begin{defi}[Fuzzy Decision-Estimation Coefficient]
\[\text{dec}^{f}_{\eps}(\hypo,\gen):=\bel{\min}{p\in\Delta\acts}\bel{\max}{\mu\in\Delta^{s}\hypo}\Set{\bel{\expec}{M,a\sim\mu,p}
[\max(f^{M})-f^{\overline{M}}(a)]}{\bel{\expec}{M,a\sim\mu,p}\left[\hells(\gen(a)\to M(a))\right]\leq\eps^{2}}\]
\end{defi}

Comparing this to the constrained DEC, we see that the fuzzy decision-estimation coefficient implements a softer cutoff for which models count as near-mimics of $\gen$. If $\hells(\gen(a)\to M(a))=2\eps^{2}$, then $\mu$ can put $0.5$ measure on $M$ and fulfill the strict distance requirement, causing the regret term in the fuzzy DEC to be half of the true regret against $M$. This imposes nontrivial regret bounds on models which aren't near-mimics of $\gen$.

We now adapt another variant of the DEC from \cite{FosterGH23}.
\begin{defi}[Offset Decision-Estimation Coefficient]
\[\text{dec}^{o}_{\gamma}(\hypo,\overline{M}):=\bel{\min}{p\in\Delta\acts}\bel{\max}{M\in\hypo}\left(\max(f^{M})-\bel{\expec}{a\sim p}\left[f^{\gen}(a)\right]-\gamma\bel{\expec}{a\sim p}\left[\hells(\gen(a)\to M(a))\right]\right)\]
\end{defi}

These quantities are closely related, and the fuzzy DEC can be viewed as the offset DEC with the optimal choice of the information-to-regret ratio $\gamma$.
\begin{prop}
If $\hypo$ is a compact subset of $\acts\to\Box\obs$ then
\[\text{dec}^{f}_{\eps}(\hypo,\gen)=\bel{\min}{\gamma\ge 0}\left(\max(\text{dec}^{o}_{\gamma}(\hypo,\gen),0)+\gamma\eps^{2}\right)\]
\end{prop}

Given $\hypo,\eps$, we can consider the worst-case value of the fuzzy DEC, and define
$$\text{dec}^{f}_{\eps}(\hypo):=\bel{\max}{\gen\in\acts\to\Box\obs}\text{dec}^{f}_{\eps}(\hypo,\gen)$$

It is natural to ask whether restricting to probabilistic beliefs of type $\acts\to\Delta\obs$ can decrease the worst-case fuzzy DEC. We answer this question negatively.
\begin{prop}
For every $\gen:\acts\to\Box\obs$, there is a probabilistic model $\overline{\theta}:\acts\to\Delta\obs$ consistent with $\gen$ where $\text{dec}^{f}_{\eps}(\hypo,\gen)\leq\text{dec}^{f}_{\eps}(\hypo,\overline{\theta})$
\end{prop}

When proving upper bounds, Proposition 2 lets us assume that $\gen$ is probabilistic.
\subsection{Upper Bounds with the E2D Algorithm}

The Estimations to Decisions (E2D) algorithm was introduced in \cite{Foster21}, and reduces decision-making to online estimation. An estimation oracle produces beliefs $\guess$ on each timestep, and actions are selected by sampling from the distribution $p$ which minimizes the DEC. For all reinforcement learning problems, this process gives an upper bound on regret which scales with the DEC and the performance of the estimation oracle. Accordingly, we must introduce some notation to quantify the quality of an estimation oracle.

Given a history and a timestep $t$, let $\guess:\acts\to\Box\obs$, $\pi_{t}:\Delta\acts$, and $\theta_{t}:\acts\to\Delta\obs$ denote the estimated model, the distribution over actions, and the behavior of the environment on that timestep. Let $\mathcal{L}:(\acts\to\Box\obs)\times(\acts\to\Box\obs)\times\acts\to\mathbb{R}^{\geq 0}$ be some function which measures how far apart two models are for a given action. By default, we have $\mathcal{L}(\gen,M,a)=\hells(\gen(a)\to M(a))$.
\begin{defi}[Inaccuracy Bound]
A function $\beta:\mathbb{N}^{>0}\times[0,1]\to\mathbb{R}^{\geq 0}$ is an inaccuracy bound on an online estimator $\estim$ for $\mathcal{L}$ if, for all $T,\delta$, algorithms $\pi$, models $M\in\hypo$, and environments $\theta\models M$, with $1-\delta$ probability according to $\theta\bowtie\pi$, we have
\[\tsum\bel{\expec}{a\sim\pi_{t}}\left[\mathcal{L}(\guess,M,a)\right]\leq\beta(T,\delta)\]
\end{defi}
\begin{defi}[Optimism Bound]
A function $\alpha:\mathbb{N}^{>0}\times[0,1]\to\mathbb{R}^{\geq 0}$ is an optimism bound on an online estimator $\estim$ if, for all $T,\delta$, algorithms $\pi$, models $M\in\hypo$, and environments $\theta\models M$, with $1-\delta$ probability according to $\theta\bowtie\pi$, we have
\[\tsum\bel{\expec}{a\sim\pi_{t}}\left[f^{\guess}(a)-f^{\theta_{t}}(a)\right]\leq\alpha(T,\delta)\]
\end{defi}

$\beta_{\estim}$ and $\alpha_{\estim}$ denote inaccuracy and optimism bounds for an estimator. In the classical case, low inaccuracy implies low optimism, but this doesn't hold here\footnote{A toy example is the multi-armed bandit problem, where a true model $M^{*}$ says that a certain arm returns $\geq 0.7$ reward. If the environment $\theta$ reliably returns $0.8$ reward on that arm, while the models $\guess$ keep predicting $0.9$ reward on that arm, the sequence $\guess$ is perfectly accurate relative to $M^{*}$, but the optimism is high.}. Both quantities appear in the regret bound for the Estimations to Decisions algorithm, so high-quality regret bounds require an estimation oracle with a low $\alpha$ and $\beta$. If an online estimator predicts by averaging past outcomes together, the optimism bound may be linear in $T$, so more sophisticated approaches are needed.

The variant of the E2D algorithm we present here depends on the time horizon $T$, for ease of analysis. However, converting it into an anytime algorithm with comparable performance may be done via the doubling trick (see e.g. \cite{Besson18,Lattimore2020}), as well as more refined methods.

\begin{algorithm}
\caption{Estimations to Decisions (E2D)}
\textbf{Parameters:} Rounds $T$, failure odds $\delta$, oracle $\estim$, loss $\mathcal{L}$, hypothesis class $\hypo$,\;

inaccuracy bound $\beta_{\estim}$ for $\estim,\mathcal{L},\hypo$.\;

$\hypo_{1}\gets\hypo$\;

$h_{<1}\gets\varnothing$\;

\For{$1\leq t\leq T$}{
    $\guess\gets\estim(h_{<t})$\;
    
    $p_{t}\gets\bel{\text{argmin}}{p\in\Delta\acts}\bel{\max}{\mu\in\Delta^{s}\hypo_{t}}\Set{\bel{\expec}{\mu,p}\left[\max(f^{M})-f^{\guess}(a)\right]}{\bel{\expec}{\mu,p}\left[\mathcal{L}(\guess,M,a)\right]\leq\frac{\beta_{\estim}(T,\delta)}{T}}$\;
    
    \Return{$a_{t}\sim p_{t}$}\;
    
    Receive $o_{t}$ from the environment.\;
    
    $\hypo_{t+1}\gets\hypo\cap\Set{M}{\sum_{k=1}^{t}\bel{\expec}{a\sim p_{k}}\left[\mathcal{L}(\widehat{M}_{k},M,a)\right]\leq\beta_{\estim}(T,\delta)}$\;
    
    $h_{<t+1}\gets h_{<t},a_{t},o_{t}$\;
}
\end{algorithm}
Given a function $\mathcal{L}$, $\text{dec}^{f,\mathcal{L}}_{\eps}(\hypo)$ is the variant of the fuzzy DEC with $\expec_{M,a\sim\mu,p}\left[\mathcal{L}(\gen,M,a)\right]$ in place of $\expec_{M,a\sim\mu,p}\left[\hells(\gen(a)\to M(a))\right]$.
\begin{theo}
For all $T,\delta,\mathcal{L}$, oracles $\estim$, and models $M\in\hypo$, if $\eps=\sqrt{\frac{\beta_{\estim}(T,\delta)}{T}}$, we have
\[\textbf{REG}(\text{E2D},M,T)\leq 2T\text{dec}^{f,\mathcal{L}}_{\eps}(\hypo)+\alpha_{\estim}(T,\delta)+2T\delta\]
\end{theo}

Theorem 1 lets us produce an upper bound on regret for any problem in our framework by picking a function $\mathcal{L}$, upper-bounding the fuzzy DEC for $\mathcal{L}$, constructing an estimator $\estim$, and upper-bounding $\beta_{\estim}$ and $\alpha_{\estim}$.

A robust reinforcement learning problem is said to be \emph{learnable} if there is some algorithm $\pi$ (which may depend on the timestep $T$), and $p<1$, where
$$\bel{\max}{M\in\hypo}(\textbf{REG}(\pi,M,T))\in\mathcal{O}(T^{p})$$

In one direction, if the fuzzy DEC shrinks as $\eps^{p}$ for some $p>0$, and there is an estimator with sub-linear inaccuracy and optimism bounds, the E2D algorithm witnesses learnability.
\begin{coro}If there is a $p>0,q>0,r<1,s<1$ and online estimator $\estim$ such that
\[\limsup_{\eps\to 0}\frac{\text{dec}^{f}_{\eps}(\hypo)}{\eps^{p}}<\infty, \limsup_{T\to\infty}\frac{\alpha_{\estim}(T,T^{-q})}{T^{r}}<\infty, \limsup_{T\to\infty}\frac{\beta_{\estim}(T,T^{-q})}{T^{s}}<\infty\]
then the E2D algorithm with $\estim$ as an oracle has $\mathcal{O}\left(T^{\max\left(1-\frac{p(1-s)}{2},r,1-q\right)}\right)$ regret\footnote{Note that the quantity in the exponent is less than 1, witnessing learnability} on all $M\in\hypo$.
\end{coro}

\subsection{Lower Bounds}
\begin{prop}
For all $p\in(0,1)$ such that $\liminf_{\eps\to 0}\frac{\text{dec}^{f}_{\eps}(\hypo)}{\eps^{p}}=\infty$, if $\eps_{T}=\sqrt{\frac{1}{T\ln(T)}}$, we have
$$\bel{\min}{\pi}\bel{\max}{M\in\hypo}(\textbf{REG}(\pi,M,T))\in\Omega\left(T\cdot\text{dec}^{f}_{\eps_{T}^{\frac{1}{1-p}}}(\hypo)\right)$$
\end{prop}

This lower bound relies on Hellinger distance, and cannot be adapted to general notions of estimation error. To compare the upper bound of Theorem 1 and the lower bound of Proposition 3, consider three cases where $\text{dec}^{f}_{\eps}(\hypo)$ scales as $\eps,\eps^{1/2}$, and $\eps^{1/3}$. Neglecting estimation complexity, the upper bound of Theorem 1 would scale as $T^{1/2},T^{3/4}$, and $T^{5/6}$. The lower bound above would be inapplicable in the first case, and scale as $T^{1/2}$ and $T^{3/4}$ in the other two cases. The lower bound on regret in the classical setting in \cite{FosterGH23} is significantly tighter than this result. However, our lower bound still establishes that slow decay of the fuzzy DEC implies unlearnability.
\begin{coro}
If, for all $p>0$, we have $\liminf_{\eps\to 0}\frac{\text{dec}^{f}_{\eps}(\hypo)}{\eps^{p}}=\infty$, then, for all $q<1$ and $\pi$ (which may depend on $T$), we have $\liminf_{T\to\infty}\frac{\max_{M\in\hypo}(\textbf{REG}(\pi,M,T))}{T^{q}}=\infty$
\end{coro}

Putting Corollaries 1 and 2 together, if the fuzzy DEC declines more slowly than $\eps^{p}$ for every $p>0$, then unlearnability holds. If the fuzzy DEC declines more quickly than $\eps^{p}$ for some $p>0$, and sub-linear estimation error is attainable, the E2D algorithm certifies learnability.
\section{Robust Online Estimation}
\subsection{The Robust Universal Estimator}
We will now present an explicit online estimation algorithm which is suitable for use as an oracle, and computable if the sets $\obs$ and $\hypo$ are finite. The RUE algorithm (Robust Universal Estimator) can be thought of as a prediction market. There is a set of bettors, $\mathcal{B}$, which have wealth and make bets against a market. The market odds $\guess:\acts\to\Delta\obs$ are generated, and the bettors bet against the market odds if they disagree. An outcome $a_{t},o_{t}$ is observed, and the bets resolve, which updates the wealth distribution $\zeta_{t}$. The key part of RUE is that $\guess$ is set to ensure that the total bettor wealth is conserved no matter which outcome occurs. A bettor which reliably profits will acquire more than all of the wealth, which is impossible.

To enforce accuracy, we associate each model $M\in\hypo$ with a bettor which reliably profits if $M$ is a true model and $\expec_{a\sim\pi_{t}}\left[\hells\left(\guess(a)\to M(a)\right)\right]$ is high\footnote{Reliable profit is impossible, so if $M$ is a true model, the sum of expected Hellinger-squared error must be low.}. To enforce non-optimism, we introduce a pessimistic bettor $\bullet$, which reliably profits if $\expec_{a\sim\pi_{t}}\left[f^{\guess}(a)-f^{\theta_{t}}(a)\right]$ is high\footnote{Reliable profit is impossible, so the sum of expected reward overestimation must be low}. There is also a uniform bettor $u$ which exists to avert division-by-zero errors. Given a bettor $B$ which isn't the pessimistic bettor, $M_{B}:\acts\to\Box\obs$ denotes their corresponding model. For the uniform bettor, their model maps all actions to the uniform distribution on $\obs$. 

Mixtures of convex functions are convex, and the squared Hellinger distance to a convex set is convex by Lemma 4, so all minimization in RUE is over convex functions.

\begin{algorithm}
\caption{Robust Universal Estimator (RUE)}
\textbf{Parameters:} Hypothesis class $\hypo$, rounds $T$, reward function $r$, prior $\zeta_{1}$\;

$\eps\gets\min\left(\frac{1}{2},\sqrt{\frac{\ln(2)}{T}}\right)$\;

\SetKwFunction{esti}{estimate}
\SetKwProg{Fn}{Function}{:}{}
\Fn{\esti{$\zeta$, $a$}}{
    \KwRet $\bel{\text{argmin}}{\mu\in\Delta\obs}\ \bel{\expec}{B\sim\zeta}\left[\text{ if }B\neq\bullet, 2\hells(\mu\to M_{B}(a)),\text{ else }\eps\cdot\bel{\expec}{o\sim\mu}\left[r(a,o)\right]\right]$\;

}
\SetKwFunction{upda}{update}
\Fn{\upda{$\zeta$, $\gen$, $a$, $o$}}{
    \For{$B\in\hypo\cup\{u\}$}{
        $\mu_{B}\gets\text{argmin}_{\mu\in M_{B}(a)}\ \hells(\gen(a),\mu)$\;

        $\xi(B)\gets\zeta(B)\cdot\left(\sqrt{\frac{\mu_{B}(o)}{\gen(a)(o)}}+\hells(\gen(a)\to M_{B}(a))\right)$\;
    }
    $\xi(\bullet)\gets\zeta(\bullet)\cdot
    \left(1+\eps(\expec_{o'\sim\gen(a)}[r(a,o')]-r(a,o))\right)$\;

    \KwRet $\xi$\;
}
\For{$1\leq t\leq T$}{
    $\guess\gets\lambda a.\esti{$\zeta_{t}$, $a$}$\;
    
    \KwRet $\guess$\;
    
    Receive $a_{t},o_{t}$ from the environment.\;
    
    $\zeta_{t+1}\gets\upda{$\zeta_{t}$, $\guess$, $a_{t}$, $o_{t}$}$\;
}
\end{algorithm}

\begin{prop}
If the uniform bettor has positive probability according to $\zeta_{1}$, the RUE algorithm never divides by zero and all $\zeta_{t}$ are probability distributions.
\end{prop}
\begin{theo} If our estimator $\estim$ is the RUE algorithm with a suitable choice of prior\footnote{Technically, the bound on $\alpha_{\estim}(T,\delta)$ in Theorem 2 only holds for $T>2$ in the limit as the uniform bettor probability according to $\zeta_{1}$ approaches zero. However, the uniform bettor is only present to avert division-by-zero errors, and any positive probability for it is sufficient to do so. If the starting probability of the uniform bettor is, say, $0.001$, this only has a minor effect on constants and doesn't affect the asymptotics with respect to $T$ and $\ln\left(\frac{1}{\delta}\right)$.}, then
\\
$\beta_{\estim}(T,\delta)\leq\ln\left(\frac{2|\hypo|}{\delta}\right)$, and $\alpha_{\estim}(T,\delta)\leq\sqrt{T}\left(2\sqrt{\ln(2)}+\sqrt{2\ln\left(\frac{1}{\delta}\right)}\right)$
\end{theo}

The "suitable choice of prior" is to assign an arbitrarily low probability $\eps'$ to the uniform bettor, $\frac{1}{2}-\eps'$ probability to the pessimistic bettor, and $\frac{1}{2|\hypo|}$ probability to all other bettors. The inaccuracy bound is close to the inaccuracy bound of Bayesian updating in the classical case.
\subsection{Covering Numbers}
The robust universal estimator only applies to finite hypothesis spaces, and must be generalized to infinite hypothesis spaces via covering numbers. We use the symmetric notion of Hausdorff distance between sets of distributions here, instead of asymmetric distance.
\begin{defi}[$\eps$-Covering Number]
The $\eps$-covering number of $\hypo$, $\mathcal{N}(\hypo,\eps)$, is defined as
$$\mathcal{N}(\hypo,\eps):=\min\Set{|X|}{X\subseteq\hypo,\forall M\in\hypo\ \exists N\in X:\bel{\max}{a\in\acts}D_{H}(N(a),M(a))\leq\eps}$$
\end{defi}
\begin{prop} If $\obs$ is finite, then for all hypothesis classes $\hypo$, there exists an online estimator $\estim$ where $\beta_{\estim}(T,\delta)\leq\bel{\min}{\eps>0}\left(2\ln\left(\frac{2\mathcal{N}(\hypo,\eps)}{\delta}\right)+8T\eps^{2}\right)$, and
$\alpha_{\estim}(T,\delta)\leq\sqrt{T}\left(2\sqrt{\ln(2)}+\sqrt{2\ln\left(\frac{1}{\delta}\right)}\right)$
\end{prop}

The estimator constructed in the proof of Proposition 5 works as follows. For each point $N$ in the minimal cover, a "fattened model" $N'$ is constructed where $N'(a)=\Set{\mu}{D_{H}(\mu\to N(a))\leq\eps}$. Then, RUE is run on the finite set of fattened models. This differs from the classical proof by using imprecision in an essential way. To simplify Proposition 5, we introduce the Minkowski-Bougliand dimension of $\hypo$, defined as
$$MB(\hypo):=\limsup_{\eps\to 0}\frac{\mathcal{N}(\hypo,\eps)}{\ln(\frac{1}{\eps})}$$

Swapping $\ln\left(\frac{2\mathcal{N}(\hypo,\eps)}{\delta}\right)$ for $\ln\left(\frac{2}{\delta}\right)+MB(\hypo)\cdot\ln\left(\frac{1}{\eps}\right)$ in Proposition 5 and minimizing over $\eps$ yields an estimation complexity in $\mathcal{O}\left(\log\left(\frac{1}{\delta}\right)+MB(\hypo)\cdot\log(T)\right)$, just as in the classical case.
\section{Novel Regret Bounds}
\subsection{Robust Linear Bandits}

We begin with a setting from \cite{Kosoy2024}, robust linear bandits. In this setting, $\acts$ and $\obs$ are large finite sets of actions and observations, and there is a reward function $r:\acts\times\obs\to[0,1]$. There is a vector space $\mathcal{Z}$, with a compact subset $\hypo$ of hypotheses. There is also an auxiliary vector space $\mathcal{W}$, and a function $F:\acts\times\mathcal{Z}\times\mathbb{R}^{\obs}\to\mathcal{W}$ that is bilinear in $\mathcal{Z}$ and $\mathbb{R}^{\obs}$. $F$ specifies how points in $\mathcal{Z}$ correspond to models. A point $z\in\mathcal{Z}$ corresponds to the model defined by
$$a\mapsto\Set{\mu\in\Delta\obs}{F(a,z,\mu)=0}$$

Kosoy's regret bound for this setting depended on several unusual quantities which are absent from the classical theory of linear bandits. $R$ is a parameter which acts as a generalized condition number, and $\text{sine}$ is another parameter which measures the angle between the walls of the probability simplex, and the set $\Set{\mu\in\Delta\obs}{F(a,z,\mu)=0}$.

If $Z,W$ are the dimensions of the spaces $\mathcal{Z}$ and $\mathcal{W}$, then the previous regret bound for this setting was of the form $\Tilde{\obs}\left(Z^{2}(\frac{1}{\text{sine}}+1)R\sqrt{WT}\right)$, which does not depend on $\acts$ or $\obs$.

To improve on this result, we upper-bound the fuzzy DEC and the $\eps$-covering number of $\hypo$.
\begin{theo}
In the robust linear bandit setting, for all $\eps<\frac{1}{e^{2}}$ (Euler's constant),
$$\text{dec}^{f}_{\eps}(\hypo)\leq 16\left(\frac{1}{\text{sine}}+1\right)R\sqrt{WZ}\eps\ln\left(\frac{1}{\eps}\right)$$
\end{theo}

\begin{prop}
For the robust linear bandit setting, $\mathcal{N}(\hypo,\eps)\leq\left(\frac{4\left(\frac{1}{\text{sine}}+1\right)RZ}{\eps^{2}}+1\right)^{Z}$
\end{prop}

The Minkowski-Bougliand dimension of $\hypo$ is then $2Z$ or less, so by Proposition 5 there is an estimator with $\beta_{\estim}(T,T^{-1/2})\in\mathcal{O}(Z\log(T))$, and $\alpha_{\estim}(T,T^{-1/2})\in\mathcal{O}(\sqrt{T\log(T)})$. Such an estimator, along with Theorems 1 and 3, show that a regret of $\Tilde{\mathcal{O}}\left(Z\left(\frac{1}{\text{sine}}+1\right)R\sqrt{WT}\right)$ is attainable in this setting, matching the classical linear bandit regret bound which scales linearly with the dimension of the hypothesis space. 
\subsection{Tabular Episodic RMDP Learning}

A Robust Markov Decision Process (RMDP) may be thought of as an MDP where the transition kernel $\imdp$ produces an imprecise belief over the next state and reward, consisting of the distributions which the environment might select.

We work in the non-stationary tabular setting where the states $\states$ and actions $\acts$ are finite sets, $H$ is the time horizon, and the transition kernel $\imdp$ has type  $\{0,...,H\}\times\states\times\acts\to\Box([0,1]\times\states)$. The initial state, initial action, and terminal state $s_{0},a_{0},s_{H+1}$ are considered to be unique, so trajectories are of the form $r_{0},s_{1},a_{1},r_{1},...,r_{H}$, which lets us specify an RMDP by the transition kernel alone.

To phrase episodic RMDP learning in our framework, we consider each episode to be a single interaction with the environment, so $T$ is the number of episodes. The action for an episode is the choice of policy. We only consider randomized nonstationary policies, also called Markov policies. This is the space $\{1,...,H\}\times\states\to\Delta\acts$, also denoted as $\Pi_{\text{RNS}}$. The observation for an episode is the trajectory. To convert an RMDP to a model of type $\Pi_{\text{RNS}}\to\Box\left([0,1]\times(\states\times\acts\times[0,1])^{\{1,...,H\}}\right)$ , we introduce the following definition.
\begin{defi}[Selection of an RMDP] \label{def5}
A function $\sigma:(\states\times\acts\times[0,1])^{\leq H}\times\states\times\acts\to\Delta([0,1]\times\states)$ is a selection of an RMDP $\imdp$ if, for all $h,tr_{<h},s,a$, we have $\sigma(tr_{<h},s,a)\in\imdp(h,s,a)$. This is denoted by $\sigma\models\imdp$.
\end{defi}

The selections of $\imdp$ are the possible behaviors of the environment. Given a selection $\sigma$, and a policy $\pi$, $\sigma\bowtie\pi$ denotes the distribution over trajectories produced by $\sigma$ interacting with $\pi$. The model associated with an RMDP $\imdp$ is then $\pi\mapsto\Set{\sigma\bowtie\pi}{\sigma\models\imdp}$. We do not require that the sum of rewards in an episode be bounded in $[0,1]$, although we impose an analogous requirement.
\begin{defi}[1-Bounded RMDP]
An RMDP $\imdp$ is 1-bounded if there is some $\sigma\models\imdp$ where, for all $\pi,h,s,a$, we have $\expec_{\sigma\bowtie\pi}\left[\sum_{k=h}^{H}r_{k}\ \middle|\ s_{h}=s,a_{h}=a\right]\leq 1$
\end{defi}

Our hypothesis class $\hypo$ of interest then consists of all 1-bounded RMDP's, with $\states,\acts,H$ fixed. The best existing regret bound in this setting is the bound of \cite{Tian0YS21}, which was of the form $\Tilde{\mathcal{O}}(H^{2}S^{1/3}A^{1/3}T^{2/3})$. \cite{Tian0YS21} did not assume 1-boundedness, which accounts for an extra factor of $H$ in their regret bound. We show that $\Tilde{\mathcal{O}}\left(\sqrt{\text{poly}(H,S,A)T}\right)$ regret is attainable.

To establish this we use Theorem 1 and a non-standard notion of loss detailed in Appendix H. The term "modified fuzzy DEC" denotes the fuzzy DEC with this new notion of loss.

In Section 5.2 of \cite{Foster21}, the DEC was upper-bounded in an analogous classical setting by an algorithm which randomized between Markov policies. We prove a comparable result in a more general setting, by showing that a \emph{single} Markov policy can certify that the modified fuzzy DEC is low.
\begin{theo}
In the episodic RMDP setting, if $\gen:\Pi_{RNS}\to\Delta([0,1]\times(\states\times\acts\times[0,1])^{\{1,...,H\}})$ is continuous and policy-coherent\footnote{A belief $\gen$ is policy-coherent if, for all $\pi$, there exists some $\sigma$ such that $\gen(\pi)=\sigma\bowtie\pi$. Policies must be mapped to distribution on trajectories which could have been produced by that policy.}, the modified fuzzy DEC fulfills $\text{dec}^{f'}_{\eps}(\hypo,\gen)\leq 2\sqrt{2(HSA+1)}\eps$
\end{theo}

We now turn to constructing an estimation oracle for $\hypo$. Unfortunately, this hypothesis class is infinite-dimensional. To address this, we retreat to a smaller hypothesis class $\hypo_{\text{parhalf}}$, of Partial Halfspace RMDP's\footnote{All Partial Halfspace RMDP's have the following form. For each $h,s$, there is a recommended action $a_{h,s}:\acts$, function $f_{h,s}:\states\to[0,1]$, and constant $c_{h,s}:[0,1]$. $\imdp(h,s,a)$ for non-recommended $a$ is $\Delta([0,1]\times\states)$, so the RMDP is "partial" by only predicting one action.
$\imdp(h,s,a_{h,s})=\Set{\mu\in\Delta([0,1]\times\states)}{\expec_{\mu}[f_{h,s}+r]\geq c_{h,s}}$, so the imprecise belief  for a recommended action is a halfspace.}, where estimation is tractable. The general insights of the RUE algorithm let us construct a custom estimator for $\hypo_{\text{parhalf}}$ with the properties we need, in Appendix L.
\begin{theo}
There is an online estimator $\estim$ where $\beta_{\estim}(T,\delta)\in\mathcal{O}\left(HS^{2}\log(T)+HS\log\left(\frac{HSAT}{\delta}\right)\right)$ and $\alpha_{\estim}(T,\delta)\in\mathcal{O}\left(H\sqrt{T}\log\left(\frac{1}{\delta}\right)\right)$, for hypotheses in $\hypo_{\text{parhalf}}$.
\end{theo}
\begin{prop}
For the estimator of Theorem 5, all estimates 
\\
$\guess:\Pi_{RNS}\to\Delta([0,1]\times(\states\times\acts\times[0,1])^{\{1,...,H\}})$ are continuous and policy-coherent.
\end{prop}

Combining these results with Theorem 1, we have an algorithm which attains $\Tilde{\mathcal{O}}\left(\sqrt{H^{2}S^{3}AT}\right)$ regret on all 1-bounded RMDP's in $\hypo_{\text{parhalf}}$. The final insight is that every $\imdp\in\hypo$ has a 1-bounded "surrogate" $\imdp'$ within $\hypo_{\text{parhalf}}$. $\imdp'$ will be a true model if $\imdp$ is, and low regret on $\imdp'$ implies low regret on $\imdp$, so an algorithm with low regret on $\hypo_{\text{parhalf}}$ has low regret on all of $\hypo$.

\begin{coro}
There is an algorithm which attains $\Tilde{\mathcal{O}}\left(\sqrt{H^{2}S^{3}AT}\right)$ regret on all 1-bounded RMDP's in the episodic RMDP setting.
\end{coro}

\section*{Acknowledgments}

This work was supported by the Machine Intelligence Research Institute in Berkeley, California, the Effective Ventures Foundation USA in San Francisco, California, the Advanced Research + Invention Agency (ARIA) in the United Kingdom, and the Survival and Flourishing Corporation.

\bibliographystyle{plain}
\bibliography{bib}

\appendix
\setcounter{theo}{0}
\setcounter{prop}{0}
\setcounter{coro}{0}
\section{Basic Propositions}
\begin{prop} \label{pro1}
If $\hypo$ is a compact subset of $\acts\to\Box\obs$, then
$$\text{dec}^{f}_{\eps}(\hypo,\gen)=\bel{\min}{\gamma\ge 0}\left(\max(\text{dec}^{o}_{\gamma}(\hypo,\gen),0)+\gamma\eps^{2}\right)$$
\end{prop}

Unpack definitions and reexpress the maximization.
$$\text{dec}^{f}_{\eps}(\hypo,\gen)$$
$$=\bel{\min}{p\in\Delta\acts}\bel{\max}{\mu\in\Delta^{s}\hypo}\Set{\bel{\expec}{M,a\sim\mu,p}\left[\max(f^{M})-f^{\gen}(a)\right]}{\bel{\expec}{M,a\sim\mu,p}\left[\hells(\gen(a)\to M(a))\right]\leq\eps^{2}}$$
$$=\bel{\min}{p\in\Delta\acts}\bel{\max}{\mu\in\Delta^{s}\hypo}\bel{\min}{\gamma\geq 0}\left(\bel{\expec}{\mu,p}\left[\max(f^{M})-f^{\gen}\right]-\gamma\left(\bel{\expec}{\mu,p}\left[\hells(\gen(a)\to M(a))\right]-\eps^{2}\right)\right)$$

If $\hypo$ is a compact subspace of $\acts\to\Box\obs$, we can use Sion's Minimax Theorem.
$$=\bel{\min}{p\in\Delta\acts}\bel{\min}{\gamma\geq 0}\bel{\max}{\mu\in\Delta^{s}\hypo}\left(\bel{\expec}{\mu,p}\left[\max(f^{M})-f^{\gen}\right]-\gamma\left(\bel{\expec}{\mu,p}\left[\hells(\gen(a)\to M(a))\right]-\eps^{2}\right)\right)$$

Now, swap the two mins, distribute the $\gamma$, regroup parentheses, and reshuffle expectations.
$$=\bel{\min}{\gamma\geq 0}\left(\bel{\min}{p\in\Delta\acts}\bel{\max}{\mu\in\Delta^{s}\hypo}\expec_{\mu}\left[\max(f^{M})-\expec_{p}\left[f^{\gen}\right]-\gamma\expec_{p}\left[\hells(\gen(a)\to M(a))\right]\right]\right)+\gamma\eps^{2}$$

For the maximization over sub-probability distributions, the maximizer will be either 100 percent probability on a specific model, or the zero measure.
$$=\bel{\min}{\gamma\geq 0}\left(\bel{\min}{p\in\Delta\acts}\max\left(0,\bel{\max}{M\in\hypo}\left(\max(f^{M})-\expec_{p}\left[f^{\gen}\right]-\gamma\expec_{p}\left[\hells(\gen(a)\to M(a))\right]\right)\right)\right)+\gamma\eps^{2}$$
$$=\bel{\min}{\gamma\geq 0}\max\left(0,\bel{\min}{p\in\Delta\acts}\bel{\max}{M\in\hypo}\left(\max(f^{M})-\expec_{p}\left[f^{\gen}\right]-\gamma\expec_{p}\left[\hells(\gen(a)\to M(a))\right]\right)\right)+\gamma\eps^{2}$$
$$=\bel{\min}{\gamma\geq 0}\left(\max(\text{dec}^{o}_{\gamma}(\hypo,\gen),0)+\gamma\eps^{2}\right)$$
$\blacksquare$
\begin{prop} \label{pro2}
For every $\gen:\acts\to\Box\obs$, there is a probabilistic model $\overline{\theta}:\acts\to\Delta\obs$ consistent with $\gen$ where $\text{dec}^{f}_{\eps}(\hypo,\gen)\leq\text{dec}^{f}_{\eps}(\hypo,\overline{\theta})$
\end{prop}

Let $\overline{\theta}(a):=\bel{\text{argmin}}{\mu\in\gen(a)}f^{\mu}(a)$. This is the probabilistic model which attains minimal expected reward while being consistent with $\gen$. Unpacking the definition of the fuzzy DEC, we have
$$\text{dec}^{f}_{\eps}(\hypo,\gen)$$
$$=\bel{\min}{p\in\Delta\acts}\bel{\max}{\mu\in\Delta^{s}\hypo}\Set{\bel{\expec}{M,a\sim\mu,p}
\left[\max(f^{M})-f^{\gen}(a)\right]}{\bel{\expec}{M,a\sim\mu,p}\left[\hells(\gen(a)\to M(a))\right]\leq\eps^{2}}$$

Swapping out $\gen$ for $\overline{\theta}$ does not affect the regret terms, because $f^{\gen}(a)=\expec_{\gen(a)}[r]=\expec_{\overline{\theta}(a)}[r]=f^{\overline{\theta}}(a)$. However, swapping out the set $\gen(a)$ for the point $\overline{\theta}(a)$ can only decrease the greatest distance from the beliefs to $M$, for all $M$. Because the Hellinger distances all go down, more sub-probability distributions $\mu\in\Delta^{s}\hypo$ count as near-mimics of our beliefs, so the maximum expected regret over near-mimics can only go up. So, we have
$$\leq\bel{\min}{p\in\Delta\acts}\bel{\max}{\mu\in\Delta^{s}\hypo}\Set{\bel{\expec}{M,a\sim\mu,p}
\left[\max(f^{M})-f^{\overline{\theta}}(a)\right]}{\bel{\expec}{M,a\sim\mu,p}\left[\hells(\overline{\theta}(a)\to M(a))\right]\leq\eps^{2}}=\text{dec}^{f}_{\eps}(\hypo,\overline{\theta})$$
$\blacksquare$
\section{Upper Bounds}
\begin{theo} \label{the1}
For all $T,\delta,\mathcal{L}$, oracles $\estim$, and models $M\in\hypo$, if $\eps=\sqrt{\frac{\beta_{\estim}(T,\delta)}{T}}$, we have
$$\textbf{REG}(\text{E2D},M,T)\leq 2T\text{dec}^{f,\mathcal{L}}_{\eps}(\hypo)+\alpha_{\estim}(T,\delta)+2T\delta$$
\end{theo}

$$\textbf{REG}(\text{E2D},M,T)=\bel{\max}{\theta\models M}\bel{\expec}{\theta\bowtie\text{E2D}}\left[\tsum\bel{\expec}{a\sim\text{E2D}_{t}}\left[\max(f^{M})-f^{\theta_{t}}(a)\right]\right]$$

Fix a maximizing environment $\theta$ which is consistent with $M$, and reexpress the regret. Abbreviate $\text{E2D}_{t}$ as $p_{t}$.
$$=\bel{\expec}{\theta\bowtie\text{E2D}}\left[\tsum\bel{\expec}{a\sim p_{t}}\left[\max(f^{M})-f^{\theta_{t}}(a)\right]\right]$$
$$=\bel{\expec}{\theta\bowtie\text{E2D}}\left[\tsum\max(f^{M})-\expec_{p_{t}}\left[f^{\guess}\right]+\expec_{p_{t}}\left[f^{\guess}-f^{\theta_{t}}\right]\right]$$

$M$ is falsely rejected from the sequence of hypotheses $\hypo_{t}$ iff 
\\
$\tsum\bel{\expec}{a\sim p_{t}}\left[\mathcal{L}(\guess,M,a)\right]>\beta_{\estim}(T,\delta)$, and by the definition of $\beta_{\estim}$ with respect to $\mathcal{L}$, this only happens with $\leq\delta$ probability. Similarly, there is a $\leq\delta$ probability of the $\alpha_{\estim}$ bound failing to apply. So, we may upper-bound the expected regret by adding $2\delta T$ to account for the regret if a failure event occurs, and passing to a history $h$ where no failure events occur but the regret is otherwise as high as possible.
$$\leq\tsum\left(\max(f^{M})-\expec_{p_{t}}\left[f^{\guess}\right]+\expec_{p_{t}}\left[f^{\guess}-f^{\theta_{t}}\right]\right)+2\delta T$$

The $\alpha_{\estim}$ bound holds for this history, so we may rewrite as
$$\leq\tsum\left(\max(f^{M})-\expec_{p_{t}}\left[f^{\guess}\right]\right)+\alpha_{\estim}(T,\delta)+2\delta T$$

Let $\eps$ be defined as $\sqrt{\frac{\beta_{\estim}(T,\delta)}{T}}$, and $\mu_{t}\in\Delta^{s}\hypo_{t}$ be defined as follows. If $\expec_{p_{t}}\left[\mathcal{L}(\guess,M,a)\right]\leq\eps^{2}$, then $\mu_{t}$ assigns $M$ a probability of 1. If not, $\mu_{t}$ assigns $M$ a probability of $\frac{\eps^{2}}{\expec_{p_{t}}\left[\mathcal{L}(\guess,M,a)\right]}$. We always have $\mu_{t}\in\Delta^{s}\hypo_{t}$ because $M$ isn't falsely rejected from any $\hypo_{t}$. We can now rewrite as
$$=\tsum\max\left(1,\frac{\expec_{p_{t}}\left[\mathcal{L}(\guess,M,a)\right]}{\eps^{2}}\right)\bel{\expec}{M'\sim\mu_{t}}\left[\max(f^{M'})-\expec_{p_{t}}\left[f^{\guess}\right]\right]+\alpha_{\estim}(T,\delta)+2\delta T$$

By the definition of the E2D algorithm, $p_{t}$ is the fuzzy-DEC (with respect to $\mathcal{L}$)-minimizing distribution on actions. Also, $\mu_{t}$ was defined to ensure that $\bel{\expec}{\mu_{t},p_{t}}\left[\mathcal{L}(\guess,M,a)\right]\leq\eps^{2}=\frac{\beta_{\estim}(T,\delta)}{T}$, so we can upper-bound with the fuzzy DEC with respect to $\mathcal{L}$.
$$\leq\tsum\max\left(1,\frac{\expec_{p_{t}}\left[\mathcal{L}(\guess,M,a)\right]}{\eps^{2}}\right)\text{dec}^{f,\mathcal{L}}_{\eps}(\hypo_{t},\guess)+\alpha_{\estim}(T,\delta)+2\delta T$$

Now, $\hypo_{t}\subseteq\hypo$. The DEC is larger for larger hypotheses classes, and worst-casing over estimates makes it larger still.
$$\leq\tsum\max\left(1,\frac{\expec_{p_{t}}\left[\mathcal{L}(\guess,M,a)\right]}{\eps^{2}}\right)\text{dec}^{f,\mathcal{L}}_{\eps}(\hypo)+\alpha_{\estim}(T,\delta)+2\delta T$$

Upper bound $\max(a,b)$ by $a+b$, and use that $\eps=\sqrt{\frac{\beta_{\estim}(T,\delta)}{T}}$.
$$\leq\tsum\left(1+\frac{T}{\beta_{\estim}(T,\delta)}\right)\expec_{p_{t}}\left[\mathcal{L}(\guess,M,a)\right]\text{dec}^{f,\mathcal{L}}_{\eps}(\hypo)+\alpha_{\estim}(T,\delta)+2\delta T$$

On this history, no failure events occur, so the sum of the expected loss is upper-bounded by $\beta_{\estim}(T,\delta)$, which cancels, yielding
$$\leq 2T\text{dec}^{f,\mathcal{L}}_{\eps}(\hypo)+\alpha_{\estim}(T,\delta)+2T\delta$$
$\blacksquare$
\begin{coro} \label{cor1}
If there is a $p>0,q>0,r<1,s<1$ and online estimator $\estim$ such that:
$$\limsup_{\eps\to 0}\frac{\text{dec}^{f}_{\eps}(\hypo)}{\eps^{p}}<\infty, \limsup_{T\to\infty}\frac{\alpha_{\estim}(T,T^{-q})}{T^{r}}<\infty, \limsup_{T\to\infty}\frac{\beta_{\estim}(T,T^{-q})}{T^{s}}<\infty$$
then the E2D algorithm with $\estim$ as an oracle has $\mathcal{O}\left(T^{\max\left(1-\frac{p(1-s)}{2},r,1-q\right)}\right)$ regret on all $M\in\hypo$.
\end{coro}

The regret bound produced by the E2D algorithm is $\mathcal{O}(T\text{dec}^{f}_{\eps}(\hypo)+\alpha_{\estim}(T,\delta)+T\delta)$ by Theorem 1. Now, let $\delta$ scale as $T^{-q}$, and use that $\alpha(T,T^{-q})\in\mathcal{O}(T^{r})$, and $\text{dec}^{f}_{\eps}(\hypo)\in\mathcal{O}(\eps^{p})$, to get a regret on the order of $\mathcal{O}(T\eps^{p}+T^{r}+T^{1-q})$. Now, use the specific choice of $\eps$ from the E2D algorithm to show
$$\eps=\sqrt{\frac{\beta_{\estim}(T,\delta)}{T}}=\sqrt{\frac{\beta_{\estim}(T,T^{-q})}{T}}\in\mathcal{O}\left(\sqrt{\frac{T^{s}}{T}}\right)=\mathcal{O}\left(T^{\frac{1}{2}(s-1)}\right)$$

Plugging this value of $\eps$ in, we get that the regret of the E2D algorithm is on the order of
$$\mathcal{O}\left(T^{1+\frac{p}{2}(s-1)}+T^{r}+T^{1-q}\right)=\mathcal{O}\left(T^{\max\left(1-\frac{p(1-s)}{2},r,1-q\right)}\right)$$
$\blacksquare$
\section{Lower Bounds}
\begin{prop} \label{pro3}
For all $p\in(0,1)$ such that $\liminf_{\eps\to 0}\frac{\text{dec}^{f}_{\eps}(\hypo)}{\eps^{p}}=\infty$, if $\eps_{T}=\sqrt{\frac{1}{T\ln(T)}}$, we have
$$\min_{\pi}\max_{M\in\hypo}(\textbf{REG}(\pi,M,T))\in\Omega(T\text{dec}^{f}_{\eps_{T}^{\frac{1}{1-p}}}(\hypo))$$
\end{prop}

For this proof, define the following quantities, which may depend on $\hypo,\pi,T,p$.

$u(T)$ is the uniform distribution on timesteps. $\gen:\acts\to\Delta\obs$ is the belief which maximizes $\text{dec}^{f}_{\eps_{T}^{\frac{1}{1-p}}}(\hypo,\gen)$. The fuzzy DEC will never be zero, or else we would get a contradiction with the assumption that the DEC shrinks slower than $\eps^{p}$ for some $p\in(0,1)$. Proposition \ref{pro2} enables us to assume that $\gen$ is probabilistic with no loss of generality.

Given a model $M\in\hypo$, we can define $\theta^{M}:\acts\to\Delta\obs$ as $M$'s closest approximation to $\gen$ in Hellinger distance.
$$\theta^{M}(a):=\bel{\text{argmin}}{\mu\in M(a)}\hells(\mu,\gen(a))$$

$p^{M}$ and $\overline{p}$, of type $\Delta\acts$, are the distributions over actions produced by the policy $\pi$ interacting with $\theta^{M}$ or $\gen$ respectively, where $\theta^{M}$ and $\gen$ are treated as environments. In the definition, $\delta_{a_{t}}$ is the probability distribution which assigns all measure to action $a_{t}$. $\overline{p}$ is defined similarly.
$$p^{M}:=\bel{\expec}{\theta^{M}\bowtie\pi}[\bel{\expec}{t\sim u(T)}[\delta_{a_{t}}]]$$

$\mu'$ is the sub-probability distribution on $\hypo$ which maximizes regret against $\overline{p}$, while being a near-mimic of $\gen$. More specifically, it is
$$\mu':=\bel{\text{argmax}}{\mu\in\Delta^{s}\hypo}\Set{\bel{\expec}{\mu,\overline{p}}\left[\max(f^{M})-f^{\gen}\right]}{\bel{\expec}{\mu,\overline{p}}\left[\hells(\gen(a)\to M(a))\right]\leq\left(\eps_{T}^{\frac{1}{1-p}}\right)^{2}}$$

$\mu'$ cannot be the all-zero sub-probability distribution, because if it was, that would witness that the fuzzy DEC for $\gen$ is zero, which is impossible. So, $\mu'$ can be uniquely written as $\lambda\mu$, where $\lambda\in(0,1]$, and $\mu\in\Delta\hypo$. We now proceed with the proof. Fix an arbitrary $\pi$, and we will lower-bound the maximum regret. Start by replacing the max over $M$ with an expectation, and chose $\theta^{M}$ for each $M$.
$$\bel{\max}{M\in\hypo}(\textbf{REG}(\pi,M,T))=\bel{\max}{M\in\hypo,\theta\models M}\bel{\expec}{\theta\bowtie\pi}\left[\tsum\max(f^{M})-r_{t}\right]$$
$$\geq\bel{\expec}{M\sim\mu}\left[\bel{\expec}{\theta^{M}\bowtie\pi}\left[\tsum\max(f^{M})-r_{t}\right]\right]$$

Now, rewrite the sum as $T$ times an expectation over the uniform distribution, and pull the $T$ and $\max(f^{M})$ out. Then swap the expectations, and expand the expectation over histories as an expectation over partial histories up to $a_{t}$, and an expectation of what $r_{t}$ will be.
$$=T\cdot\expec_{\mu}\left[\max(f^{M})-\bel{\expec}{\theta^{M}\bowtie\pi}[\bel{\expec}{t\sim u(T)}[r_{t}]]\right]=T\cdot\expec_{\mu}\left[\max(f^{M})-\expec_{u(T)}\left[\bel{\expec}{\theta^{M}\bowtie\pi}\left[\bel{\expec}{\theta^{M}(a_{t})}[r]\right]\right]\right]$$

Rewrite the inner expectation as $f^{\theta^{M}}(a)$, and interchange expectations again.
$$=T\cdot\expec_{\mu}\left[\max(f^{M})-\expec_{u(T)}\left[\bel{\expec}{\theta^{M}\bowtie\pi}\left[f^{\theta^{M}}(a_{t})\right]\right]\right]=T\cdot\expec_{\mu}\left[\max(f^{M})-\bel{\expec}{\theta^{M}\bowtie\pi}\left[\expec_{u(T)}\left[f^{\theta^{M}}(a_{t})\right]\right]\right]$$

Note that $\theta^{M}$ doesn't depend on the history, just the action, and the action was generated by sampling a random history and random timestep, which can be viewed as sampling from $p^{M}$.
$$=T\cdot\expec_{\mu}\left[\max(f^{M})-\bel{\expec}{a\sim p^{M}}\left[f^{\theta^{M}}(a)\right]\right]$$
$$=T\left(\bel{\expec}{\mu,\overline{p}}\left[\max(f^{M})-f^{\gen}\right]+\bel{\expec}{\mu,\overline{p}}\left[f^{\gen}-f^{\theta^{M}}\right]+\expec_{\mu}\left[\expec_{\overline{p}}\left[f^{\theta^{M}}\right]-\expec_{p^{M}}\left[f^{\theta^{M}}\right]\right]\right)$$

This can be lower-bounded by the first term, minus the absolute value of the second two terms. Then move the absolute value in.
$$\geq T\left(\bel{\expec}{\mu,\overline{p}}\left[\max(f^{M})-f^{\gen}\right]-\left|\bel{\expec}{\mu,\overline{p}}\left[f^{\gen}-f^{\theta^{M}}\right]\right|-\left|\expec_{\mu}\left[\expec_{\overline{p}}\left[f^{\theta^{M}}\right]-\expec_{p^{M}}\left[f^{\theta^{M}}\right]\right]\right|\right)$$
$$\geq T\left(\bel{\expec}{\mu,\overline{p}}\left[\max(f^{M})-f^{\gen}\right]-\bel{\expec}{\mu,\overline{p}}\left[\left|f^{\gen}-f^{\theta^{M}}\right|\right]-\expec_{\mu}\left[\left|\expec_{\overline{p}}\left[f^{\theta^{M}}\right]-\expec_{p^{M}}\left[f^{\theta^{M}}\right]\right|\right]\right)$$

These terms can be bounded by the total variation distance between $\gen(a)$ and $\theta^{M}(a)$, and the total variation distance between $\overline{p}$ and $p^{M}$, because the expected rewards are in $[0,1]$.
$$\geq T\left(\bel{\expec}{\mu,\overline{p}}\left[\max(f^{M})-f^{\gen}\right]-\bel{\expec}{\mu,\overline{p}}\left[D_{TV}(\gen(a),\theta^{M}(a))\right]-\expec_{\mu}\left[D_{TV}(\overline{p},p^{M})\right]\right)$$

By the data processing inequality, $D_{TV}(\overline{p},p^{M})$ can be upper-bounded by the total variation distance between $\gen\bowtie\pi$ and $\theta^{M}\bowtie\pi$. Then use that total variation distance is upper-bounded by $\sqrt{2}$ times the Hellinger distance.
$$\geq T\left(\bel{\expec}{\mu,\overline{p}}\left[\max(f^{M})-f^{\gen}\right]-\bel{\expec}{\mu,\overline{p}}\left[D_{TV}(\gen(a),\theta^{M}(a))\right]-\expec_{\mu}\left[D_{TV}(\gen\bowtie\pi,\theta^{M}\bowtie\pi)\right]\right)$$
$$\geq T\left(\bel{\expec}{\mu,\overline{p}}\left[\max(f^{M})-f^{\gen}\right]-\bel{\expec}{\mu,\overline{p}}\left[\sqrt{2\hells(\gen(a),\theta^{M}(a))}\right]-\expec_{\mu}\left[\sqrt{2\hells(\gen\bowtie\pi,\theta^{M}\bowtie\pi)}\right]\right)$$

By Lemma A.13 of \cite{Foster21}, that latter Hellinger-squared term can be upper bounded by $100T\ln(T)$ times the expected (under $\overline{p}$) Hellinger distance between $\gen(a)$ and $\theta^{M}(a)$. Then apply concavity of square root.
$$\geq T(\bel{\expec}{\mu,\overline{p}}\left[\max(f^{M})-f^{\gen}\right]-\bel{\expec}{\mu,\overline{p}}\left[\sqrt{2\hells(\gen(a),\theta^{M}(a))}\right]$$
$$-\expec_{\mu}\left[\sqrt{200T\ln(T)\expec_{\overline{p}}\left[\hells(\gen(a),\theta^{M}(a))\right]}\right])$$
$$\geq T(\bel{\expec}{\mu,\overline{p}}\left[\max(f^{M})-f^{\gen}\right]-\sqrt{2\bel{\expec}{\mu,\overline{p}}\left[\hells(\gen(a),\theta^{M}(a))\right]}$$
$$-\sqrt{200T\ln(T)\bel{\expec}{\mu,\overline{p}}\left[\hells(\gen(a),\theta^{M}(a))\right]})$$

Rewrite the expectation over $\mu$ (a distribution) as an expectation over $\mu'$ times $\frac{1}{\lambda}$, then use that $\theta^{M}(a)$ was picked to minimize Hellinger distance to $\gen(a)$, so 
\\
$\hells(\gen(a),\theta^{M}(a))=\hells(\gen(a)\to M(a))$.
$$\geq T(\frac{1}{\lambda}\bel{\expec}{\mu',\overline{p}}\left[\max(f^{M})-f^{\gen}\right]-\sqrt{\frac{2}{\lambda}\bel{\expec}{\mu',\overline{p}}\left[\hells(\gen(a)\to M(a))\right]}$$
$$-\sqrt{\frac{200T\ln(T)}{\lambda}\bel{\expec}{\mu',\overline{p}}\left[\hells(\gen(a)\to M(a))\right]})$$

By how $\gen$ and $\mu'$ were constructed, the expectation of the regret gap exceeds $\text{dec}^{f}_{\eps_{T}^{\frac{1}{1-p}}}(\hypo)$ and the Hellinger expectation is less than $\eps_{T}^{\frac{2}{1-p}}$. Then use that $\eps_{T}=(T\ln(T))^{-1/2}$, so $T\ln(T)=\eps_{T}^{-2}$.
$$\geq T\left(\frac{1}{\lambda}\text{dec}^{f}_{\eps_{T}^{\frac{1}{1-p}}}(\hypo)-\sqrt{\frac{2}{\lambda}\eps_{T}^{\frac{2}{1-p}}}-\sqrt{\frac{200T\ln(T)}{\lambda}\eps_{T}^{\frac{2}{1-p}}}\right)$$
$$=T\left(\frac{1}{\lambda}\text{dec}^{f}_{\eps_{T}^{\frac{1}{1-p}}}(\hypo)-\sqrt{\frac{2}{\lambda}\eps_{T}^{\frac{2}{1-p}}}-\sqrt{\frac{200}{\lambda}\eps_{T}^{\frac{2}{1-p}-2}}\right)$$

At this point, the value of $\lambda$ is the only quantity remaining which depends on the choice of policy $\pi$. This function is convex in $\lambda$, so we will compute the minimizing value of $\lambda$, which is
$$\left(\frac{2\text{dec}^{f}_{\eps_{T}^{\frac{1}{1-p}}}(\hypo)}{\eps_{T}^{\frac{1}{1-p}}\left(\sqrt{2}+\sqrt{200}\frac{1}{\eps_{T}}\right)}\right)^{2}$$

If this minimizer exceeds 1, then because $\lambda\in(0,1]$, the true minimizer will be 1. We will now show that, for all sufficiently large $T$, this occurs. The minimizing $\lambda$ is
$$\geq\left(\frac{2\text{dec}^{f}_{\eps_{T}^{\frac{1}{1-p}}}(\hypo)}{\eps_{T}^{\frac{1}{1-p}}\left(2\sqrt{200}\frac{1}{\eps_{T}}\right)}\right)^{2}=\left(\frac{\text{dec}^{f}_{\eps_{T}^{\frac{1}{1-p}}}(\hypo)}{\sqrt{200}\eps_{T}^{\frac{1}{1-p}-1}}\right)^{2}=\left(\frac{\text{dec}^{f}_{\eps_{T}^{\frac{1}{1-p}}}(\hypo)}{\sqrt{200}\left(\eps_{T}^{\frac{1}{1-p}}\right)^{p}}\right)^{2}$$

Since we assumed that $\liminf_{\eps\to 0}\frac{dec^{f}_{\eps}(\hypo)}{\eps^{p}}=\infty$, the above quantity diverges to infinity. Therefore, for sufficiently large $T$ (which doesn't depend on the choice of $\pi$), the minimizing $\lambda$ is 1. So we continue to lower bound by
$$\geq T\left(\text{dec}^{f}_{\eps_{T}^{\frac{1}{1-p}}}(\hypo)-\sqrt{2\eps_{T}^{\frac{2}{1-p}}}-\sqrt{200\eps_{T}^{\frac{2}{1-p}-2}}\right)=T\left(\text{dec}^{f}_{\eps_{T}^{\frac{1}{1-p}}}(\hypo)-\sqrt{2}\eps_{T}^{\frac{1}{1-p}}-\sqrt{200}\eps_{T}^{\frac{1}{1-p}-1}\right)$$

Using that $\eps_{T}<1$, and $\frac{1}{1-p}-1=\frac{p}{1-p}$, we can proceed to
$$\geq T\left(\text{dec}^{f}_{\eps_{T}^{\frac{1}{1-p}}}(\hypo)-2\sqrt{200}\eps_{T}^{\frac{1}{1-p}-1}\right)= T\left(\text{dec}^{f}_{\eps_{T}^{\frac{1}{1-p}}}(\hypo)-2\sqrt{200}\left(\eps_{T}^{\frac{1}{1-p}}\right)^{p}\right)$$

Putting our inequalities together, given a $p$ where $\liminf_{\eps\to 0}\frac{\text{dec}^{f}_{\eps}(\hypo)}{\eps^{p}}=\infty$, for all $T$ above some finite threshold, every policy $\pi$ has the property that
$$\bel{\max}{M\in\hypo}(\textbf{REG}(\pi,M,T))\geq T\left(\text{dec}^{f}_{\eps_{T}^{\frac{1}{1-p}}}(\hypo)-2\sqrt{200}\left(\eps_{T}^{\frac{1}{1-p}}\right)^{p}\right)$$

As $T$ rises, $\eps^{\frac{1}{1-p}}_{T}$ shrinks. Applying our assumption on $p$, in the limit, the DEC term far exceeds the $2\sqrt{200}$ term, so we have $\min_{\pi}\max_{M\in\hypo}(\textbf{REG}(\pi,M,T))\in\Omega(T\text{dec}^{f}_{\eps^{\frac{1}{1-p}}_{T}}(\hypo))$ as desired. $\blacksquare$

\begin{coro} \label{cor2}
If, for all $p>0$, we have $\liminf_{\eps\to 0}\frac{\text{dec}^{f}_{\eps}(\hypo)}{\eps^{p}}=\infty$, then, for all $q<1$ and $\pi$ (which may depend on $T$), we have $\liminf_{T\to\infty}\frac{\max_{M\in\hypo}(\textbf{REG}(\pi,M,T))}{T^{q}}=\infty$
\end{coro}

Let $q<1$. This implies $\frac{2-2q}{3-2q}\in(0,1)$. Accordingly, let $p$ be an arbitrary positive number strictly less than $\frac{2-2q}{3-2q}$. By Proposition 3, we can show
$$\liminf_{T\to\infty}\frac{\max_{M\in\hypo}(\textbf{REG}(\pi,M,T))}{T^{q}}\geq\liminf_{T\to\infty}T^{1-q}\cdot\text{dec}^{f}_{\eps_{T}^{\frac{1}{1-p}}}(\hypo))=\liminf_{T\to\infty}T^{1-q}\eps_{T}^{\frac{p}{1-p}}\frac{\text{dec}^{f}_{\eps_{T}^{\frac{1}{1-p}}}(\hypo)}{\left(\eps_{T}^{\frac{1}{1-p}}\right)^{p}}$$

Apply our assumption that, for every $p$, the liminf of that fraction diverges to infinity, and use that $\eps_{T}$ was defined to be $(T\ln(T))^{-1/2}$, to get
$$\geq\liminf_{T\to\infty}T^{1-q}\eps_{T}^{\frac{p}{1-p}}=\liminf_{T\to\infty}T^{1-q}(T\ln(T))^{\frac{-p}{2(1-p)}}$$

If the exponent on $T$ exceeds 0, then it will outgrow the $\ln(T)$ raised to a negative power, establishing our desired result that this quantity diverges to infinity. We now switch to proving $1-q-\frac{p}{2(1-p)}>0$. By recalling that $p<\frac{2-2q}{3-2q}$, we can compute
$$1-q-\frac{p}{2(1-p)}>1-q-\frac{\frac{2-2q}{3-2q}}{2\left(1-\frac{2-2q}{3-2q}\right)}=1-q-\frac{2-2q}{2((3-2q)-(2-2q))}=1-q-(1-q)=0$$

And our result follows. $\blacksquare$
\section{Hellinger Distance Lemmas}
\begin{lemm} \label{lem1}
Let $X$ be a nonempty compact convex Polish space, and $f:X\to\mathbb{R}$ be continuous and strictly convex. Then there is a unique minimizer of $f$.
\end{lemm}

$f$ is continuous, and, because $X$ is a compact space, a minimizer exists. To prove that the minimizer is unique, assume there are two distinct minimizers, $x,x'$. In such a case, by the strict convexity of $f$, we'd have 
$$f(0.5x+0.5x')<0.5f(x)+0.5f(x')=\min_{x}f(x)$$

By convexity of $X$, $0.5x+0.5x'$ is also in $X$, but this choice of value makes $f$ lower than its minimum value, and we have a contradiction. Therefore, the minimizer must be unique. $\blacksquare$

\begin{lemm} \label{lem2}
Let $X$ be a nonempty compact convex Polish space, and $Y$ be a Polish space, and $f:X\times Y\to\mathbb{R}$ be a function which is continuous in $X\times Y$, and strictly convex in $X$ for all $y$. Then the function $\lambda y.\text{argmin}_{x\in X}f(x,y)$ is continuous.
\end{lemm}

Use $x_{y}$ as an abbreviation for $\text{argmin}_{x\in X}f(x,y)$. This denotes a unique point, by Lemma \ref{lem1}. Fix an arbitrary sequence $y_{n}$ converging to $y_{\infty}$ in $Y$. By compactness of $X$, the sequence $x_{y_{n}}$ has at least one limit point. Fix an arbitrary limit point $x_{\infty}$, and a subsequence where the $x_{y_{n}}$ converge to $x_{\infty}$. Then by continuity of $f$, the fact that $x_{y_{n}}$ is the minimizer for $y_{n}$, continuity of $f$, and the fact that $x_{y_{\infty}}$ is the minimizer for $y_{\infty}$, we have
$$f(x_{y_{\infty}},y_{\infty})=\lim_{n\to\infty}f(x_{y_{\infty}},y_{n})\geq\lim_{n\to\infty}f(x_{y_{n}},y_{n})=f(x_{\infty},y_{\infty})\geq f(x_{y_{\infty}},y_{\infty})$$

The left and right sides are equal, so all inequalities are equalities, and we have $f(x_{\infty},y_{\infty})=f(x_{y_{\infty}},y_{\infty})$. However, by Lemma 1, minimizers are unique, so $x_{\infty}=x_{y_{\infty}}$. $x_{\infty}$ was an arbitrary limit point of the $x_{y_{n}}$, so all limit points are equal, so $x_{y_{n}}$ converges to $x_{y_{\infty}}$. Our convergent sequence $y_{n}$ was arbitrary, so the function $\lambda y.x_{y}$ is continuous. Unpacking the definition of $x_{y}$, this shows continuity of $\lambda y.\text{argmin}_{x\in X}f(x,y)$, as desired. $\blacksquare$

\begin{lemm} \label{lem3}
Let $X$ be a nonempty compact convex Polish space, and $Y$ be a convex open subset of a normed vector space $A$, and $f:X\times Y\to\mathbb{R}$ be a function which is continuous in $X\times Y$, convex in $Y$ for all $x$, strictly convex in $X$ for all $y$, Frechet-differentiable in $Y$ for all $x$, and (letting $df^{x}:Y\to A^{*}$ be the Frechet derivative in $Y$ at $x$), has $\lambda x.df^{x}_{y}:X\to A^{*}$ being continuous for all $y$. In such a case, letting $g(y):=\min_{x\in X}f(x,y)$, $g$ is also Frechet-differentiable, and $dg_{y}=df^{\text{argmin}_{x\in X}f(x,y)}_{y}$.
\end{lemm}

Use the notation $x_{y}$ as an abbreviation for $\text{argmin}_{x\in X}f(x,y)$. By Lemma \ref{lem1}, this denotes a unique point. To show Frechet-differentiability of $g$, and that the Frechet differential is as desired, we must show that
$$\lim_{h\to 0}\frac{g(y+h)-g(y)-df^{x_{y}}_{y}(h)}{||h||}=0$$

The $h$ are vectors in the space $A$. We will prove this by showing that the limsup is below 0 and the liminf is above 0. For the upper bound, we compute
$$\limsup_{h\to 0}\frac{g(y+h)-g(y)-df^{x_{y}}_{y}(h)}{||h||}=\limsup_{h\to 0}\frac{f(x_{y+h},y+h)-f(x_{y},y)-df^{x_{y}}_{y}(h)}{||h||}$$
$$\leq\limsup_{h\to 0}\frac{f(x_{y},y+h)-f(x_{y},y)-df^{x_{y}}_{y}(h)}{||h||}=0$$

In order, this was by expanding the definition of $g$, using that $f(x_{y+h},y+h)\leq f(x_{y},y+h)$, and using that $df^{x_{y}}_{y}$ being the Frechet differential of $f$ at $x_{y}$ and $y$ implies the relevant limit is zero, by the definition of the Frechet derivative. For the lower bound, we have
$$\liminf_{h\to 0}\frac{g(y+h)-g(y)-df^{x_{y}}_{y}(h)}{||h||}=\liminf_{h\to 0}\frac{f(x_{y+h},y+h)-f(x_{y},y)-df^{x_{y}}_{y}(h)}{||h||}$$
$$\geq\liminf_{h\to 0}\frac{f(x_{y+h},y+h)-f(x_{y+h},y)-df^{x_{y}}_{y}(h)}{||h||}$$
$$\geq\liminf_{h\to 0}\frac{f(x_{y+h},y)+df^{x_{y+h}}_{y}(h)-f(x_{y+h},y)-df^{x_{y}}_{y}(h)}{||h||}$$
$$=\liminf_{h\to 0}\frac{(df^{x_{y+h}}_{y}-df^{x_{y}}_{y})(h)}{||h||}\geq\liminf_{h\to 0}\frac{-||df^{x_{y+h}}_{y}-df^{x_{y}}_{y}||\cdot||h||}{||h||}$$
$$=\liminf_{h\to 0}-||df^{x_{y+h}}_{y}-df^{x_{y}}_{y}||=0$$

In order, this was expanding definitions, and using that $f(x_{y},y)\leq f(x_{y+h},y)$. To derive the third line, we used the convexity of $f$ for all $x$, specifically the part where the linear approximation to a convex function at a point always undershoots the original convex function. Then we cancel, use that the differentials are linear functions, lower-bound with the operator norm, and cancel again. The final step uses Lemma \ref{lem2}, that $x_{y}$ is continuous in $y$, to show that $x_{y+h}$ converges to $x$. Combining this with the starting assumption that $\lambda x.df^{x}_{y}:X\to A^{*}$ is continuous for all $y$, $df^{x_{y+h}}_{y}$ converges to $df^{x_{y}}_{y}$, so the operator norm of their difference shrinks to zero.

This establishes that the Frechet derivative of $g$ exists and equals $df^{x_{y}}_{y}$ at $y$. The lemma then follows because $x_{y}=\text{argmin}_{x\in X}f(x,y)$. $\blacksquare$

\begin{lemm} \label{lem4}
Given a finite space of observations $\obs$, a $\nu\in\Delta\obs$ with full support, and an imprecise belief $\Psi$, $\hells(\nu\to\Psi)$ is convex and Frechet-differentiable in $\nu$. Letting $\mu_{\nu}:=\text{argmin}_{\mu\in\Psi}\hells(\mu,\nu)$, we have that $d(\lambda\nu'.\hells(\nu'\to\Psi))_{\nu}=d(\lambda\nu'.\hells(\mu_{\nu},\nu'))_{\nu}$.
\end{lemm}

We apply Lemma \ref{lem3}. To verify the conditions, the compact convex Polish space is $\Psi$, because imprecise beliefs are nonempty compact convex subsets of $\Delta\obs$. Letting $u$ denote the uniform measure over $\obs$, then $Y$ would be $\Set{\nu-u}{\nu\in\Delta\obs,\nu\text{ has full support}}$. This is the convex open set of probability distributions with full support, but moved so that it's in a subspace of $\mathbb{R}^{\obs}$. For Lemma \ref{lem3}, $f$ would be $f(\mu,\nu-u)=\hells(\mu,\nu)$, and $\hells(\mu,\nu)=1-\sum_{o}\sqrt{\mu(o)\cdot\nu(o)}$. This is clearly seen to be continuous in both arguments, convex in its second argument (by concavity of square root), and strictly convex in its first argument (by strict concavity of square root, and $\nu(o)$ being nonzero for all observations). Now we only need to verify the Frechet-differentiability conditions on Hellinger distance to invoke Lemma \ref{lem3}. Computing the Frechet derivative for some fixed $\mu$, and letting $h$ be in $\mathbb{R}^{\obs}$, we get
$$d(\lambda\nu-u.f(\mu,\nu-u))(h)=d\left(\lambda\nu.1-\sum_{o}\sqrt{\mu(o)\cdot\nu(o)}\right)(h)$$
$$=\sum_{o}\frac{-1}{2\sqrt{\nu(o)}}\sqrt{\mu(o)}d(\lambda\nu.\nu(o))(h)=\sum_{o}\frac{-1}{2}\sqrt{\frac{\mu(o)}{\nu(o)}}h(o)=\left\langle-\frac{1}{2}\sqrt{\frac{\mu}{\nu}},h\right\rangle$$

Because this derivative can be written as an inner product with a vector which is well-defined in all entries (because all $\nu$ are assumed to have full support), $f$ is Frechet-differentiable in $Y$ for all $\mu$. Further, for all $\nu$ with full support, continuity in $\mu$ holds. All conditions have been verified, so we can apply Lemma \ref{lem3} to show
$$d(\lambda\nu'.\hells(\nu'\to\Psi))_{\nu}=d(\lambda\nu'.\min_{\mu\in\Psi}\hells(\mu,\nu'))_{\nu}=d(\lambda\mu,\nu'.\hells(\mu,\nu'))^{\mu_{\nu}}_{\nu}=d(\lambda\nu'.\hells(\mu_{\nu},\nu'))_{\nu}$$

This establishes Frechet-differentiability, but not convexity. To show convexity, we compute
$$\hells(p\nu+(1-p)\nu'\to\Psi)=\bel{\min}{\mu\in\Psi}\hells(p\nu+(1-p)\nu',\mu)$$

Using $\mu_{\nu},\mu_{\nu'}$ for the Hellinger minimizers, we then use that $\Psi$ is convex by the definition of an imprecise belief, so $p\mu_{\nu}+(1-p)\mu_{\nu'}\in\Psi$, then apply convexity of Hellinger-squared error in both arguments.
$$\leq\hells(p\nu+(1-p)\nu',p\mu_{\nu}+(1-p)\mu_{\nu'})\leq p\hells(\nu,\mu_{\nu})+(1-p)\hells(\nu,\mu_{\nu'})$$
$$=p\hells(\nu\to\Psi)+(1-p)\hells(\nu'\to\Psi)$$

And convexity is shown. $\blacksquare$

\begin{lemm} \label{lem5}
For imprecise beliefs $\Psi,\Phi,\Theta$, $\hells(\Psi\to\Theta)\leq 2\hells(\Psi\to\Phi)+2\hells(\Phi\to\Theta)$
\end{lemm}

First, we establish the analogous fact for probability distributions, that 
\\
$\hells(\mu,\xi)\leq 2\hells(\mu,\nu)+2\hells(\nu,\xi)$. Applying the triangle inequality, squaring both sides, and using the AM/GM inequality yields the result.
$$\hells(\mu,\xi)\leq(D_{H}(\mu,\nu)+D_{H}(\nu,\xi))^{2}=\hells(\mu,\nu)+\hells(\nu,\xi)+2D_{H}(\mu,\nu)D_{H}(\nu,\xi)$$
$$=\hells(\mu,\nu)+\hells(\nu,\xi)+2\sqrt{\hells(\mu,\nu)\hells(\nu,\xi)}\leq2\hells(\mu,\nu)+2\hells(\nu,\xi)$$

Letting $\nu_{\mu}$ be the point in $\Phi$ which is closest to $\mu$, we can use definitions and our probabilistic result above to yield
$$\hells(\Psi\to\Theta)=\bel{\max}{\mu\in\Psi}\bel{\min}{\zeta\in\Theta}\hells(\mu,\zeta)\leq\bel{\max}{\mu\in\Psi}\bel{\min}{\zeta\in\Theta}(2\hells(\mu,\nu_{\mu})+2\hells(\nu_{\mu},\zeta))$$
$$=2\bel{\max}{\mu\in\Psi}\hells(\mu,\nu_{\mu})+2\bel{\min}{\zeta\in\Theta}\hells(\nu_{\mu},\zeta)\leq 2\bel{\max}{\mu\in\Psi}\hells(\mu,\nu_{\mu})+2\bel{\max}{\nu'\in\Phi}\bel{\min}{\zeta\in\Theta}\hells(\nu',\zeta)$$
$$=2\bel{\max}{\mu\in\Psi}\bel{\min}{\nu\in\Phi}\hells(\mu,\nu)+2\bel{\max}{\nu'\in\Phi}\bel{\min}{\zeta\in\Theta}\hells(\nu',\zeta))=2\hells(\Psi\to\Phi)+2\hells(\Phi\to\Theta)$$
$\blacksquare$
\section{Robust Online Estimation}
\begin{lemm} \label{lem6}
Given a finite set $\mathcal{B}$, a $T:\mathbb{N}^{>0}$, a history $h:(\acts\times\obs)^{\leq T}$, a sequence $\guess:\acts\to\Delta\obs$, a distribution $\zeta_{1}:\Delta\mathcal{B}$ with full support, and a betting function $\text{bet}:\mathcal{B}\times(\acts\to\Delta\obs)\times\acts\times\obs\to\mathbb{R}^{\geq 0}$, we can recursively define
$$\bigstar_{t}:=\expec_{B\sim\zeta_{t}}[\text{bet}(B,\guess,a_t,o_t)]$$
$$\zeta_{t+1}(B):=\frac{\zeta_{t}(B)\cdot\text{bet}(B,\guess,a_t,o_t)}{\bigstar_{t}}$$
If all $\bigstar_{t}>0$, then every $B\in\mathcal{B}$ fulfills
$$\ln(\zeta_{T+1}(B))-\ln(\zeta_{1}(B))=\tsum\ln(\text{bet}(B,\guess,a_{t},o_{t}))-\tsum\ln(\bigstar_{t})$$
\end{lemm}

The proof is as follows.
$$\ln(\zeta_{T+1}(B))-\ln(\zeta_{1}(B))=\tsum\ln(\zeta_{t+1}(B))-\ln(\zeta_{t}(B))=\tsum\ln\left(\frac{\zeta_{t+1}(B)}{\zeta_{t}(B)}\right)$$
$$=\tsum\ln\left(\frac{\frac{\zeta_{t}(B)\cdot\text{bet}(B,\guess,a_{t},o_{t})}{\bigstar_{t}}}{\zeta_{t}(B)}\right)=\tsum\ln\left(\frac{\text{bet}(B,\guess,a_{t},o_{t})}{\bigstar_{t}}\right)$$
$$=\tsum\ln(\text{bet}(B,\guess,a_{t},o_{t}))-\tsum\ln(\bigstar_{t})$$

This proof above only works if all $\zeta_{t}$ assign nonzero probability to $B$. If some $\zeta_{t}$ assigns zero probability to $B$, then the same holds for all greater $t$, and this can only have occurred by one of the bets returning 0. So, as long as we still have $\bigstar_{t}>0$ for all $t$, we can verify our desired equation via
$$\ln(\zeta_{T+1}(B))-\ln(\zeta_{1}(B))=-\infty=\tsum\ln(\text{bet}(B,\guess,a_{t},o_{t}))-\tsum\ln(\bigstar_{t})$$

This is because $\zeta_{T+1}(B)=0$, $\zeta_{1}(B)>0$, some $t$ has $\text{bet}(B,\guess,a_{t},o_{t})=0$, and all $\bigstar_{t}>0$. $\blacksquare$

\begin{prop} \label{pro4}
If the uniform bettor has positive probability according to $\zeta_{1}$, the RUE algorithm never divides by zero and all $\zeta_{t}$ are probability distributions.
\end{prop}

This proof proceeds in three phases. First, we show that if the uniform bettor $B_{\text{uniform}}$ has nonzero probability according to some $\zeta_{t}$ which is a probability distribution, then for all $a$, $\guess(a)$ will have full support, eliminating division-by-zero errors if the precondition holds. Second, we show that if the uniform bettor has nonzero probability according to some $\zeta_{t}$ which is a probability distribution, then in the language of Lemma \ref{lem6}, for all $a,o$, $\expec_{B\sim\zeta_{t}}[\text{bet}(B,\guess,a,o)]=1$ holds. Finally, with these two results, we carry through an inductive proof that the uniform bettor always has nonzero probability and all $\zeta_{t}$ are probability distributions. Therefore, $\guess(a)$ always has full support, eliminating all division-by-zero errors by the first phase of the proof.

Let $\eps'$ be $\zeta_{1}(B_{\text{uniform}})$. To recap some notation, given an bettor $B$ which isn't the pessimism bettor, $M_{B}$ denotes the model in $\hypo$ which $B$ corresponds to, or the function mapping all actions to the uniform distribution over $\obs$ for the uniform bettor. We use $\mu_{\guess,B,a}$ for $\bel{\text{argmin}}{\mu\in M_{B}(a)}\hells(\mu,\guess(a))$. If $\guess(a)$ has full support, this distribution is unique, but until then, it is an arbitrary distribution.

For the first phase, assume for purposes of contradiction that there is some $a,o'$ such that $\mathbb{P}_{\guess(a)}(o')=0$ and yet the uniform bettor has nonzero probability according to $\zeta_{t}$ which is a probability distribution. Fix an observation $o_{a}$ which $\guess(a)$ assigns nonzero probability to. Restating definitions, we have
$$\guess(a)=\bel{\text{argmin}}{\mu\in\Delta\obs}\bel{\expec}{B\sim\zeta_{t}}[\text{ if }B=\bullet,\eps\bel{\expec}{o\sim\mu}[r(a,o)]\text{ else }2\hells(\mu\to M_{B}(a))]$$
Perturb the minimizing $\guess(a)$ to $\guess'(a)$ by moving an arbitrarily small $\eps''$ amount of probability measure from $o_{a}$ to the (zero probability) observation $o'$. Now we compute.
$$\bel{\expec}{B\sim\zeta_{t}}[\text{ if }B=\bullet,\eps\bel{\expec}{o\sim\guess(a)}[r(a,o)]\text{ else }2\hells(\guess(a)\to M_{B}(a))]$$
$$-\bel{\expec}{B\sim\zeta_{t}}[\text{ if }B=\bullet,\eps\bel{\expec}{o\sim\guess'(a)}[r(a,o)]\text{ else }2\hells(\guess'(a)\to M_{B}(a))]$$
$$=\bel{\expec}{B\sim\zeta_{t}}[\text{ if }\bullet,\eps(\bel{\expec}{o\sim\guess(a)}[r(a,o)]-\bel{\expec}{o\sim\guess'(a)}[r(a,o)])$$
$$\text{ else }2(\hells(\guess(a)\to M_{B}(a))-\hells(\guess'(a)\to M_{B}(a)))]$$
$$\geq\bel{\expec}{B\sim\zeta_{t}}\left[\text{if }\bullet,-\eps\cdot D_{TV}(\guess(a),\guess'(a))\text{ else }2\left(\hells(\guess(a),\mu_{\guess,B,a})-\hells(\guess'(a),\mu_{\guess,B,a})\right)\right]$$

The above line was because the reward was in $[0,1]$, so we can lower-bound by the negative total variation distance. For the Hellinger-squared error term, we use that $\mu_{\guess,B,a}$ was defined as the error-minimizer, and swapping "distance from a point to a set" for "distance from a point to a point" can only increase the distance, which leads to a decrease because of the negative sign. Now, split out the uniform bettor from the rest $\hypo$, use that $\mu_{\guess,B_{\text{uniform}},a}$ is the uniform distribution (denoted as $u$), decrease further by minimizing over all $\mu$, and write $\min_{\mu}(f(\mu))$ as $-\max_{\mu}(-f(\mu))$.
$$\geq\bel{\expec}{B\sim\zeta_{t}}[\text{if }\bullet,-\eps\cdot D_{TV}(\guess(a),\guess'(a))\text{ else if }B_{\text{uniform}},$$
$$-2\left(\hells(\guess'(a),u)-\hells(\guess(a),u)\right)\text{ else }-2\bel{\max}{\mu}\left(\hells(\guess'(a),\mu)-\hells(\guess(a),\mu)\right)]$$

Unpacking the Hellinger-squared error as $1-\sum_{o}\sqrt{\guess(a)(o)\cdot\mu(o)}$, and canceling out as much as we can because $\guess(a)$ and $\guess'(a)$ are the same except on two observations, we get
$$=\bel{\expec}{B\sim\zeta_{t}}[\text{ if }\bullet,-\eps\cdot D_{TV}(\guess(a),\guess'(a))\text{ else if }B_{\text{uniform}},$$
$$-2\left(\sqrt{\guess(a)(o')u(o')}+\sqrt{\guess(a)(o_{a})u(o_{a})}-\sqrt{\guess'(a)(o')u(o')}-\sqrt{\guess'(a)(o_{a})u(o_{a})}\right)\text{ else }$$
$$-2\bel{\max}{\mu}\left(\sqrt{\guess(a)(o') \mu(o')}+\sqrt{\guess(a)(o_{a})\mu(o_{a})}-\sqrt{\guess'(a)(o')\mu(o')}-\sqrt{\guess'(a)(o_{a})\mu(o_{a})}\right)]$$

We now use that $\guess(a)(o')$ is zero, $\guess'(a)(o')$ is $\eps''$, and $\guess'(a)(o_{a})=\guess(a)(o_{a})-\eps''$.
$$=\bel{\expec}{B\sim\zeta_{t}}[\text{if }\bullet,-\eps\cdot D_{TV}(\guess(a),\guess'(a))$$
$$\text{ else if }B_{\text{uniform}},-2\left(\sqrt{u(o_{a})}\left(\sqrt{\guess(a)(o_{a})}-\sqrt{\guess(a)(o_{a})-\eps''}\right)-\sqrt{\eps''\cdot u(o')}\right)$$
$$\text{ else }-2\bel{\max}{\mu}\left(\sqrt{\mu(o_{a})}\left(\sqrt{\guess(a)(o_{a})}-\sqrt{\guess(a)(o_{a})-\eps''}\right)-\sqrt{\eps''\cdot\mu(o')}\right)]$$

Remove some positive terms, and pull one of them out of the expectation.
$$\geq\bel{\expec}{B\sim\zeta_{t}}[\text{if }\bullet,-\eps\cdot D_{TV}(\guess(a),\guess'(a))$$
$$\text{ else if }B_{\text{uniform}},-2\sqrt{u(o_{a})}\left(\sqrt{\guess(a)(o_{a})}-\sqrt{\guess(a)(o_{a})-\eps''}\right)$$
$$\text{ else }-2\bel{\max}{\mu}\sqrt{\mu(o_{a})}\left(\sqrt{\guess(a)(o_{a})}-\sqrt{\guess(a)(o_{a})-\eps''}\right)]+2\zeta_{t}(B_{\text{uniform}})\sqrt{\eps''\cdot u(o')}$$

Then upper-bound $\mu(o_{a})$ and $u(o_{a})$ by 1, letting us fold the uniform case, and the $B\in\hypo$ case together.
$$\geq\bel{\expec}{B\sim\zeta_{t}}\left[\text{ if }\bullet,-\eps\cdot D_{TV}(\guess(a),\guess'(a))\text{ else}-2\left(\sqrt{\guess(a)(o_{a})}-\sqrt{\guess(a)(o_{a})-\eps''}\right)\right]$$
$$+2\zeta_{t}(B_{\text{uniform}})\sqrt{\eps''\cdot u(o')}$$

Upper-bound $\eps$ by 1, and use that the total variation distance between $\guess(a)$ and $\guess'(a)$ is $\eps''$. Further, $\guess(a)(o_{a})$ was stipulated to be $>0$, so we can take a limit as $\eps''$ approaches zero, to yield, to within $\mathcal{O}\left(\eps''^{2}\right)$,
$$\geq\bel{\expec}{B\sim\zeta_{t}}\left[\text{if }\bullet,-\eps'',\text{ else}-\frac{1}{\sqrt{\guess(a)(o_{a})}}\eps''\right]+2\zeta_{t}(B_{\text{uniform}})\sqrt{\eps''\cdot u(o')}$$
$$\geq-\frac{1}{\sqrt{\guess(a)(o_{a})}}\eps'+2\zeta_{t}(B_{\text{uniform}})\sqrt{\eps'\cdot u(o')}$$

However, $\guess(a)(o_{a})$ was assumed to be positive, as was $\zeta_{t}(B_{\text{uniform}})$, and the outcome space $\obs$ was assumed to be finite, so this is a term on the order of $\sqrt{\eps''}$ minus a term on the order of $\eps''$, which is positive when $\eps''$ is sufficiently small. So our net inequality, for sufficiently small $\eps''$, is
$$\bel{\expec}{B\sim\zeta_{t}}[\text{ if }B=\bullet,\eps\bel{\expec}{o\sim\guess(a)}[r(a,o)]\text{ else }2\hells(\guess(a)\to M_{B}(a))]$$
$$-\bel{\expec}{B\sim\zeta_{t}}[\text{if }B=\bullet,\eps\bel{\expec}{o\sim\guess'(a)}[r(a,o)]\text{ else }2\hells(\guess'(a)\to M_{B}(a))]>0$$

However, this is impossible, because $\guess(a)$ was assumed to be the minimizer of the equation, so perturbing it cannot produce a strictly lower value. By contradiction, if the uniform bettor $B_{\text{uniform}}$ has nonzero probability according to some $\zeta_{t}$ which is a probability distribution, then for all $a$, $\guess(a)$ has full support.
\\

Now we move on to phase two, and show that if the uniform bettor has nonzero probability according to some $\zeta_{t}$ which is a probability distribution (implying that $\guess(a)$ always has full support), then for \emph{every} $a,o$, we have $\bel{\expec}{B\sim\zeta_{t}}[\text{bet}(B,\guess,a,o)]=1$.
Restating the estimate of RUE, it is
$$\guess(a)=\bel{\text{argmin}}{\nu\in\Delta\obs}\bel{\expec}{B\sim\zeta_{t}}[\text{if }B=\bullet,\eps\bel{\expec}{o\sim\nu}[r(a,o)]\text{ else }2\hells(\nu\to M_{B}(a))]$$
To abbreviate this, let $g^{B,a}:\Delta\obs\to\mathbb{R}$ be $\lambda\nu.\eps\bel{\expec}{o\sim\nu}[r(a,o)]$ for $B=\bullet$ (the pessimism bettor), and $\lambda\nu.2\hells(\nu\to M_{B}(a))$ otherwise. This lets us restate the estimate as $\guess(a)=\bel{\text{argmin}}{\nu\in\Delta\obs}\bel{\expec}{B\sim\zeta_{t}}[g^{B,a}]$. Note that, for all $B,a$, $g^{B,a}$ is Frechet-differentiable on the interior of $\Delta\obs$, and convex. This is easy to show for the pessimism bettor, but for Hellinger-squared error, we appeal to Lemma \ref{lem4}, because $\guess(a)$ has full support. $\bel{\expec}{B\sim\zeta_{t}}[g^{B,a}]$ is a finite number of Frechet-differentiable convex functions mixed together, so it is a Frechet-differentiable convex function.

Because $\guess(a)$ is the minimizer of a differentiable convex function, the derivative in all directions is 0. Let $o$ be an arbitrary observation, and consider our direction of movement to be $\guess(a)-\textbf{1}_{o}$, where $\textbf{1}_{o}$ is the distribution which places all measure on $o$. This is a probability distribution minus a probability distribution, and represents moving from certainty in an observation towards our prediction. Use that the derivative in all directions is 0, the derivative of a mixture of functions is the mixture of the derivatives, and linearity. Then unpack the definition of $g^{B,a}$.
$$0=d\left(\lambda\nu.\bel{\expec}{B\sim\zeta_{t}}\left[g^{B,a}(\nu)\right]\right)_{\guess(a)}(\guess(a)-\textbf{1}_{o})=\expec_{\zeta_{t}}\left[d(g^{B,a})_{\guess(a)}(\guess(a)-\textbf{1}_{o})\right]$$
$$=\expec_{\zeta_{t}}[\text{if }B=\bullet,d(\lambda\nu.\eps\bel{\expec}{o\sim\nu}[r(a,o)])_{\guess(a)}(\guess(a)-\textbf{1}_{o})$$
$$\text{ else }d(\lambda\nu.2\hells(\nu\to M_{B}(a)))_{\guess(a)}(\guess(a)-\textbf{1}_{o})]$$

Applying Lemma \ref{lem4}, we get
$$=\expec_{\zeta_{t}}[\text{if }B=\bullet,d(\lambda\nu.\eps\bel{\expec}{o\sim\nu}[r(a,o)])_{\guess(a)}(\guess(a)-\textbf{1}_{o})$$
$$\text{ else }d(\lambda\nu.2\hells(\mu_{\guess,B,a},\nu))_{\guess(a)}(\guess(a)-\textbf{1}_{o})]$$

In Lemma \ref{lem4} we computed the differential for Hellinger distance and expressed it as an inner product, yielding
$$d(\lambda\nu.2\hells(\mu_{\guess,B,a},\nu))_{\guess(a)}(\guess(a)-\textbf{1}_{o})=\left\langle-\sqrt{\frac{\mu_{\guess,B,a}}{\guess(a)}},\guess(a)-\textbf{1}_{o}\right\rangle$$

For the other differential, it is easy to compute.
$$d(\lambda\nu.\eps\bel{\expec}{o\sim\nu}[r(a,o)])_{\guess(a)}(\guess(a)-\textbf{1}_{o})=\left\langle\lambda o'.\eps\cdot  r(a,o'),\guess(a)-\textbf{1}_{o}\right\rangle$$

Substituting these in, we have an overall equality of
$$0=\expec_{\zeta_{t}}\left[\text{if }B=\bullet,\left\langle\lambda o'.\eps\cdot r(a,o'),\guess(a)-\textbf{1}_{o}\right\rangle\text{ else }\left\langle-\sqrt{\frac{\mu_{\guess,B,a}}{\guess(a)}},\guess(a)-\textbf{1}_{o}\right\rangle\right]$$

We can rewrite this as
$$0=\expec_{\zeta_{t}}[\text{if }\bullet,\eps\left(\bel{\expec}{o'\sim\guess(a)}[r(a,o')]-r(a,o)\right)$$
$$\text{ else }\sqrt{\frac{\mu_{\guess,B,a}(o)}{\guess(a)(o)}}-\sum_{o'}\sqrt{\mu_{\guess,B,a}(o')\cdot\guess(a)(o')}]$$

Adding 1 to both sides, and using that $\hells(\mu,\nu)=1-\sum_{o}\sqrt{\mu(o)\nu(o)}$, and that
\\
$\hells(\mu_{\guess,B,a},\guess(a))=\hells(\guess(a)\to M_{B}(a))$, we get
$$1=\expec_{\zeta_{t}}[\text{if }B=\bullet,1+\eps\left(\bel{\expec}{o'\sim\guess(a)}[r(a,o')]-r(a,o)\right)$$
$$\text{ else }\sqrt{\frac{\mu_{\guess,B,a}(o)}{\guess(a)(o)}}+\hells(\guess(a)\to M_{B}(a))]$$

$\zeta_{t}(B)$ times the associated bet of $B$ is $\zeta_{t+1}(B)$ according to the RUE algorithm, so the above equation shows that $\zeta_{t+1}$ is a probability distribution because $\sum_{B\in\mathcal{B}}\zeta_{t+1}(B)=1$. In the language of Lemma \ref{lem6}, the above is precisely the definition of the betting functions for RUE, so we get an overall equality of $1=\bel{\expec}{B\sim\zeta_{t}}\left[\text{bet}(B,\guess,a,o)\right]$. This argument works for any $a,o,t$ if $\zeta_{t}$ of the uniform bettor is nonzero. So, also, in the language of Lemma \ref{lem6}, we have just proven that $\zeta_{t}(B_{\text{uniform}})>0$ and $\zeta_{t}$ being a probability distribution implies that $\bigstar_{t}=1$ and $\zeta_{t+1}$ is a probability distribution.
\\

For our final phase, we give an inductive proof that the uniform bettor always has nonzero probability and all $\zeta_{t}$ are probability distributions. In the base case, this holds by assumption. For the induction step, if $\zeta_{t}(B_{\text{uniform}})>0$ and $\zeta_{t}$ is a probability distribution, then $\zeta_{t+1}$ is also a probability distribution, as proved earlier. Then we have $\zeta_{t+1}(B_{\text{uniform}})=\zeta_{t}(B_{\text{uniform}})\text{bet}(B_{\text{uniform}},\guess,a_{t},o_{t})$. The probability of the uniform bettor according to $\zeta_{t}$ is nonzero (by induction assumption), so we just need to show that the bet made is nonzero. We may compute the bet of the uniform bettor (because it always bets on the uniform distribution $u$) as
$$\sqrt{\frac{u(o_{t})}{\guess(a_{t})(o_{t})}}+\hells(\guess(a_{t}),u)$$

Because the observation space is finite, nonzero probability is assigned to all observations, so the square root is always nonzero, establishing $\zeta_{t+1}(B_{\text{uniform}})>0$. By induction, our desired result then follows, that the uniform bettor always has nonzero probability and all $\zeta_{t}$ are probability distributions. All $\bigstar_{t}$ are 1 by the second phase of the proof, and by the first phase of the proof, we can conclude that $\guess(a)$ has full support for all $a,t$, and division-by-zero errors are impossible. $\blacksquare$

\begin{theo} \label{the2}
If our estimator $\estim$ is the RUE algorithm with a suitable choice of prior, 
\\
$\beta_{\estim}(T,\delta)\leq\ln\left(\frac{2|\hypo|}{\delta}\right)$, and $\alpha_{\estim}(T,\delta)\leq\sqrt{T}\left(2\sqrt{\ln(2)}+\sqrt{2\ln\left(\frac{1}{\delta}\right)}\right)$
\end{theo}

The choice of prior is $\frac{1}{2|\hypo|}$ probability assigned to all bettors corresponding to $M\in\hypo$, some arbitrarily low $\eps'$ probability assigned to the uniform bettor, and $\frac{1}{2}-\eps'$ probability assigned to the pessimism bettor. Technically, the bound on $\beta$ will only hold for $\eps'\leq\frac{1}{2}$, and the bound on $\alpha$ only holds for $T>2$ in the $\eps'\to 0$ limit.

We use $\mu_{\guess,B,a}$ to denote $\bel{\text{argmin}}{\mu\in M_{B}(a)}\hells(\mu,\guess(a))$. The RUE algorithm is well-defined by Proposition \ref{pro4}. Now, we can invoke Lemma \ref{lem6} to analyze the estimation complexity.
$$\ln(\zeta_{T+1}(B))-\ln(\zeta_{1}(B))=\tsum\ln(\text{bet}(B,\guess,a_{t},o_{t}))-\tsum\ln(\bigstar_{t})$$

This can be simplified further. All $\bigstar_{t}=1$, as proved in Proposition \ref{pro4}. Our choice of $\zeta_{1}$ was half of the measure uniformly distributed on $\hypo$, and approximately half on the pessimism bettor, and an arbitrarily small quantity on the uniform bettor. Substituting these in, we have shown that for all $B$ which aren't the pessimism or uniform bettor, we have $\ln\left(2|\hypo|\right)\geq\tsum\ln(\text{bet}(B,\guess,a_{t},o_{t}))$, and for the pessimism bettor, the left-hand side is just $\ln(\frac{1}{\frac{1}{2}-\eps'})$, which is approximately $\ln(2)$.
\\

We will now bound $\beta_{\guess}$ and $\alpha_{\guess}$ using martingale arguments. Fix some true hypothesis 
\\
$M^{*}\in\hypo$, environment $\theta\models M^{*}$, and algorithm $\pi$. Let $B^{*}$ denote the bettor corresponding to $M^{*}$. By Lemma A.4 from \cite{Foster21}, with $1-\delta$ probability according to $\theta\bowtie\pi$, we will have
$$\tsum\ln(\text{bet}(B^{*},\guess,a_{t},o_{t}))\geq-\tsum\ln\left(\bel{\expec}{a\sim\pi_{t},o\sim\theta_{t}(a)}\left[e^{-\ln(\text{bet}(B^{*},\guess,a,o))}\right]\right)-\ln\left(\frac{1}{\delta}\right)$$

The history dependence is implicit in the notation $\pi_{t},\theta_{t}$. Composing this with the previous inequality, and moving the logarithm over, we have
$$\ln\left(2|\hypo|\right)+\ln\left(\frac{1}{\delta}\right)\geq-\tsum\ln\left(\bel{\expec}{a\sim\pi_{t},o\sim\theta_{t}(a)}\left[e^{-\ln(\text{bet}(B^{*},\guess,a,o))}\right]\right)$$

We can re-express this as
$$\ln\left(\frac{2|\hypo|}{\delta}\right)\geq-\tsum\ln\left(\bel{\expec}{a\sim\pi_{t},o\sim\theta_{t}(a)}\left[\frac{1}{\text{bet}(B^{*},\guess,a,o)}\right]\right)$$

Unpacking the definition of the betting function in the robust universal estimator, we have
$$=-\tsum\ln\left(\expec_{\pi_{t},\theta_{t}}\left[\frac{1}{\sqrt{\frac{\mu_{\guess,B^{*},a}(o)}{\guess(a)(o)}}+\hells(\guess(a)\to M^{*}(a))}\right]\right)$$

Removing the Hellinger-squared error makes this term smaller, and permits us to flip the square root. Re-expressing the expectations, we have
$$\geq-\tsum\ln\left(\expec_{\pi_{t}}\left[\expec_{\theta_{t}(a)}\left[\sqrt{\frac{\guess(a)(o)}{\mu_{\guess,B^{*},a}(o)}}\right]\right]\right)$$

Now, if we start at $\mu_{\guess,B^{*},a}$ and travel in the direction of $\theta_{t}(a)$, the Hellinger distance to $\guess(a)$ can only increase, because $\mu_{\guess,B^{*},a}$ is the Hellinger distance minimizer to $\guess(a)$ within $M^{*}(a)$, and $\theta_{t}(a)$ is also a distribution within $M^{*}(a)$, by the definition of $\theta$ being consistent with $M^{*}$. This justifies the equation
$$0\leq d(\lambda\mu.2\hells(\mu,\guess(a)))_{\mu_{\guess,B^{*},a}}(\theta_{t}(a)-\mu_{\guess,B^{*},a})$$

By explicitly computing this derivative, we arrive at
$$0\leq\left\langle-\sqrt{\frac{\guess(a)}{\mu_{\guess,B^{*},a}}},\theta_{t}(a)-\mu_{\guess,B^{*},a}\right\rangle$$

Writing the inner product as an expectation and reshuffling, we have derived
$$\expec_{\theta_{t}(a)}\left[\sqrt{\frac{\guess(a)(o)}{\mu_{\guess,M^{*},a}(o)}}\right]\leq\expec_{\mu_{\guess,M^{*},a}}\left[\sqrt{\frac{\guess(a)(o)}{\mu_{\guess,M^{*},a}(o)}}\right]$$

Because of this fact, along with the definition of Hellinger-squared error, we can derive
$$-\tsum\ln\left(\expec_{\pi_{t}}\left[\expec_{\theta_{t}(a)}\left[\sqrt{\frac{\guess(a)(o)}{\mu_{\guess,B^{*},a}(o)}}\right]\right]\right)\geq-\tsum\ln\left(\expec_{\pi_{t}}\left[\expec_{\mu_{\guess,B^{*},a}}\left[\sqrt{\frac{\guess(a)(o)}{\mu_{\guess,B^{*},a}(o)}}\right]\right]\right)$$
$$=-\tsum\ln\left(\expec_{\pi_{t}}\left[\sum_{o}\sqrt{\guess(a)(o)\cdot\mu_{\guess,B^{*},a}(o)}\right]\right)=-\tsum\ln\left(\expec_{\pi_{t}}\left[1-\hells(\mu_{\guess,B^{*},a},\guess(a))\right]\right)$$
$$=-\tsum\ln\left(1-\expec_{\pi_{t}}\left[\hells(\guess(a)\to M^{*}(a))\right]\right)$$

Chaining all our inequalities together, and finishing off with $-\ln(1-x)\geq x$, we have derived that
$$\ln\left(\frac{2|\hypo|}{\delta}\right)\geq\tsum\bel{\expec}{a\sim\pi_{t}}\left[\hells(\guess(a)\to M^{*}(a))\right]$$
holds with $>1-\delta$ probability according to $\theta\bowtie\pi$. $M^{*}$, $\theta$, $\delta$, and $\pi$ were arbitrary, so this provides a value for $\beta_{\estim}(T,\delta)$.
\\

To compute $\alpha_{\estim}(T,\delta)$, we will use Azuma's inequality. For the pessimistic better, we were previously at $\ln(2)\geq\tsum\ln(\text{bet}(B,\guess,a_{t},o_{t}))$ from Lemma \ref{lem6}. This holds approximately, and the left-hand side is technically $\ln\left(\frac{1}{\frac{1}{2}-\eps'}\right)$, but in the $\eps'\to 0$ limit, this inequality holds. Substituting in the pessimism bet, we get
$$\ln(2)\geq\tsum\ln(1+\eps(\expec_{\guess(a_{t})}[r(a_{t},o)]-r(a_{t},o_{t})))$$

$\eps$ was defined to ensure that it was always $\frac{1}{2}$ or less, so we'll use the fact that $\ln(1+x)\geq x-x^{2}$ over the interval $[-0.5,0.5]$, to get
$$\ln(2)\geq\tsum\eps(\expec_{\guess(a_{t})}[r(a_{t},o)]-r(a_{t},o_{t}))-\eps^{2}(\expec_{\guess(a_{t})}[r(a_{t},o)]-r(a_{t},o_{t}))^{2}$$

Upper-bounding the difference in rewards by 1, and adding to both sides, we get
$$\ln(2)+T\eps^{2}\geq\tsum\eps(\expec_{\guess(a_{t})}[r(a_{t},o)]-r(a_{t},o_{t}))$$

Azuma's inequality can be used to show that, no matter what $\theta,\pi$ are, with $1-\delta$ probability under $\theta\bowtie\pi$, we have
$$\tsum\eps\bel{\expec}{a\sim\pi_{t},o\sim\theta_{t}(a)}\left[\bel{\expec}{o'\sim\guess(a)}[r(a,o')]-r(a,o)\right]$$
$$\leq\tsum\eps(\bel{\expec}{o\sim\guess(a_{t})}[r(a_{t},o)]-r(a_{t},o_{t}))+\sqrt{2\ln\left(\frac{1}{\delta}\right)\tsum\eps^{2}}$$

Rearranging and combining with the previous inequality, we get
$$\ln(2)+T\eps^{2}+\sqrt{2\ln\left(\frac{1}{\delta}\right)\tsum\eps^{2}}\geq\tsum\eps\bel{\expec}{a\sim\pi_{t},o\sim\theta_{t}(a)}\left[\bel{\expec}{o'\sim\guess(a)}[r(a,o')]-r(a,o)\right]$$

Simplifying the contents of the square root, using the $f$ notation to abbreviate expected rewards, and dividing both sides by epsilon, we get
$$\frac{\ln(2)}{\eps}+T\eps+\sqrt{2\ln\left(\frac{1}{\delta}\right)T}\geq\tsum\bel{\expec}{a\sim\pi_{t}}\left[f^{\guess}(a)-f^{\theta_{t}}(a)\right]$$

Finally, if $\sqrt{\frac{\ln(2)}{T}}\leq\frac{1}{2}$ (which holds for all $T\geq 3$), then $\eps=\sqrt{\frac{\ln(2)}{T}}$ and plugging that in yields a bound of
$$\sqrt{T}\left(2\sqrt{\ln(2)}+\sqrt{2\ln\left(\frac{1}{\delta}\right)}\right)\geq\tsum\bel{\expec}{a\sim\pi_{t}}\left[f^{\guess}(a)-f^{\theta_{t}}(a)\right]$$

This bound covers all $T\geq 3$. This establishes a value for $\alpha_{\estim}(T,\delta)$. $\blacksquare$

\begin{prop} \label{pro5}
If $\obs$ is finite, then for all hypothesis classes $\hypo$, there exists an online estimator $\estim$ where $\beta_{\estim}(T,\delta)\leq\bel{\min}{\eps>0}\left(2\ln\left(\frac{2\mathcal{N}(\hypo,\eps)}{\delta}\right)+8T\eps^{2}\right)$, and
$\alpha_{\estim}(T,\delta)\leq\sqrt{T}\left(2\sqrt{\ln(2)}+\sqrt{2\ln\left(\frac{1}{\delta}\right)}\right)$
\end{prop}

Given a model $N$, $B_{N,\eps}$ will denote the ball of all models which are $\eps$-close to $N$ in Hellinger distance. 
$$B_{N,\eps}=\Set{N}{\forall a\in\acts:D_{H}(N(a),M(a))\leq\eps}$$
Here we are using Hausdorff distance between the sets $N(a)$ and $M(a)$, instead of an asymmetric notion of distance. Given a model $N$, let $N_{\eps}$ denote the $\eps$-thickened version of $N$, defined as
$$N_{\eps}(a):=\Set{\mu}{D_{H}(\mu\to N(a))\leq\eps}$$
By the definition of $\mathcal{N}(\hypo,\eps)$ (the minimum number of models needed for an $\eps$-approximate cover of $\hypo$), there is some finite set $\mathcal{C}_{\eps}$ of models, where $\hypo\subseteq\bel{\bigcup}{M'\in\mathcal{C}}B_{M',\eps}$, and $|\mathcal{C}|=\mathcal{N}(\hypo,\eps)$.

Our estimator will be RUE, with its finite hypothesis space being the $N_{\eps}$ models, for $N\in\mathcal{C}_{\eps}$. This establishes the value of $\alpha_{\estim}(T,\delta)$, by Theorem \ref{the2}. We will now compute $\beta_{\estim}(T,\delta)$ for this estimator. If $M^{*}$ is a true hypothesis, let $N^{*}$ denote some model $N\in\mathcal{C}_{\eps}$ such that $M^{*}\in B_{N,\eps}$. Such a model will always exist because the balls cover $\hypo$. By Lemma \ref{lem5} applied twice, for any algorithm $\pi$, we have
$$\tsum\bel{\expec}{a\sim\pi_{t}}\left[\hells(\guess(a)\to M^{*}(a))\right]$$
$$\leq 2\tsum\expec_{\pi_{t}}\left[\hells(\guess(a)\to N_{\eps}^{*}(a))\right]+2\tsum\expec_{\pi_{t}}\left[\hells(N^{*}_{\eps}(a)\to M^{*}(a))\right]$$
$$\leq2\tsum\expec_{\pi_{t}}\left[\hells(\guess(a)\to N_{\eps}^{*}(a))\right]+4\tsum\expec_{\pi_{t}}\left[\hells(N^{*}_{\eps}(a)\to N^{*}(a))+\hells(N^{*}(a)\to M^{*}(a))\right]$$

$N^{*}_{\eps}(a)$ consists of all distributions with a Hellinger distance of $\eps$ or less from $N^{*}(a)$, so the middle term is at most $\eps^{2}$. $N^{*}(a)$ has a symmetric Hellinger distance of $\eps$ or less from $M^{*}(a)$ because $M^{*}\in B_{N^{*},\eps}$, so the latter sum is at most $\eps^{2}$ as well, yielding
$$\leq2\tsum\expec_{\pi_{t}}\left[\hells(\guess(a)\to N_{\eps}^{*}(a))\right]+8T\eps^{2}$$

For the first term, observe that, because all actions have $M^{*}(a)\subseteq N^{*}_{\eps}(a)$, and the environment is consistent with $M^{*}$, $N^{*}_{\eps}$ is a true hypothesis as well. Therefore, by Theorem \ref{the2} along with $|\mathcal{C}_{\eps}|=\mathcal{N}(\hypo,\eps)$, with $1-\delta$ probability according to $\theta\bowtie\pi$, we have
$$\leq2\ln\left(\frac{2\mathcal{N}(\hypo,\eps)}{\delta}\right)+8T\eps^{2}$$

Because $\eps$ can be whatever we wish, we can take the minimum over all values of $\eps$, yielding our conclusion. $\blacksquare$
\section{Definitions for Robust Linear Bandits:}
$\acts,\obs$ are finite sets of actions and observations. We consider the space $\Delta\obs$ to be embedded in the vector space $\mathbb{R}^{\obs}$ in the typical way. $r:\acts\times\obs\to[0,1]$ is some known reward function.

$\mathcal{Z},\mathcal{W}$ are finite-dimensional vector spaces. $Z$ and $W$ are their dimensions. $\hypo$ is a compact subset of $\mathcal{Z}$, consisting of the hypotheses.

$F:\acts\times\mathcal{Z}\times\mathbb{R}^{\obs}\to\mathcal{W}$ is a function which is bilinear in $\mathcal{Z}$ and $\mathbb{R}^{\obs}$. Given some $M:\mathcal{Z}$ and $a:\acts$, $F_{a,M}:\mathbb{R}^{\obs}\to\mathcal{W}$ is the induced linear function, and $K_{a,M}\subseteq\mathbb{R}^{\obs}$ is the nullspace of $F_{a,M}$.

Given an $M\in\hypo$, we will overload notation so that $M$ also denotes a model of type $\acts\to\Box\obs$, via the definition $M(a):=K_{a,M}\cap\Delta\obs$. We will also overload notation so that $\hypo$ denotes the set of models which $\hypo$ induces.

$a^{*}_{M}$ is $\bel{\text{argmax}}{a\in\acts}f^{M}(a)$, the action which maximizes the worst-case expected reward for $M$.
\\

The spaces $\mathbb{R}^{\obs},\mathcal{W},\mathcal{Z}$ are equipped with a variety of norms. Given a norm $||.||_{\text{norm}}$, $B^{u}_{\text{norm}}$ denotes the unit ball of that norm, and $D_{\text{norm}}$ denotes distances measured with respect to that norm.

The space $\mathbb{R}^{\obs}$ is equipped with the L1 norm. $||y||_{\mathbb{R}^{\obs}}$ denotes this norm.

The space $\mathcal{W}$ is equipped with the following three norms. $||w||_{\text{small},W}$ is the norm where 
\\
$B^{u}_{\text{small},W}:=\bel{\bigcap}{a\in\acts,M\in\hypo}F_{a,M}(B^{u}_{\mathbb{R}^{\obs}})$. Equivalently, $||w||_{\text{small},W}:=\bel{\max}{a\in\acts,M\in\hypo}\bel{\min}{y:F_{a,M}(y)=w}||y||_{\mathbb{R}^{\obs}}$.

$||w||_{\text{big},W}$ is the norm where $||w||_{\text{big},W}:=\frac{||w||_{\text{small},W}}{\bel{\max}{a\in\acts,M\in\hypo,y\in B^{u}_{\mathbb{R}^{\obs}}}||F_{a,M}(y)||_{\text{small},W}}$. $B^{u}_{\text{big},W}$ can be thought of as $B^{u}_{\text{small},W}$, expanded so that it engulfs the set $\bel{\bigcup}{a,M\in\hypo}F_{a,M}(B^{u}_{\mathbb{R}^{\obs}})$.

Finally, $||w||_{2,W}$ is the norm where $B^{u}_{2,W}$ is the outer John ellipsoid of $B^{u}_{\text{big},W}$. It is an L2 norm and induces an inner product on $\mathcal{W}$.
\\

The space $\mathcal{Z}$ is equipped with the following two norms. 
$$||M||_{Z}:=\max_{a\in\acts,y\in B^{u}_{\mathbb{R}^{\obs}}}||F_{a,M}(y)||_{\text{big},W}$$
$||M||_{2,Z}$ is the norm where $B^{u}_{2,Z}$ is the outer John ellipsoid of $B^{u}_{Z}$. It is an L2 norm.

The quantity $R$ is $\bel{\max}{a\in\acts,M\in\hypo,y\in B^{u}_{\mathbb{R}^{\obs}}}||F_{a,M}(y)||_{\text{small},W}$. We have $R||w||_{\text{big},W}=||w||_{\text{small},W}$. $\text{sine}$ is a parameter which is defined in \cite{Kosoy2024}.
\section{Robust Linear Bandit Theorems}
\begin{lemm} \label{lem7}
In the robust linear bandits setting, letting $\zeta:\Delta\hypo$, and $\gen:\acts\to\Delta\obs$, and $\mathcal{G}$ be a set of functions of type $\acts\to\mathbb{R}^{\geq 0}$ defined as $\mathcal{G}:=\Set{\lambda a.||F_{a,M}(\gen(a))||_{2,W}}{M\in\hypo}$, we have that for any $\Delta',\eps'>0$, Foster's disagreement coefficient, $\text{dis}(\mathcal{G},\Delta',\eps',\zeta)\leq Z$.
\end{lemm}

Foster's disagreement coefficient, from \cite{Foster21}, is defined as
$$\text{dis}(\mathcal{G},\Delta',\eps',\zeta)=\bel{\max}{\Delta\geq\Delta',\eps\geq\eps'}\frac{\Delta^{2}}{\eps^{2}}\bel{\mathbb{P}}{N\sim\zeta}\left(\exists g\in\mathcal{G}:g(a^{*}_{N})>\Delta,\bel{\expec}{M\sim\zeta}[g^{2}(a^{*}_{M})]\leq\eps^{2}\right)$$

Define the following subset of $\mathcal{Z}$.
$$G_{\eps,\zeta}:=\Set{M\in\mathcal{Z}}{\bel{\expec}{N\sim\zeta}[g^{2}_{M}(a^{*}_{N})]\leq\eps^{2}}$$

Note that the second condition in the existence statement of Foster's disagreement coefficient is precisely the requirement that the function $g$ correspond to some $M\in G_{\eps,\zeta}\cap\hypo$. Therefore, we can upper-bound the disagreement coefficient by
$$\leq\bel{\max}{\Delta\geq\Delta',\eps\geq\eps'}\frac{\Delta^{2}}{\eps^{2}}\bel{\mathbb{P}}{N\sim\zeta}\left(\exists M\in G_{\eps,\zeta}:g_{M}(a^{*}_{N})>\Delta\right)=\bel{\max}{\Delta\geq\Delta',\eps\geq\eps'}\frac{\Delta^{2}}{\eps^{2}}\bel{\mathbb{P}}{N\sim\zeta}\left(\max_{M\in G_{\eps,\zeta}}g^{2}_{M}(a^{*}_{N})>\Delta^{2}\right)$$

By Markov's inequality, we can derive
$$\leq\bel{\max}{\Delta\geq\Delta',\eps\geq\eps'}\frac{\Delta^{2}}{\eps^{2}}\frac{\bel{\expec}{N\sim\zeta}\left[\bel{\max}{M\in G_{\eps,\zeta}}g^{2}_{M}(a^{*}_{N})\right]}{\Delta^{2}}=\bel{\max}{\eps\geq\eps'}\frac{\bel{\expec}{N\sim\zeta}\left[\bel{\max}{M\in G_{\eps,\zeta}}g^{2}_{M}(a^{*}_{N})\right]}{\eps^{2}}$$

We will now upper-bound the expectation. In the robust linear bandit setting, the space $\mathcal{W}$ is equipped with a notion of inner product. We'll use this to equip the hypothesis space $\mathcal{Z}$ with something which is almost an inner product (but which does not correspond to either of the two specified norms on $\mathcal{Z}$), namely, 
$$\langle M,M'\rangle_{\zeta}:=\bel{\expec}{N\sim\zeta}\left[\left\langle F_{a^{*}_{N},M}(\gen(a^{*}_{N})),F_{a^{*}_{N},M'}(\gen(a^{*}_{N}))\right\rangle_{W}\right]$$

The reason for this definition is
$$g^{2}_{M}(a^{*}_{N})=||F_{a^{*}_{N},M}(\gen(a^{*}_{N}))||^{2}_{2,W}=\langle F_{a^{*}_{N},M}(\gen(a^{*}_{N})),F_{a^{*}_{N},M}(\gen(a^{*}_{N}))\rangle_{W}$$

Which implies that $\bel{\expec}{N\sim\zeta}[g^{2}_{M}(a^{*}_{N})]=\langle M,M\rangle_{\zeta}$.
Due to the inner product on $\mathcal{W}$, and the linearity of $F$ in the second subscript argument, this "inner product" on $\mathcal{Z}$ obeys all properties of an inner product except one. There might be nonzero vectors $M$ where $\langle M,M\rangle_{\zeta}=0$. These are exactly the $M$ where, with probability one according to $N\sim\zeta$, $F_{a^{*}_{N},M}(\gen(a^{*}_{N}))=0$.

By linear algebra, we have the following. There is a subspace $\mathcal{Z}'$ where this "inner product" is a true inner product, and a canonical surjection $pr:\mathcal{Z}\to\mathcal{Z'}$, and for all $M\in\mathcal{Z}$, we have $\langle M,M\rangle_{\zeta}=\langle pr(M),pr(M)\rangle_{\zeta}$.

We'll now prove that, with probability 1 according to $N\sim\zeta$, we have that for all $M\in\mathcal{Z}$, 
\\
$g^{2}_{M}(a^{*}_{N})=g^{2}_{pr(M)}(a^{*}_{N})$. This occurs because, by the definition of $g$, the inner product on $\mathcal{W}$, and linearity of $F$ in its second subscript argument, we have
$$g^{2}_{M}(a^{*}_{N})=||F_{a^{*}_{N},M}(\gen(a^{*}_{N}))||^{2}_{2,W}=\langle F_{a^{*}_{N},M}(\gen(a^{*}_{N})),F_{a^{*}_{N},M}(\gen(a^{*}_{N}))\rangle_{W}=$$
$$\left\langle F_{a^{*}_{N},pr(M)}(\gen(a^{*}_{N}))+F_{a^{*}_{N},M-pr(M)}(\gen(a^{*}_{N})),F_{a^{*}_{N},pr(M)}(\overline{M}(a^{*}_{N}))+F_{a^{*}_{N},M-pr(M)}(\gen(a^{*}_{N}))\right\rangle_{W}$$

There's a finite collection of vectors $z_{i}$ which serve as a basis  for the subspace of vectors of the form $M-pr(M)$, and for all of these vectors, with probability 1 when $N\sim\zeta$, $F_{a^{*}_{N},z_{i}}(\gen(a^{*}_{N}))=0$. So, with probability 1, for every $M$, we have
$$=\langle F_{a^{*}_{N},pr(M)}(\gen(a^{*}_{N})),F_{a^{*}_{N},pr(M)}(\gen(a^{*}_{N}))\rangle_{W}=g^{2}_{pr(M)}(a^{*}_{N})$$

Now, we can return to the term we were trying to upper-bound, and go
$$\bel{\max}{\eps\geq\eps'}\frac{\bel{\expec}{N\sim\zeta}\left[\bel{\max}{M\in G_{\eps,\zeta}}g^{2}_{M}(a^{*}_{N})\right]}{\eps^{2}}=\bel{\max}{\eps\geq\eps'}\frac{\bel{\expec}{N\sim\zeta}\left[\bel{\max}{M\in G_{\eps,\zeta}}g^{2}_{pr(M)}(a^{*}_{N})\right]}{\eps^{2}}$$

Then, by the definition of $G_{\eps,\zeta}$ as the $M$ where $\bel{\expec}{N\sim\zeta}[g^{2}_{M}(a^{*}_{N})]\leq\eps^{2}$, and our rephrasing of this as an "inner product", and the fact that this "inner product" doesn't change under projection, we can rewrite as
$$=\bel{\max}{\eps\geq\eps'}\frac{\bel{\expec}{N\sim\zeta}\left[\bel{\max}{M\in\mathcal{Z}:\langle M,M\rangle_{\zeta}\leq\eps^{2}}g^{2}_{pr(M)}(a^{*}_{N})\right]}{\eps^{2}}=\bel{\max}{\eps\geq\eps'}\frac{\bel{\expec}{N\sim\zeta}\left[\bel{\max}{M\in\mathcal{Z}:\langle pr(M),pr(M)\rangle_{\zeta}\leq\eps^{2}}g^{2}_{pr(M)}(a^{*}_{N})\right]}{\eps^{2}}$$
$$=\bel{\max}{\eps\geq\eps'}\frac{\bel{\expec}{N\sim\zeta}\left[\bel{\max}{M'\in\mathcal{Z}':\langle M',M'\rangle_{\zeta}\leq\eps^{2}}g^{2}_{M'}(a^{*}_{N})\right]}{\eps^{2}}$$

Now, since the "inner product" is a true inner product on $\mathcal{Z}'$, we're maximizing over an L2 ball, so every $M'$ fulfilling those conditions can be written as a spherical combination of orthogonal basis vectors $z_{i}\in\mathcal{Z}'$, which all have an inner product of $\eps^{2}$. Then unpack definitions and use linearity of $F$ in its second subscript argument.
$$=\bel{\max}{\eps\geq\eps'}\frac{\bel{\expec}{N\sim\zeta}\left[\bel{\max}{\alpha_{i}:\sum_{i}\alpha^{2}_{i}\leq 1}g^{2}_{\sum_{i}\alpha_{i}z_{i}}(a^{*}_{N})\right]}{\eps^{2}}=\bel{\max}{\eps\geq\eps'}\frac{\bel{\expec}{N\sim\zeta}\left[\bel{\max}{\alpha_{i}:\sum_{i}\alpha^{2}_{i}\leq 1}||F_{a^{*}_{N},\sum_{i}\alpha_{i}z_{i}}(\gen(a^{*}_{N}))||^{2}_{2,W}\right]}{\eps^{2}}$$
$$=\bel{\max}{\eps\geq\eps'}\frac{\bel{\expec}{N\sim\zeta}\left[\bel{\max}{\alpha_{i}:\sum_{i}\alpha^{2}_{i}\leq 1}||\sum_{i}\alpha_{i}F_{a^{*}_{N},z_{i}}(\gen(a^{*}_{N}))||^{2}_{2,W}\right]}{\eps^{2}}$$

Upper-bound the norm of the sum by the sum of the norms, and use the Cauchy-Schwartz inequality.
$$\leq\bel{\max}{\eps\geq\eps'}\frac{\bel{\expec}{N\sim\zeta}\left[\bel{\max}{\alpha_{i}:\sum_{i}\alpha^{2}_{i}\leq 1}\left(\sum_{i}|\alpha_{i}|\cdot||F_{a^{*}_{N},z_{i}}(\gen(a^{*}_{N}))||_{2,W}\right)^{2}\right]}{\eps^{2}}$$
$$\leq\bel{\max}{\eps\geq\eps'}\frac{\bel{\expec}{N\sim\zeta}\left[\bel{\max}{\alpha_{i}:\sum_{i}\alpha^{2}_{i}\leq 1}\left(\sum_{i}\alpha^{2}_{i}\right)\left(\sum_{i}||F_{a^{*}_{N},z_{i}}(\gen(a^{*}_{N}))||^{2}_{2,W}\right)\right]}{\eps^{2}}$$
$$=\bel{\max}{\eps\geq\eps'}\frac{\bel{\expec}{N\sim\zeta}\left[\sum_{i=1}^{\text{dim}_{Z'}}||F_{a^{*}_{N},z_{i}}(\gen(a^{*}_{N}))||^{2}_{2,W}\right]}{\eps^{2}}=\bel{\max}{\eps\geq\eps'}\frac{\sum_{i=1}^{\text{dim}_{Z'}}\bel{\expec}{N\sim\zeta}\left[||F_{a^{*}_{N},z_{i}}(\gen(a^{*}_{N}))||^{2}_{2,W}\right]}{\eps^{2}}$$

Pack it back up as an inner product, use that the $z_{i}$ were selected to all have an inner product of $\eps^{2}$, and use that $Z$ is the dimension of $\mathcal{Z}$.
$$=\bel{\max}{\eps\geq\eps'}\frac{\sum_{i=1}^{\text{dim}_{Z'}}\langle z_{i},z_{i}\rangle_{\zeta}}{\eps^{2}}\leq\bel{\max}{\eps\geq\eps'}\frac{Z\eps^{2}}{\eps^{2}}=Z$$

And so the disagreement coefficient has a maximum value of $Z$, the dimension of the hypothesis space. $\blacksquare$

\begin{lemm} \label{lem8}
Let $\kappa$ be $7\left(\frac{1}{\text{sine}}+1\right)R\sqrt{WZ}$ and $\gamma$ be $\geq 2e^{2}\kappa$ where $e$ is Euler's constant. We then have that for all beliefs $\gen:\acts\to\Delta\obs$,
$$\text{dec}^{o}_{\gamma}(\hypo,\gen)\leq\frac{\kappa^{2}\ln^{2}\left(\frac{\gamma}{2\kappa}\right)}{\gamma}$$
\end{lemm}

Begin by unpacking the definition of the offset DEC.
$$\text{dec}^{o}_{\gamma}(\hypo,\gen)=\bel{\min}{p\in\Delta\acts}\bel{\max}{\mu\in\Delta\hypo}\bel{\expec}{M\sim\mu}[\max(f^{M})]-\bel{\expec}{a\sim p}[f^{\gen}(a)]-\gamma\bel{\expec}{a,M\sim p,\mu}[\hells(\gen(a)\to M(a))]$$

This is affine in both arguments. The hypothesis space and action space are compact, so $\Delta\hypo$ and $\Delta\acts$ are compact. Therefore, we can use Sion's minimax theorem, to swap the max and the min.
$$=\bel{\max}{\mu\in\Delta\hypo}\bel{\min}{p\in\Delta\acts}\expec_{\mu}\left[\max(f^{M})\right]-\expec_{p}\left[f^{\overline{M}}\right]-\gamma\expec_{p,\mu}[\hells(\gen(a)\to M(a))]$$

Given $\mu$, let $p$ be the distribution over actions produced by sampling $M\sim\mu$, and selecting the optimal action, $a^{*}_{M}$.
$$\leq\bel{\max}{\mu\in\Delta\hypo}\bel{\expec}{M\sim\mu}\left[f^{M}(a^{*}_{M})-f^{\gen}(a^{*}_{M})\right]-\gamma\bel{\expec}{M,N\sim\mu}[\hells(\gen(a^{*}_{N})\to M(a^{*}_{N}))]$$

Use that $\sqrt{2}$ times the Hellinger distance exceeds total variation distance, so 2 times the Hellinger-squared distance exceeds the total variation distance squared.
$$\leq\bel{\max}{\mu\in\Delta\hypo}\expec_{\mu}\left[f^{M}(a^{*}_{M})-f^{\gen}(a^{*}_{M})\right]-\frac{\gamma}{2}\bel{\expec}{M,N\sim\mu}\left[D^{2}_{TV}(\gen(a^{*}_{N})\to M(a^{*}_{N}))\right]$$

Letting $\nu_{M}$ be the point in $M(a^{*}_{M})$ which minimizes total variation distance to $\gen(a^{*}_{M})$, we have
$$f^{M}(a^{*}_{M})-f^{\gen}(a^{*}_{M})=\bel{\min}{\nu\in M(a^{*}_{M})}\bel{\expec}{o\sim\nu}[r(a^{*}_{M},o)]-\bel{\expec}{o\sim\gen(a^{*}_{M})}[r(a^{*}_{M},o)]$$
$$\leq\expec_{\nu_{M}}[r(a^{*}_{M},o)]-\expec_{\gen(a^{*}_{M})}[r(a^{*}_{M},o)]\leq D_{TV}(\nu_{M},\gen(a^{*}_{M}))=D_{TV}(\gen(a^{*}_{M})\to M(a^{*}_{M}))$$

So, we can continue to upper-bound by 
$$\leq\bel{\max}{\mu\in\Delta\hypo}\expec_{\mu}[D_{TV}(\gen(a^{*}_{M})\to M(a^{*}_{M}))]-\frac{\gamma}{2}\bel{\expec}{M,N\sim\mu}\left[D^{2}_{TV}(\gen(a^{*}_{N})\to M(a^{*}_{N}))\right]$$

At this point, we use that the norm on $\mathbb{R}^{\obs}$ is the L1 norm, which coincides with total variation distance, to rewrite as 
$$\leq\bel{\max}{\mu\in\Delta\hypo}\expec_{\mu}[D_{\mathbb{R}^{\obs}}(\gen(a^{*}_{M})\to M(a^{*}_{M}))]-\frac{\gamma}{2}\bel{\expec}{M,N\sim\mu}\left[D^{2}_{\mathbb{R}^{\obs}}(\gen(a^{*}_{N})\to M(a^{*}_{N}))\right]$$

Now, note that the squared distance term (with $\gamma$) being 1 or more on all $\mu$ implies that the offset DEC is 0 or less and the lemma holds, so for the $\mu$ which is picked, we may assume that the squared distance times $\frac{\gamma}{2}$ is $\leq 1$. $\gamma\geq 14e^{2}$ because we assumed that $\gamma\geq 2\kappa e^{2}$ and $\kappa\geq 7$, so the squared distance term (without $\gamma$) is $\leq\frac{1}{7e^{2}}$. This will be used later. By Lemmas A.2 and A.7 from \cite{Kosoy2024}, we have
$$\leq\bel{\max}{\mu\in\Delta\hypo}\expec_{\mu}\left[4\left(\frac{1}{\text{sine}(K_{a^{*}_{M},M},\Delta\obs)}+1\right)D_{\mathbb{R}^{\obs}}(\gen(a^{*}_{M})\to K_{a^{*}_{M},M})\right]$$
$$-\frac{\gamma}{2}\bel{\expec}{M,N\sim\mu}\left[D^{2}_{\mathbb{R}^{\obs}}(\gen(a^{*}_{N})\to M(a^{*}_{N}))\right]$$

Letting $\text{sine}$ be defined as $\bel{\min}{M\in\hypo,a\in\acts}\text{sine}(K^{b}_{a,M}(a),\Delta\obs)$, we can then reexpress as
$$\leq\bel{\max}{\mu\in\Delta\hypo}\expec_{\mu}\left[4\left(\frac{1}{\text{sine}}+1\right)D_{\mathbb{R}^{\obs}}(\gen(a^{*}_{M})\to K_{a^{*}_{M},M})\right]-\frac{\gamma}{2}\bel{\expec}{M,N\sim\mu}\left[D^{2}_{\mathbb{R}^{\obs}}(\gen(a^{*}_{N})\to M(a^{*}_{N}))\right]$$

Now, we'll move to measuring distance in $\mathcal{W}$. It is immediate from the definitions of the relevant norms that, for any action $a$, model $M$, and points $y,y'\in\mathbb{R}^{\obs}$, 
\\
$D_{\mathbb{R}^{\obs}}(y,y')\geq||F_{a,M}(y)-F_{a,M}(y')||_{2,W}$, because all $F_{a,M}$ map the unit ball of the norm on $\mathbb{R}^{\obs}$ inside the unit ball of the L2 norm on $\mathcal{W}$. Using this fact, along with the fact that $M(a^{*}_{N})$ consists of all the points in $\Delta\obs$ which $F_{a^{*}_{N},M}$ maps to zero, we have that 
\\
$D^{2}_{\mathbb{R}^{\obs}}(\gen(a^{*}_{N})\to M(a^{*}_{N}))\geq||F_{a^{*}_{N},M}(\gen(a^{*}_{N}))||^{2}_{2,W}$ which yields
$$\leq\bel{\max}{\mu\in\Delta\hypo}\expec_{\mu}\left[4\left(\frac{1}{\text{sine}}+1\right)D_{\mathbb{R}^{\obs}}(\gen(a^{*}_{M})\to K_{a^{*}_{M},M})\right]-\frac{\gamma}{2}\bel{\expec}{M,N\sim\mu}\left[||F_{a^{*}_{N},M}(\gen(a^{*}_{N}))||^{2}_{2,W}\right]$$

In the other direction, we have $D_{\mathbb{R}^{\obs}}(y,y')\leq||F_{a,M}(y)-F_{a,M}(y')||_{\text{small},W}$ by the definition of the small norm on $\mathcal{W}$. Again using that $K_{a^{*}_{M},M}$ consists of all the points which $F_{a^{*}_{M},M}$ maps to zero, we have $D_{\mathbb{R}^{\obs}}(\gen(a^{*}_{M})\to K_{a^{*}_{M},M})\leq||F_{a^{*}_{M},M}(\gen(a^{*}_{M}))||_{\text{small},W}$ which yields
$$\leq\bel{\max}{\mu\in\Delta\hypo}\expec_{\mu}\left[4\left(\frac{1}{\text{sine}}+1\right)||F_{a^{*}_{M},M}(\gen(a^{*}_{M}))||_{\text{small},W}\right]-\frac{\gamma}{2}\bel{\expec}{M,N\sim\mu}\left[||F_{a^{*}_{N},M}(\gen(a^{*}_{N}))||^{2}_{2,W}\right]$$

Now we use that $||w||_{\text{small},W}=R\cdot||w||_{\text{big},W}\leq R\sqrt{W}||w||_{2,W}$, by the definition of the big $\mathcal{W}$ norm, the definition of the L2 norm on $\mathcal{W}$, and John's Ellipsoid Theorem.
$$\leq\bel{\max}{\mu\in\Delta\hypo}\expec_{\mu}\left[4\left(\frac{1}{\text{sine}}+1\right)R\sqrt{W}||F_{a^{*}_{M},M}(\gen(a^{*}_{M}))||_{2,W}\right]-\frac{\gamma}{2}\bel{\expec}{M,N\sim\mu}\left[||F_{a^{*}_{N},M}(\gen(a^{*}_{N}))||^{2}_{2,W}\right]$$

At this point, we introduce the abbreviations
$$g_{M}(a):=||F_{a,M}(\gen(a^{*}_{M}))||_{2,W}$$
$$H:=\expec_{M,N\sim\mu}[g^{2}_{M}(a^{*}_{N})]$$

To abbreviate our equation as
\begin{equation} \label{eq:1}
\leq\bel{\max}{\mu\in\Delta\hypo}4\left(\frac{1}{\text{sine}}+1\right)R\sqrt{W}\bel{\expec}{M\sim\mu}[g_{M}(a^{*}_{M})]-\frac{\gamma}{2}H
\end{equation}

Now, we take a detour to upper-bound that expectation. Via page 105, Lemma E.2 of \cite{Foster21}, we can upper-bound the expectation term with Foster's disagreement coefficient. We have that $g_{M}(a)\leq 1$, because regardless of action or hypothesis, $F$ maps $\Delta\obs$ into the unit ball of our L2 norm, so Lemma E.2 can be invoked. Expressing the result in our notation, define 
\\
$\mathcal{G}:=\Set{g_{M}}{M\in\hypo}$. Foster's result from Lemma E.2 was
$$\bel{\expec}{M\sim\mu}[g_{M}(a^{*}_{M})]$$
$$\leq\bel{\min}{\Delta,\eps\in(0,1],\eta>0}\Delta+\frac{\eta\eps^{2}}{2}+\frac{1}{2\eta}\left(4\text{ dis}\left(\mathcal{G},\Delta,\eps,\mu\right)\ln\left(\frac{1}{\eps}\right)\ln\left(\frac{1}{\Delta}\right)+1\right)+\frac{\eta}{2}\bel{\expec}{M,N\sim\mu}[g^{2}_{M}(a^{*}_{N})]$$

Where $\text{dis}(\mathcal{G},\Delta,\eps,\mu)$ is the disagreement coefficient. Applying Lemma \ref{lem7} to upper-bound the disagreement coefficient by $Z$, and using our abbreviation $H$, we can proceed to 
$$\bel{\expec}{M\sim\mu}[g_{M}(a^{*}_{M})]\leq\bel{\min}{\Delta,\eps\in(0,1],\eta>0}\Delta+\frac{\eta\eps^{2}}{2}+\frac{1}{2\eta}\left(4Z\ln\left(\frac{1}{\eps}\right)\ln\left(\frac{1}{\Delta}\right)+1\right)+\frac{\eta}{2}H$$

Let $\Delta,\eps$ be $\sqrt{H}$, and rewrite to
$$\leq\bel{\min}{\eta>0}\sqrt{H}+\eta H+\frac{1}{2\eta}\left(Z\ln^{2}\left(\frac{1}{H}\right)+1\right)$$

Now, choose $\eta$ to be $\sqrt{\frac{Z}{2H}}\ln\left(\frac{1}{H}\right)$, and remember that $Z=\text{dim}(\mathcal{Z})\geq 1$ to get
$$\leq\sqrt{H}+\frac{1}{\sqrt{2}}\sqrt{ZH}\ln\left(\frac{1}{H}\right)+\frac{1}{\sqrt{2}}\sqrt{ZH}\ln\left(\frac{1}{H}\right)+\frac{1}{\sqrt{2}}\sqrt{\frac{H}{Z}}\ln^{-1}\left(\frac{1}{H}\right)$$
$$=\sqrt{\frac{ZH}{2}}\ln\left(\frac{1}{H}\right)\left(\sqrt{\frac{2}{Z}}\ln^{-1}\left(\frac{1}{H}\right)+2+\frac{1}{Z}\ln^{-2}\left(\frac{1}{H}\right)\right)$$
$$\leq\sqrt{\frac{ZH}{2}}\ln\left(\frac{1}{H}\right)\left(\sqrt{2}\ln^{-1}\left(\frac{1}{H}\right)+2+\ln^{-2}\left(\frac{1}{H}\right)\right)$$

Following the definition of that $H$ term, it equals $\bel{\expec}{M,N\sim\mu}\left[||F_{a,M}(\gen(a^{*}_{M}))||^{2}_{2,W}\right]$. We had previously derived that $D^{2}_{\mathbb{R}^{\obs}}(\gen(a^{*}_{N})\to M(a^{*}_{N}))\geq||F_{a^{*}_{N},M}(\gen(a^{*}_{N}))||^{2}_{2,W}$, so $H$ is upper-bounded by $\bel{\expec}{M,N\sim\mu}\left[D^{2}_{\mathbb{R}^{\obs}}(\gen(a^{*}_{N})\to M(a^{*}_{N}))\right]$. This quantity was previously demonstrated to be $\leq\frac{1}{7e^{2}}$. Putting these inequalities together, we can upper-bound $H$ by $\frac{1}{7e^{2}}$, and proceed to an upper bound of
$$\leq 1.72\sqrt{ZH}\ln\left(\frac{1}{H}\right)$$

Plugging this upper bound back into \ref{eq:1} and using that $4\cdot 1.72\leq 7$, we get
$$\leq\bel{\max}{\mu\in\Delta\hypo}7\left(\frac{1}{\text{sine}}+1\right)R\sqrt{WZH}\ln\left(\frac{1}{H}\right)-\frac{\gamma}{2}H$$

The only quantity that depends on $\mu$ is $H$, though it is suppressed in the notation. We shift to maximizing over the value of $H$ explicitly. $H\in[0,1]$ by its definition and the fact that the $g$ functions are in $[0,1]$. We also abbreviate $7\left(\frac{1}{\text{sine}}+1\right)R\sqrt{WZ}$ as $\kappa$, to get
$$\leq\bel{\max}{H\in[0,1]}7\left(\frac{1}{\text{sine}}+1\right)R\sqrt{WZH}\ln\left(\frac{1}{H}\right)-\frac{\gamma}{2}H=\bel{\max}{H\in[0,1]}\kappa\sqrt{H}\ln\left(\frac{1}{H}\right)-\frac{\gamma}{2} H$$
$$\leq\max\left(\bel{\max}{H\in\left[0,\frac{4\kappa^{2}}{\gamma^{2}}\right]}\left(\kappa\sqrt{H}\ln\left(\frac{1}{H}\right)-\frac{\gamma}{2} H\right),\bel{\max}{H\in\left[\frac{4\kappa^{2}}{\gamma^{2}},1\right]}\left(\kappa\sqrt{H}\ln\left(\frac{\gamma^{2}}{4\kappa^{2}}\right)-\frac{\gamma}{2}H\right)\right)$$

We'll show that the first term has a positive derivative (so the maximizing $H$ must be at the edge, and subsumed by the second term), and then exactly maximize the second term. For the first term, take the derivative to yield
$$-\frac{\kappa}{\sqrt{H}}+\frac{\kappa}{2\sqrt{H}}\ln\left(\frac{1}{H}\right)-\frac{\gamma}{2}\geq-\frac{\kappa}{\sqrt{H}}+\frac{\kappa}{2\sqrt{H}}\ln\left(\frac{\gamma^{2}}{4\kappa^{2}}\right)-\frac{\gamma}{2}=-\frac{\kappa}{\sqrt{H}}+\frac{\kappa}{\sqrt{H}}\ln\left(\frac{\gamma}{2\kappa}\right)-\frac{\gamma}{2}$$
$$=\frac{\kappa}{\sqrt{H}}\left(\ln\left(\frac{\gamma}{2\kappa}\right)-1\right)-\frac{\gamma}{2}\geq\frac{\gamma}{2}\left(\ln\left(\frac{\gamma}{2\kappa}\right)-1\right)-\frac{\gamma}{2}=\frac{\gamma}{2}\left(\ln\left(\frac{\gamma}{2\kappa}\right)-2\right)\geq 0$$

The first and second inequality follow because $\frac{1}{H}\geq\frac{\gamma^{2}}{4\kappa^{2}}$. The third inequality was our starting assumption that $\gamma\geq 2e^{2}\kappa$ and $\ln(e^{2})=2$. So, because the derivative is never negative, the maximizing value for $H$ is $\frac{4\kappa^{2}}{\gamma^{2}}$, which is subsumed by the later max, so our maximization over $H$ can be written as
$$=\bel{\max}{H\in\left[\frac{4\kappa^{2}}{\gamma^{2}},1\right]}\left(\kappa\sqrt{H}\ln\left(\frac{\gamma^{2}}{4\kappa^{2}}\right)-\frac{\gamma}{2} H\right)=\bel{\max}{H\in\left[\frac{4\kappa^{2}}{\gamma^{2}},1\right]}\left(2\kappa\sqrt{H}\ln\left(\frac{\gamma}{2\kappa}\right)-\frac{\gamma}{2}H\right)$$

The exact maximizing value of $H$ is $\frac{4\kappa^{2}\ln^{2}\left(\frac{\gamma}{2\kappa}\right)}{\gamma^{2}}$. Because we assumed that $\gamma\geq 2e^{2}\kappa$, this value is compatible with the range that $H$ may lie within. The exact maximum is then
$$=\frac{2\kappa^{2}\ln^{2}\left(\frac{\gamma}{2\kappa}\right)}{\gamma}$$

And our result follows by chaining all inequalities together. $\blacksquare$

\begin{theo} \label{the3}
In the robust linear bandit setting, for all $\eps<\frac{1}{e^{2}}$ (Euler's constant),
$$\text{dec}^{f}_{\eps}(\hypo)\leq 16\left(\frac{1}{\text{sine}}+1\right)R\sqrt{WZ}\eps\ln\left(\frac{1}{\eps}\right)$$
\end{theo}

By Proposition \ref{pro1}, to express the fuzzy DEC in terms of the offset DEC, and Proposition \ref{pro2}, to assume, without loss of generality, that probabilistic $\gen$ suffice to upper-bound the fuzzy DEC with respect to squared Hellinger loss, we may consider only probabilistic $\gen$, and compute
$$\text{dec}^{f}_{\eps}(\hypo,\gen)=\bel{\min}{\gamma\geq 0}\max(\text{dec}^{o}_{\gamma}(\hypo,\gen),0)+\gamma\eps^{2}$$

Select $\gamma:=\frac{\sqrt{2}\kappa}{\eps}\ln\left(\frac{1}{\eps}\right)$, where $\kappa:=7\left(\frac{1}{\text{sine}}+1\right)R\sqrt{WZ}$, as defined in Lemma \ref{lem8}. Note that, in particular, since $\eps\leq e^{-2}$ by assumption, we have $\gamma\geq 2e^{2}\kappa$, so we can safely invoke Lemma \ref{lem8} to bound the offset DEC. So, we can rewrite as
$$\leq\frac{2\kappa^{2}\ln^{2}\left(\frac{\gamma}{2\kappa}\right)}{\gamma}+\gamma\eps^{2}=\sqrt{2}\kappa\eps\left(\frac{\ln^{2}\left(\frac{1}{\sqrt{2}\eps}\ln\left(\frac{1}{\eps}\right)\right)}{\ln\left(\frac{1}{\eps}\right)}+\ln\left(\frac{1}{\eps}\right)\right)$$

The ratio of the fraction on the left side, and $\ln(\frac{1}{\eps})$, never exceeds 1.6 for $\eps$ in our range of interest, as can be verified by a graphing calculator, so we can upper-bound by
$$<\sqrt{2}\kappa\eps\cdot2.6\ln\left(\frac{1}{\eps}\right)<16\left(\frac{1}{\text{sine}}+1\right)R\sqrt{WZ}\eps\ln\left(\frac{1}{\eps}\right)$$
$\blacksquare$

\begin{lemm} \label{lem9}
For any two hypotheses $M,M'\in\hypo$, 
\\
$\max_{a}\hells(M(a),M'(a))\leq 4\left(\frac{1}{\text{sine}}+1\right)R\sqrt{Z}||M'-M||_{Z,2}$, where the Hellinger distance is symmetric Hellinger distance, and the norm is the L2 norm on $\mathcal{Z}$ from the definitions for robust linear bandits.
\end{lemm}

We start with the fact that total variation distance upper-bounds Hellinger-squared distance, and that total variation distance coincides with the norm on $\mathbb{R}^{\obs}$.
$$\bel{\max}{a}\hells(M(a)\to M'(a))\leq\bel{\max}{a}D_{TV}(M(a)\to M'(a))$$
$$=\bel{\max}{a}D_{\mathbb{R}^{\obs}}(M(a)\to M'(a))=\bel{\max}{a}\bel{\max}{\mu\in M(a)}d_{\mathbb{R}^{\obs}}(\mu\to M'(a))$$

Using Lemmas A.2 and A.7 from \cite{Kosoy2024}, and Kosoy's definition of the sine parameter, we have that $D_{\mathbb{R}^{\obs}}(\mu\to M'(a))\leq 4\left(\frac{1}{\text{sine}}+1\right)D_{\mathbb{R}^{\obs}}(\mu\to K_{a,M'})$, yielding
$$\leq 4\left(\frac{1}{\text{sine}}+1\right)\bel{\max}{a}\bel{\max}{\mu\in M(a)}D_{\mathbb{R}^{\obs}}(\mu\to K_{a,M'})$$

We have that, for all $y\in\mathbb{R}^{\obs}$, $D_{\mathbb{R}^{\obs}}(\mu,y)\leq||F_{a,M'}(\mu)-F_{a,M'}(y)||_{\text{small},W}$ by the definition of the small norm on $\mathcal{W}$. Using that $K_{a,M'}$ consists of all the points which $F_{a,M'}$ maps to zero, we have that $D_{\mathbb{R}^{\obs}}(\mu\to K_{a,M'})\leq||F_{a,M'}(\mu)||_{\text{small},W}$. Additionally, we have that $||w||_{\text{small},W}=R\cdot||w||_{\text{big},W}$, so we may upper-bound by
$$\leq 4\left(\frac{1}{\text{sine}}+1\right)R\bel{\max}{a}\bel{\max}{\mu\in M(a)}||F_{a,M'}(\mu)||_{W,\text{big}}$$

Now, because $\mu\in M(a)$, $F_{a,M}(\mu)=0$. Therefore, with linearity of $F$, the definitions of our norms on $\mathcal{Z}$, and John's ellipsoid theorem, we get
$$=4\left(\frac{1}{\text{sine}}+1\right)R\bel{\max}{a}\bel{\max}{y\in M(a)}||F_{a,M'-M}(y)||_{W,\text{big}}$$
$$\leq 4\left(\frac{1}{\text{sine}}+1\right)R\bel{\max}{a}\bel{\max}{y:||y||_{\mathbb{R}^{\obs}}\leq 1}||F_{a,M'-M}(y)||_{W,\text{big}}$$
$$=4\left(\frac{1}{\text{sine}}+1\right)R||M'-M||_{Z}\leq 4\left(\frac{1}{\text{sine}}+1\right)R\sqrt{Z}||M'-M||_{Z,2}$$

And so, our overall bound, for any $M,M'$, is
$$\bel{\max}{a}\hells(M(a)\to M'(a))\leq 4\left(\frac{1}{\text{sine}}+1\right)R\sqrt{Z}||M'-M||_{Z,2}$$

This same bound applies if we switch $M$ and $M'$, so our desired upper bound on 
\\
$\max_{a}\hells(M(a),M'(a))$ follows. $\blacksquare$

\begin{lemm} \label{lem10}
An upper bound on the number of L2 balls of radius $\eps$ needed to cover the cube $[-1,1]^{Z}$ is $\left(\frac{\sqrt{Z}}{2\eps}+1\right)^{Z}$.
\end{lemm}

Cover the $[-1,1]^{Z}$ cube with a uniform grid that has one point evenly spaced every $\frac{2\eps}{\sqrt{Z}}$ distance. There are, at most, $\left(\frac{\sqrt{Z}}{2\eps}+1\right)^{Z}$-many grid points. To show that this cover is suitable, and every point is within an L2 distance of $\eps$ of a point on this grid, we can take an arbitrary point $x$ and round it off to the closest value on the grid, in every coordinate, to produce some $x'$. The difference in every coordinate is $\leq\frac{\eps}{\sqrt{Z}}$, and then we can compute
$$||x-x'||_{2}=\sqrt{\sum_{i=1}^{Z}(x_{i}-x'_{i})^{2}}\leq\sqrt{\sum_{i=1}^{Z}\left(\frac{\eps}{\sqrt{Z}}\right)^{2}}=\frac{\eps}{\sqrt{Z}}\sqrt{Z}=\eps$$

Thereby verifying that we have constructed an appropriately fine cover. $\blacksquare$

\begin{prop} \label{pro6}
For the robust linear bandit setting, $\mathcal{N}(\hypo,\eps)\leq\left(\frac{4\left(\frac{1}{\text{sine}}+1\right)RZ}{\eps^{2}}+1\right)^{Z}$
\end{prop}

By Lemma \ref{lem10}, there is a cover of $[-1,1]^{Z}$ (the unit ball in $\mathcal{Z}$ according to $L_{\infty}$ distance, with the orthonormal basis given by the $L_{2}$ norm on $\mathcal{Z}$) with $\left(\frac{4\left(\frac{1}{\text{sine}}+1\right)RZ}{\eps^{2}}+1\right)^{Z}$-many balls of radius $\frac{\eps^{2}}{8\left(\frac{1}{\text{sine}}+1\right)R\sqrt{Z}}$. This produces an $\eps$-cover of $\hypo$ with respect to Hellinger distance, by taking every ball with a nonempty intersection with $\hypo$, and picking an arbitrary point in it. Here is why.

Given any $M\in\hypo$, it is in some occupied L2 ball, which has had an arbitrary $M'\in\hypo$ selected from it. By the triangle inequality, $||M'-M||_{Z,2}\leq\frac{\eps^{2}}{4\left(\frac{1}{\text{sine}}+1\right)R\sqrt{Z}}$. Then, by Lemma 9, this yields 
$$\max_{a}\hells(M(a),M'(a))\leq\eps^{2}$$

So, every $M\in\hypo$ has some $M'$ in our finite covering set, where, for all actions, 
\\
$\eps^{2}\geq\max_{a}\hells(M(a),M'(a))$, which occurs iff $\eps\geq\max_{a}D_{H}(M(a),M'(a))$. Therefore, we have constructed an $\eps$-cover of our hypothesis space. $\blacksquare$
\section{Definitions for Robust Markov Decision Processes}
$H$ is the number of timesteps in an episode. $T$ is the number of episodes. $\states$ is a finite state space, and $S$ is its cardinality. $\acts$ is a finite space of actions, and $A$ is its cardinality. Timesteps, states, actions, and rewards are denoted by $h,s,a,r$. $s_{h},a_{h},r_{h}$ denote the state, action, and reward on timestep $h$.

To denote a set of timesteps, we use the notation $[H]$ to denote the set $\{1,2...H\}$, and $[H]_{0}$ to denote the set $\{0,1,...H\}$. $\mathbb{P}$ is used for probabilities of events, instead of expectations.

The space of policies $\Pi_{RNS}$ is defined as $[H]\times\states\to\Delta\acts$. Elements are denoted $\pi$. $\pi(h,s)$ denotes the probability distribution over actions that $\pi$ plays when it is in state $s$ on timestep $h$, so $\pi(h,s)(a)$ is the probability of action $a$. Given a selection $\sigma$ of an RMDP, defined in Definition 5, $\sigma\bowtie\pi$ is the corresponding distribution over trajectories produced by $\sigma$ interacting with $\pi$.
\\

We identify RMDP's with their transition kernels $\imdp$, of type $[H]_{0}\times\states\times\acts\to\Box([0,1]\times\states)$. The initial state, initial action, and terminal state $s_{0},a_{0},s_{H+1}$ are considered to be unique, so trajectories are of the form $r_{0},s_{1},a_{1},r_{1},...,r_{H}$. We also identify RMDP's with their imprecise models $\pi\mapsto\Set{\sigma\bowtie\pi}{\sigma\models\imdp}$. $\sigma\models\imdp$ is defined in Definition 5.
\\

Given a distribution over trajectories $\mu$, $\mu|h,s,a$ is an abbreviation for $\mu|s_{h}=s,a_{h}=a$. Similarly, taking the probability of $h,s,a$ is an abbreviation for the probability of $s_{h}=s,a_{h}=a$. The same pattern generalizes to conditioning on, or asking for the probability of, $h,s$. $\mu_{\downarrow h}$ denotes this distribution projected to the $r_{h},s_{h+1}$ coordinates.
\\

$\imdp$ is 1-bounded if there is a selection $\sigma\models\imdp$ where, for all $\pi,h,s,a$, $\bel{\expec}{\sigma\bowtie\pi|h,s,a}\left[\sum_{k=h}^{H}r_{k}\right]\leq 1$.
For a fixed $H,\states,\acts$, $\hypo$ denotes the set of all 1-bounded RMDP's.

A function $\gen$ of type $\Pi_{RNS}\to\Delta\left([0,1]\times(\states\times\acts\times[0,1])^{[H]}\right)$ is policy-coherent if it fulfills the following property. For all policies, partial trajectories that end in a state $s$, and actions, the probability of $a$ according to $\gen(\pi)$ conditioned on the trajectory equals $\pi(h,s)(a)$. Intuitively, $\gen$ is policy-coherent if $\gen(\pi)$ is always a distribution over trajectories that looks like the actions were generated by the policy $\pi$.
\\

There is a natural function of type $[0,1]\to\Delta\{0,1\}$, which induces an affine function $\text{convert}:\Delta([0,1]\times\states)\to\Delta(\{0,1\}\times\states)$, which induces a function $\text{convert}:\Box([0,1]\times\states)\to\Box(\{0,1\}\times\states)$. $\hellsc(\Psi\to\Phi)$ is an abbreviation for $\hells(\text{convert}(\Psi)\to\text{convert}(\Phi))$.
\\

For the RMDP proofs, our notion of loss $\mathcal{L}(\gen,\imdp,\pi)$ is
$$\expec_{\gen(\pi)}\left[\sum_{h=0}^{H}\hellsc\left((\gen(\pi)|h,s_{h},a_{h})_{\downarrow h}\to\imdp(h,s_{h},a_{h})\right)\right]$$

$\hellsc$ measures squared Hellinger error, after converting distributions over $[0,1]$ to distributions over $\{0,1\}$. $\gen(\pi)$ is a distribution over trajectories, so we have to condition it on arriving at a specific state and action on timestep $h$, and project out the next reward $r_{h}$ and next state $s_{h+1}$, to find its prediction for what happens next. $\imdp(h,s,a)$ is just the transition kernel for our RMDP of choice. The modified offset DEC and modified fuzzy DEC are defined by taking the definitions of the offset DEC and fuzzy DEC, and swapping out Hellinger-squared loss for the above notion of loss.
\\

The function $\text{odec}(\gamma,p,M,N):\mathbb{R}^{\geq 0}\times\Delta\acts\times(\acts\to[0,2])\times(\acts\to[0,2])\to(-\infty,1]$ is defined as
$$\text{odec}(\gamma,p,M,N):=\bel{\max}{a'}N(a')-\bel{\expec}{a\sim p}[M(a)]-\gamma\bel{\expec}{a\sim p}\left[\left(N(a)-M(a)\right)^{2}\right]$$

Given a policy-coherent belief $\gen$, a policy $\pi$, and a $\gamma\geq 0$, $b^{\gen(\pi)}_{h,s}$ (the reward bonus at $h,s$), $\bonval^{\gen(\pi)}_{h,s}$ (the expected future reward plus bonuses at $h,s$, but clipped to be 1 at most) and $\overline{Q}^{\gen(\pi)}_{h,s,a}$ (the Q-value of $\bonval^{\gen(\pi)}$ at $h,s,a$) are defined by downwards induction as follows. $\overline{V}^{\gen(\pi)}_{H+1,s_{H+1}}$ is taken to be zero. Then, we have
$$\overline{Q}^{\gen(\pi)}_{h,s,a}=\bel{\expec}{\gen(\pi)|h,s,a}\left[r_{h}+\bonval^{\gen(\pi)}_{h+1,s_{h+1}}\right]$$
$$b^{\gen(\pi)}_{h,s}:=\bel{\min}{p}\ \bel{\max}{N}\ \text{odec}\left(\frac{\gamma}{8}\mathbb{P}_{\gen(\pi)}(h,s),p,\lambda a.\overline{Q}^{\gen(\pi)}_{h,s,a},N\right)$$
$$\bonval^{\gen(\pi)}_{h,s}:=\min\left(1,b^{\gen(\pi)}_{h,s}+\expec_{a\sim\pi(h,s)}\left[\overline{Q}^{\gen(\pi)}_{h,s,a}\right]\right)$$

This is well-defined for all $h,s$ and $h,s,a$ which $\gen(\pi)$ does not assign zero probability to. The definitions may be extended to zero with our convention that $s_{0}$ is an initial state and $a_{0}$ is an initial action. $\overline{Q}^{\gen(\pi)}_{0,s_{0},a_{0}}$ reduces to $\expec_{\gen(\pi)}\left[r_{0}+\bonval^{\gen(\pi)}_{1,s_{1}}\right]$. $b^{\gen(\pi)}_{0,s_{0}}$ is rendered well-defined and nonzero by looking at the value of the odec function for the one-armed bandit, instead of the $A$-armed bandit. Finally, we have $\bonval^{\gen(\pi)}_{0,s_{0}}=\min\left(1,b^{\gen(\pi)}_{h,s}+\overline{Q}^{\gen(\pi)}_{0,s_{0},a_{0}}\right)$.
\\

$\Delta_{\geq\eps}\acts$ denotes the space of all probability distributions on $\acts$ such that, for every action, the probability of that action is $\geq\frac{\eps}{A}$. $\textbf{1}$ is used to denote the indicator function which is 1 if an event happens and 0 if it doesn't.
\section{RMDP Lemmas}
\begin{lemm} \label{lem11}
Given two compact Polish spaces $A,B$, and a continuous function $f:A\times B\to\mathbb{R}$, the function $\lambda b.\bel{\min}{a}f(a,b)$ is continuous, as is $\lambda b.\bel{\max}{a}f(a,b)$.
\end{lemm}

By the definition of "Polish space", we can equip $A,B$ with metrics $D_{A},D_{B}$, to make them into compact metric spaces. The product of compact metric spaces can be made into a compact metric space via a metric of $D((a,a'),(b,b'))=\max(D_{A}(a,a'),D_{B}(b,b'))$. We then establish continuity of $\lambda b.\bel{\max}{a}f(a,b)$ as follows. If $b_{n}$ limits to $b$, then the functions $\lambda a.f(a,b_{n})$ limit (in the sup-norm) to $\lambda a.f(a,b)$, by
$$\limsup_{n\to\infty}\bel{\max}{a}|f(a,b_{n})-f(a,b)|\leq\limsup_{n\to\infty}\bel{\max}{a,b':D((a,b'),(a,b))\leq D_{B}(b_{n},b)}|f(a,b')-f(a,b)|=0$$

The inequality is a simple consequence of our definitions of distance. The equality is because, by the Heine-Cantor theorem, $f$ is uniformly continuous. For any $\eps>0$, uniform continuity of $f$ yields a $\delta$ where points being $\delta$ apart implies that $f$ can only vary by $\eps$ between those two points, and as $n$ increases, $b_{n}$ will eventually stay only $\delta$ apart from $b$, witnessing that the limsup is $\eps$ or less. This argument works for any $\eps$, so the limsup is zero. 

Therefore, $\lambda a.f(a,b_{n})$ limits to $\lambda a.f(a,b)$ in the sup-norm, so the difference between their maximum values limit to zero, and we have that $\bel{\max}{a}f(a,b_{n})$ converges to $\bel{\max}{a}f(a,b)$. This argument works for any convergent sequence $b_{n}$, so $\lambda b.\bel{\max}{a}f(a,b)$ is a continuous function. The same line of argument works for min as well. $\blacksquare$

\begin{lemm} \label{lem12}
Given three compact Polish spaces $X,Y,Z$, and a continuous function 
\\
$f:X\times Y\times Z\to\mathbb{R}$, the function $\lambda z.\bel{\min}{x}\ \bel{\max}{y}\ f(x,y,z)$ is continuous.
\end{lemm}

By assumption, $\lambda x,y,z.f(x,y,z)$ is continuous and everything is a compact Polish space. A product of two compact Polish spaces is compact Polish. Invoking Lemma \ref{lem11} with $A=Y$ and $B=X\times Z$ we get that $\lambda x,z.\bel{\max}{y}f(x,y,z)$ is continuous. Invoking Lemma \ref{lem11} a second time with $A=X$ and $B=Z$, we get that $\lambda z.\bel{\min}{x}\ \bel{\max}{y}\ f(x,y,z)$ is continuous and Lemma 12 is proven. $\blacksquare$
\\
\\

For Lemma 13, define the following construction. Given a compact Polish space $X$, $X^{\dagger}$ denotes the quotient of the space $X\times[0,1]$ formed by identifying all $(x,0)$ pairs.

\begin{lemm} \label{lem13}
If a policy-coherent belief $\gen:\Pi_{RNS}\to\Delta([0,1]\times(\states\times\acts\times[0,1])^{[H]})$ is continuous, then for all $h,s,a$ and $\eps>0$, the following functions of type signature
\\
$\left([H]\times\states\to\Delta_{\geq\eps}\acts\right)\to[0,2]^{\dagger}$ are continuous. The Q-value function $\lambda\pi.\left(\overline{Q}^{\gen(\pi)}_{h,s,a},\mathbb{P}_{\gen(\pi)}(h,s)\right)$, the bonus function
\\
$\lambda\pi.\left(b^{\gen(\pi)}_{h,s},\mathbb{P}_{\gen(\pi)}(h,s)\right)$ and the expected value function $\lambda\pi.\left(\bonval^{\gen(\pi)}_{h,s},\mathbb{P}_{\gen(\pi)}(h,s)\right)$
\end{lemm}

To establish this, we use a downwards induction proof, where we assume continuity of these functions at level $h+1$, and derive their continuity at level $h$. To show the base case that continuity holds at $h=H+1$, we use that the expected value $\bonval^{\gen(\pi)}_{H+1,s}$ is always zero by definition and does not depend on the choice of $\pi$. The probability of $H+1,s$ is zero if $s$ isn't the terminal state, and one if it is, and again, there is no dependence on $\pi$. Constant functions are continuous, so our base case is proven.

For the induction step, we first notice that, for the various functions, by the definition of $[0,2]^{\dagger}$, continuity is proven by showing that, if $\pi_{n}$ converges to $\pi$, the probability term converges, and if the probability converges to $>0$, the values converge as well. 

Convergence of probabilities holds because, for any $h,s$, by the starting assumption of the continuity of $\gen$, it is immediate that $\mathbb{P}_{\gen(\pi_{n})}(h,s)$ converges to $\mathbb{P}_{\gen(\pi)}(h,s)$.
\\

Now we must show convergence of values if the limiting probability for $h,s$ is nonzero  so we may freely assume that $\mathbb{P}_{\gen(\pi)}(h,s)>0$.

For Q-values, we can expand the definition.
$$\lim_{n\to\infty}\overline{Q}^{\gen(\pi_{n})}_{h,s,a}=\lim_{n\to\infty}\bel{\expec}{\gen(\pi_{n})|h,s,a}\left[r_{h}+\bonval^{\gen(\pi_{n})}_{h+1,s_{h+1}}\right]$$
$$=\lim_{n\to\infty}\frac{\expec_{\gen(\pi_{n})}\left[\textbf{1}_{s_{h}=s}\cdot\textbf{1}_{a_{h}=a}\cdot\left(r_{h}+\bonval^{\gen(\pi_{n})}_{h+1,s_{h+1}}\right)\right]}{\expec_{\gen(\pi_{n})}\left[\textbf{1}_{s_{h}=s}\cdot\textbf{1}_{a_{h}=a}\right]}$$
$$=\lim_{n\to\infty}\frac{\expec_{\gen(\pi_{n})}\left[\textbf{1}_{s_{h}=s}\cdot\textbf{1}_{a_{h}=a}\cdot\left(r_{h}+\bonval^{\gen(\pi_{n})}_{h+1,s_{h+1}}\right)\right]}{\mathbb{P}_{\gen(\pi_{n})}(h,s)\cdot\pi_{n}(h,s)(a)}$$

The quantity in the denominator clearly converges to $\mathbb{P}_{\gen(\pi)}(h,s)\cdot\pi(h,s)(a)$. This limiting value is nonzero, because $\mathbb{P}_{\gen(\pi)}(h,s)>0$ by assumption, and $\pi(h,s)(a)\geq\frac{\eps}{A}$ because we are restricting our attention to policies of type $[H]\times\states\to\Delta_{\geq\eps}\acts$.
$$=\lim_{n\to\infty}\frac{\expec_{\gen(\pi_{n})}\left[\textbf{1}_{s_{h}=s}\cdot\textbf{1}_{a_{h}=a}\cdot\left(r_{h}+\bonval^{\gen(\pi_{n})}_{h+1,s_{h+1}}\right)\right]}{\mathbb{P}_{\gen(\pi)}(h,s)\cdot\pi(h,s)(a)}$$

Now, as $\pi_{n}$ limits to $\pi$, for any given $s'$, by our induction assumption of continuity at $h+1$, either $\bonval^{\gen(\pi_{n})}_{h+1,s'}$ limits to $\bonval^{\gen(\pi)}_{h+1,s'}$, or the probability of $s'$ at time $h+1$ limits to zero. If we have convergence of values only holding for states with nonzero limiting probability, the expectation values will converge anyways, and we have
$$=\lim_{n\to\infty}\frac{\expec_{\gen(\pi_{n})}\left[\textbf{1}_{s_{h}=s}\cdot\textbf{1}_{a_{h}=a}\cdot\left(r_{h}+\bonval^{\gen(\pi)}_{h+1,s_{h+1}}\right)\right]}{\mathbb{P}_{\gen(\pi)}(h,s)\cdot\pi(h,s)(a)}$$

Now, by continuity of $\gen$, $\gen(\pi_{n})$ converges to $\gen(\pi)$. The contents of the expectation in the numerator are continuous as a function from trajectories to $\mathbb{R}$, so their expectation values converge, and we have
$$=\frac{\expec_{\gen(\pi)}\left[\textbf{1}_{s_{h}=s}\cdot\textbf{1}_{a_{h}=a}\cdot\left(r_{h}+\bonval^{\gen(\pi)}_{h+1,s_{h+1}}\right)\right]}{\mathbb{P}_{\gen(\pi)}(h,s)\cdot\pi(h,s)(a)}=\bel{\expec}{\gen(\pi)|h,s,a}\left[r_{h}+\bonval^{\gen(\pi)}_{h+1,s_{h+1}}\right]=\overline{Q}^{\gen(\pi)}_{h,s,a}$$

And we have shown continuity for the Q-values for $h$ and all $s,a$.
\\

To show continuity of the bonus values for $h$ and all $s$, we use continuity of the Q-values for $h$ and all $s,a$, and can freely assume that $\mathbb{P}_{\gen(\pi)}(h,s)>0$. By definition, we have
$$\lim_{n\to\infty}b^{\gen(\pi_{n})}_{h,s}=\lim_{n\to\infty}\bel{\min}{p}\ \bel{\max}{N}\ \text{odec}\left(\frac{\gamma}{8}\mathbb{P}_{\gen(\pi_{n})}(h,s),p,\lambda a.\overline{Q}^{\gen(\pi_{n})}_{h,s,a},N\right)$$

The odec function can be verified, by casual inspection of its definition, to be continuous in all arguments. All parameters come from compact Polish spaces, namely $[0,\frac{\gamma}{8}]$, $\Delta\acts$, $\acts\to[0,2]$, and $\acts\to[0,2]$. Therefore, we can invoke Lemma \ref{lem12} to get
$$=\bel{\min}{p}\ \bel{\max}{N}\ \text{odec}\left(\lim_{n\to\infty}\frac{\gamma}{8}\mathbb{P}_{\gen(\pi_{n})}(h,s),p,\lim_{n\to\infty}\left(\lambda a.\overline{Q}^{\gen(\pi_{n})}_{h,s,a}\right),N\right)$$

The probabilities converge, as previously argued, and we have already proven that the Q-values converge, as long as the limiting probability of $h,s$ is above zero, which it is assumed to be, so we have
$$=\bel{\min}{p}\ \bel{\max}{N}\ \text{odec}\left(\frac{\gamma}{8}\mathbb{P}_{\gen(\pi)}(h,s),p,\lambda a.\overline{Q}^{\gen(\pi)}_{h,s,a},N\right)=b^{\gen(\pi)}_{h,s}$$

And continuity is shown for the bonus values. 
\\

Now, to show continuity for the expected values we have
$$\lim_{n\to\infty}\bonval^{\gen(\pi_{n})}_{h,s}=\lim_{n\to\infty}\min\left(1,b^{\gen(\pi_{n})}_{h,s}+\bel{\expec}{a\sim\pi_{n}(h,s)}\left[\overline{Q}^{\gen(\pi_{n})}_{h,s,a}\right]\right)$$

By continuity of the bonus values and Q-values (which we have just proven in our induction step) and our assumption that $h,s$ has a nonzero limiting probability, the expected values converge.
$$=\min\left(1,b^{\gen(\pi)}_{h,s}+\bel{\expec}{a\sim\pi(h,s)}\left[\overline{Q}^{\gen(\pi)}_{h,s,a}\right]\right)=\bonval^{\gen(\pi)}_{h,s}$$

And so, continuity is established for the expected values. The downwards induction step has been proven, so Lemma 13 follows, that for all $h,s,a$, the Q-values, bonus values, and expected values are continuous as a function of $\pi$. $\blacksquare$

\begin{lemm} \label{lem14}
If a policy-coherent belief $\gen:\Pi_{RNS}\to\Delta([0,1]\times(\states\times\acts\times[0,1])^{[H]})$ is continuous, then for any $\eps>0,\gamma>0$, there exists a policy $\poli\in\Pi_{RNS}$, where, for all $h,s\in[H]_{0}\times\states$ that $\gen(\poli)$ assigns nonzero probability,
$$b^{\gen(\poli)}_{h,s}+\left(2+\frac{\gamma}{2}\right)\eps$$
$$\geq\bel{\max}{N\in\acts\to[0,2]}\left(\bel{\max}{a'}N(a')-\bel{\expec}{a\sim\poli(h,s)}\left[\overline{Q}^{\gen(\poli)}_{h,s,a}\right]-\frac{\gamma}{8}\mathbb{P}_{\gen(\poli)}(h,s)\bel{\expec}{a\sim\poli(h,s)}\left[\left(N(a)-\overline{Q}^{\gen(\poli)}_{h,s,a}\right)^{2}\right]\right)$$
\end{lemm}

As in the definitions, $\Delta_{\geq\eps}\acts$ is the set of all probability distributions which assign $\geq\frac{\eps}{A}$ probability to every action. Using the definitions, and letting $u_{\acts}$ be the uniform distribution on actions, consider the following set-valued function $([H]\times\states\to\Delta_{\geq\eps}\acts)\to\mathcal{P}([H]\times\states\to\Delta_{\geq\eps}\acts)$. $\Pi_{h,s\in[H]\times\states}$ denotes the Cartesian product of sets in this formula.
\begin{equation} \label{eq:2}
\lambda\pi.\Pi_{h,s\in[H]\times\states}\text{ if }\mathbb{P}_{\gen(\pi)}(h,s)=0,\Set{\eps u_{\acts}+(1-\eps)\nu}{\nu\in\Delta\acts}
\end{equation}
$$\text{else}\Set{\eps u_{\acts}+(1-\eps)\nu}{\nu\in\bel{\text{argmin}}{p}\ \bel{\max}{N}\ \text{odec}\left(\frac{\gamma}{8}\mathbb{P}_{\gen(\pi)}(h,s),p,\lambda a.\overline{Q}^{\gen(\pi)}_{h,s,a},N\right)}$$

First, we will show that any fixpoint of this set-valued function has our desired property. Then we will show that, by Kakutani's fixpoint theorem, a suitable fixpoint exists.

Accordingly, let $\poli$ be a fixpoint of the above equation. Then, for any $h,s$ with nonzero probability according to $\gen(\poli)$, $\poli(h,s)$ is of the form $\eps u_{\acts}+(1-\eps)\nu$ where $\nu$ is the distribution over actions which minimizes the odec function.
$$\bel{\max}{N\in\acts\to[0,2]}\bel{\max}{a'}N(a')-\bel{\expec}{a\sim\poli(h,s)}\left[\overline{Q}^{\gen(\poli)}_{h,s,a}\right]-\frac{\gamma}{8}\mathbb{P}_{\gen(\poli)}(h,s)\bel{\expec}{a\sim\poli(h,s)}\left[\left(N(a)-\overline{Q}^{\gen(\poli)}_{h,s,a}\right)^{2}\right]$$
$$=\bel{\max}{N\in\acts\to[0,2]}\bel{\max}{a'}N(a')-\bel{\expec}{a\sim\eps u_{\acts}+(1-\eps)\nu}\left[\overline{Q}^{\gen(\poli)}_{h,s,a}\right]$$
$$-\frac{\gamma}{8}\mathbb{P}_{\gen(\poli)}(h,s)\bel{\expec}{a\sim\eps u_{\acts}+(1-\eps)\nu}\left[\left(N(a)-\overline{Q}^{\gen(\poli)}_{h,s,a}\right)^{2}\right]$$

Because these functions are bounded in $[0,2]$ (for the Q values, it is because reward is bounded in $[0,1]$ and so is $\bonval$), we can upper-bound by
$$\leq\bel{\max}{N\in\acts\to[0,2]}\bel{\max}{a'}N(a')-\bel{\expec}{a\sim\nu}\left[\overline{Q}^{\gen(\poli)}_{h,s,a}\right]+2D_{TV}(\eps u_{A}+(1-\eps)\nu,\nu)$$
$$-\frac{\gamma}{8}\mathbb{P}_{\gen(\poli)}(h,s)\left(\bel{\expec}{a\sim\nu}\left[\left(N(a)-\overline{Q}^{\gen(\poli)}_{h,s,a}\right)^{2}\right]-4D_{TV}(\eps u_{A}+(1-\eps)\nu,\nu)\right)$$

The total variation terms are both upper-bounded by $\eps$, which can be pulled out of the maximum. Upper-bounding the probability term by 1, we get
$$\leq\left(2+\frac{\gamma}{2}\right)\eps+\bel{\max}{N\in\acts\to[0,2]}\bel{\max}{a'}N(a')-\bel{\expec}{a\sim\nu}\left[\overline{Q}^{\gen(\poli)}_{h,s,a}\right]-\frac{\gamma}{8}\mathbb{P}_{\gen(\poli)}(h,s)\left(\bel{\expec}{a\sim\nu}\left[\left(N(a)-\overline{Q}^{\gen(\poli)}_{h,s,a}\right)^{2}\right]\right)$$

The above equation can be rephrased as
$$=\left(2+\frac{\gamma}{2}\right)\eps+\bel{\max}{N\in\acts\to[0,2]}\text{odec}\left(\frac{\gamma}{8}\mathbb{P}_{\gen(\poli)}(h,s),\nu,\lambda a.\overline{Q}^{\gen(\poli)}_{h,s,a},N\right)$$

Using that $\nu$ is the minimizer of the above equation, we can rephrase, and pack up the definition of $b^{\gen(\poli)}_{h,s}$.
$$=\left(2+\frac{\gamma}{2}\right)\eps+\bel{\min}{p}\bel{\max}{N\in\acts\to[0,2]}\text{odec}\left(\frac{\gamma}{8}\mathbb{P}_{\gen(\poli)}(h,s),p,\lambda a.\overline{Q}^{\gen(\poli)}_{h,s,a},N\right)=\left(2+\frac{\gamma}{2}\right)\eps+b^{\gen(\poli)}_{h,s}$$

Putting all inequalities together, we have proven our desired property, as long as there is a $\poli$ which is a fixpoint of \ref{eq:2}. 
\\

We now shift to proving that a fixpoint exists. The topological conditions for Kakutani's fixpoint theorem are easy to verify. $[H]\times\states\to\Delta_{\geq\eps}\acts$ is clearly a compact, convex, nonempty subset of a Euclidean space. That just leaves showing nonemptiness, convexity, and closed graph of the set-valued function.
\\

To show nonemptiness, the product of nonempty sets is nonempty, so it suffices to prove nonemptiness for all component sets of the product. If $\mathbb{P}_{\gen(\pi)}(h,s)=0$, nonemptiness is trivial. If it exceeds zero, then Lemma \ref{lem13} shows the Q-values are well-defined, and then the arguments in Lemma \ref{lem12}, along with the continuity of the odec function, prove that the function
\\
$\lambda p.\bel{\max}{N}\ \text{odec}\left(\frac{\gamma}{8}\mathbb{P}_{\gen(\pi)}(h,s),p,\lambda a.\overline{Q}^{\gen(\pi)}_{h,s,a},N\right)$
is continuous. The distribution $p$ is from $\Delta\acts$ which is compact, and continuous functions from compact spaces to $\mathbb{R}$ have minimizers, so the set of minimizers is nonempty. And so, nonemptiness has been shown.
\\

To show convexity, the product of convex sets is convex, so it suffices to prove convexity for all the component sets of the product. Given a convex set of probability distributions $C$ mixing $\eps$ of the uniform distribution with all distributions in $C$ preserves convexity, so we just need to show convexity holds before mixing with the uniform distribution. If $\mathbb{P}_{\gen(\pi)}(h,s)=0$, this is trivial, because $\Delta\acts$ is convex. If it exceeds zero, then Lemma \ref{lem13} shows the Q-values are well-defined, and then, letting $\nu_{1}$ and $\nu_{2}$ be minimizers, and $q\in[0,1]$, we have, by the odec function being affine in the distribution over actions, convexity of maximization, and the defining property of $\nu_{1}$ and $\nu_{2}$,
$$\bel{\max}{N}\ \text{odec}\left(\frac{\gamma}{8}\mathbb{P}_{\gen(\pi)}(h,s),q\nu_{1}+(1-q)\nu_{2},\lambda a.\overline{Q}^{\gen(\pi)}_{h,s,a},N\right)$$
$$=\bel{\max}{N}(q\cdot\text{odec}\left(\frac{\gamma}{8}\mathbb{P}_{\gen(\pi)}(h,s),\nu_{1},\lambda a.\overline{Q}^{\gen(\pi)}_{h,s,a},N\right)$$
$$+(1-q)\text{odec}\left(\frac{\gamma}{8}\mathbb{P}_{\gen(\pi)}(h,s),\nu_{2},\lambda a.\overline{Q}^{\gen(\pi)}_{h,s,a},N\right))$$
$$\leq q\ \bel{\max}{N}\ \text{odec}\left(\frac{\gamma}{8}\mathbb{P}_{\gen(\pi)}(h,s),\nu_{1},\lambda a.\overline{Q}^{\gen(\pi)}_{h,s,a},N\right)$$
$$+(1-q)\bel{\max}{N}\ \text{odec}\left(\frac{\gamma}{8}\mathbb{P}_{\gen(\pi)}(h,s),\nu_{2},\lambda a.\overline{Q}^{\gen(\pi)}_{h,s,a},N\right)$$
$$=\bel{\min}{p\in\Delta\acts}\ \bel{\max}{N}\ \text{odec}\left(\frac{\gamma}{8}\mathbb{P}_{\gen(\pi)}(h,s),p,\lambda a.\overline{Q}^{\gen(\pi)}_{h,s,a},N\right)$$

However, this inequality cannot be strict because otherwise $q\nu_{1}+(1-q)\nu_{2}$ would yield a lower value than the minimum possible, so it must be an equality, witnessing that $q\nu_{1}+(1-q)\nu_{2}$ is a minimizer as well. $q$ and the choices of $\nu_{1}$ and $\nu_{2}$ were arbitrary, so any mixture of minimizers is a minimizer, and the argmin set is convex. Convexity is thereby established.
\\

To show closed graph, let $\pi_{n}$ limit to $\pi$ and assume that $\pi'_{n}$ are in the sets associated with $\pi_{n}$, and that they limit to $\pi'$. Showing that the limiting $\pi'$ lies in the set associated with $\pi$ verifies the closed graph property. This occurs iff, for all $h,s$, we have 
$$\text{ if }\mathbb{P}_{\gen(\pi)}(h,s)=0,\pi'(h,s)\in\Set{\eps u_{\acts}+(1-\eps)\nu}{\nu\in\Delta\acts}$$
$$\text{else }\pi'(h,s)\in\Set{\eps u_{\acts}+(1-\eps)\nu}{\nu\in\bel{\text{argmin}}{p}\ \bel{\max}{N}\ \text{odec}\left(\frac{\gamma}{8}\mathbb{P}_{\gen(\pi)}(h,s),p,\lambda a.\overline{Q}^{\gen(\pi)}_{h,s,a},N\right)}$$

Now, given $\pi'(h,s)$, it is possible to uniquely recover the corresponding distribution $\nu$ via $\frac{\pi'(h,s)-\eps u_{\acts}}{(1-\eps)}$. This is a continuous function of the policy, so from the $\pi'_{n}(h,s)$ sequence, we get a sequence $\nu_{n}$ limiting to some $\nu$, where, for all $n$, $\pi'_{n}(h,s)=\eps u_{\acts}+(1-\eps)\nu_{n}$, and the same holds for the limiting $\pi'$ and $\nu$. Therefore, we can show that $\pi'(h,s)$ lies in the needed set by showing that the limiting $\nu$ fulfills the appropriate condition. Proving this for arbitrary $h,s$ verifies the closed graph property, so let $h,s$ be arbitrary.

If $\mathbb{P}_{\gen(\pi)}(h,s)=0$, we only need to verify that the limiting $\nu$ lies in $\Delta\acts$, which is trivial. If this quantity exceeds zero, we must verify that
$$\bel{\max}{N}\ \text{odec}\left(\frac{\gamma}{8}\mathbb{P}_{\gen(\pi)}(h,s),\nu,\lambda a.\overline{Q}^{\gen(\pi)}_{h,s,a},N\right)=\bel{\min}{p}\ \bel{\max}{N}\ \text{odec}\left(\frac{\gamma}{8}\mathbb{P}_{\gen(\pi)}(h,s),p,\lambda a.\overline{Q}^{\gen(\pi)}_{h,s,a},N\right)$$

This equality can be verified as follows. By Lemma \ref{lem13}, and $\nu$ being the limit of the $\nu_{n}$, we have
$$\bel{\max}{N}\ \text{odec}\left(\frac{\gamma}{8}\mathbb{P}_{\gen(\pi)}(h,s),\nu,\lambda a.\overline{Q}^{\gen(\pi)}_{h,s,a},N\right)$$
$$=\bel{\max}{N}\ \text{odec}\left(\lim_{n\to\infty}\frac{\gamma}{8}\mathbb{P}_{\gen(\pi_{n})}(h,s),\lim_{n\to\infty}\nu_{n},\lim_{n\to\infty}\lambda a.\overline{Q}^{\gen(\pi_{n})}_{h,s,a},N\right)$$

By Lemma \ref{lem11}, and continuity of the odec function, we can show that the function 
\\
$\lambda\gamma,\nu,M.\bel{\max}{N}\ \text{dec}(\gamma,\nu,M,N)$ is continuous, yielding
$$=\lim_{n\to\infty}\bel{\max}{N}\ \text{odec}\left(\frac{\gamma}{8}\mathbb{P}_{\gen(\pi_{n})}(h,s),\nu_{n},\lambda a.\overline{Q}^{\gen(\pi_{n})}_{h,s,a},N\right)$$

Then since $\mathbb{P}_{\gen(\pi)}(h,s)>0$, all but finitely many of the $\pi_{n}$ fulfill $\mathbb{P}_{\gen(\pi_{n})}(h,s)>0$, so their corresponding $\nu_{n}$ fulfill 
$$\bel{\max}{N}\ \text{odec}\left(\frac{\gamma}{8}\mathbb{P}_{\gen(\pi_{n})}(h,s),\nu_{n},\lambda a.\overline{Q}^{\gen(\pi_{n})}_{h,s,a},N\right)$$
$$=\bel{\min}{p}\ \bel{\max}{N}\ \text{odec}\left(\frac{\gamma}{8}\mathbb{P}_{\gen(\pi_{n})}(h,s),p,\lambda a.\overline{Q}^{\gen(\pi_{n})}_{h,s,a},N\right)$$

This yields 
$$=\lim_{n\to\infty}\bel{\min}{p}\ \bel{\max}{N}\ \text{odec}\left(\frac{\gamma}{8}\mathbb{P}_{\gen(\pi_{n})}(h,s),p,\lambda a.\overline{Q}^{\gen(\pi_{n})}_{h,s,a},N\right)$$

Then, by Lemma \ref{lem12} and the continuity of the odec function, along with Lemma \ref{lem13} witnessing that the Q-values and probabilities are continuous, we have
$$=\bel{\min}{p}\ \bel{\max}{N}\ \text{odec}\left(\lim_{n\to\infty}\frac{\gamma}{8}\mathbb{P}_{\gen(\pi_{n})}(h,s),p,\lim_{n\to\infty}\lambda a.\overline{Q}^{\gen(\pi_{n})}_{h,s,a},N\right)$$
$$=\bel{\min}{p}\ \bel{\max}{N}\ \text{odec}\left(\lim_{n\to\infty}\frac{\gamma}{8}\mathbb{P}_{\gen(\pi)}(h,s),p,\lim_{n\to\infty}\lambda a.\overline{Q}^{\gen(\pi)}_{h,s,a},N\right)$$

Our necessary equality, to show that the limiting $\nu$ is a minimizer, has been verified. This establishes that $\pi'(h,s)$ lies in the appropriate set. $h,s$ were arbitrary, and so we have established closed graph for our set-valued function.

As nonemptiness, convexity, and closed graph have been verified, a fixpoint exists by Kakutani's Fixpoint Theorem, and our result follows. $\blacksquare$
\section{RMDP Theorems}
\begin{lemm} \label{lem15}
In the episodic RMDP setting, if $\gen:\Pi_{RNS}\to\Delta([0,1]\times(\states\times\acts\times[0,1])^{H})$ is continuous and policy-coherent, the modified offset DEC fulfills $\text{dec}^{o'}_{\gamma}(\hypo,\gen)\leq\frac{2(HSA+1)}{\gamma}$
\end{lemm}

First, we unpack the definition of the modified offset DEC, and distribute the expectations over actions inwards, using policy-coherence of $\gen$.
$$\bel{\min}{p\in\Delta\Pi_{RNM}}\bel{\max}{\imdp\in\hypo}\max(f^{\imdp})-\bel{\expec}{\pi\sim p}\left[f^{\gen}(\pi)\right]$$
$$-\gamma\bel{\expec}{\pi\sim p}\left[\expec_{\gen(\pi)}\left[\sum_{h=0}^{H}\bel{\expec}{a\sim\pi(h,s_{h})}\left[\hellsc\left(\left(\gen(\pi)|h,s_{h},a\right)_{\downarrow h}\to\imdp(h,s_{h},a)\right)\right]\right]\right]$$

Because $\gen$ is assumed to be continuous, we can invoke Lemma \ref{lem14} for an arbitrarily low $\eps$ and the $\gamma$ of choice to construct a suitable policy $\poli$. Let $p$ be the deterministic choice of $\poli$.
$$\leq\bel{\max}{\imdp\in\hypo}\max(f^{\imdp})-f^{\gen}(\poli)-\gamma\expec_{\gen(\poli)}\left[\sum_{h=0}^{H}\bel{\expec}{a\sim\poli(h,s_{h})}\left[\hellsc\left(\left(\gen(\poli)|h,s_{h},a\right)_{\downarrow h}\to\imdp(h,s_{h},a)\right)\right]\right]$$

Fix the optimal choice of $\imdp$. Let $\pi_{\imdp}$ be the optimal choice of policy to maximize worst-case expected reward. We can now reexpress the reward terms.
\begin{equation} \label{eq:3}
=\bel{\min}{\sigma\models\imdp}\expec_{\sigma\bowtie\pi_{\imdp}}\left[\sum_{h=0}^{H}r_{h}\right]-\expec_{\gen(\poli)}\left[\sum_{h=0}^{H}r_{h}\right]
\end{equation}
$$-\gamma\expec_{\gen(\poli)}\left[\sum_{h=0}^{H}\bel{\expec}{a\sim\poli(h,s_{h})}\left[\hellsc\left(\left(\gen(\poli)|h,s_{h},a\right)_{\downarrow h}\to\imdp(h,s_{h},a)\right)\right]\right]$$

We now define a new term. Fix the convention that $\bonval^{\gen(\poli)}_{h,s}=1$ if $\mathbb{P}_{\gen(\poli)}(h,s)=0$, rendering $\bonval^{\gen(\poli)}_{h,s}$ well-defined everywhere. Call $h,s$ as "bad state" if $\bonval^{\gen(\poli)}_{h,s}=1$. Given a trajectory $s_{1},a_{1},r_{1}...s_{H},a_{H},r_{H}$, define $t^{*}$ (the dependence on trajectory is suppressed in the notation) as
$$t^{*}:=\max\Set{h}{0\leq h\leq H+1\wedge\forall0\leq k<h:\bonval^{\gen(\poli)}_{k,s_{k}}<1}$$

Intuitively, $t^{*}$ is the first timestep where we encounter a bad state, and is $H+1$ otherwise.
\\

To digress, letting $\sigma^{*}$ be $$\sigma^{*}:=\bel{\text{argmin}}{\sigma\models\imdp}\ \expec_{\sigma\bowtie\pi_{\imdp}}\left[\sum_{h=0}^{t^{*}-1}r_{h}+\textbf{1}_{t^{*}=h}\right]$$
we can modify $\sigma^{*}$ to construct a $\sigma'$ which is also a selection of $\imdp$ as follows. $\sigma'$ copies $\sigma^{*}$, except that if it ever hits a bad state, the transition probabilities from that point onward will behave in the way (consistent with $\imdp$) which minimizes the expected sum of future rewards. We have
$$\bel{\min}{\sigma\models\imdp}\bel{\expec}{\sigma\bowtie\pi_{\imdp}}\left[\sum_{h=0}^{H}r_{h}\right]\leq\bel{\expec}{\sigma'\bowtie\pi_{\imdp}}\left[\sum_{h=0}^{H}r_{h}\right]\leq\bel{\expec}{\sigma'\bowtie\pi_{\imdp}}\left[\sum_{h=0}^{t^{*}-1}r_{h}+\textbf{1}_{t^{*}=h}\right]$$
$$=\bel{\expec}{\sigma^{*}\bowtie\pi_{\imdp}}\left[\sum_{h=0}^{t^{*}-1}r_{h}+\textbf{1}_{t^{*}=h}\right]=\bel{\min}{\sigma\models\imdp}\bel{\expec}{\sigma\bowtie\pi_{\imdp}}\left[\sum_{h=0}^{t^{*}-1}r_{h}+\textbf{1}_{t^{*}=h}\right]$$

The first inequality and the last equality are obvious. The second inequality holds because one of our assumptions was that $\hypo$ consisted of the 1-bounded RMDP's, so for all $h,s,\pi$, there was a way that the RMDP could behave which would guarantee that the expected sum of future rewards was in $[0,1]$. Switching from "receive 1 reward when a bad state is reached" to "receive the worst-case sum of future rewards when a bad state is reached" can only decrease the expected sum of rewards. The first equality holds because $\nu'$ only behaves differently than $\nu$ after a bad state is reached, so if the reward only depends on the trajectory up to the first bad state, the expected rewards must be the same. Combining this inequality with \ref{eq:3}, we get
$$\leq\bel{\min}{\sigma\models\imdp}\bel{\expec}{\sigma\bowtie\pi_{\imdp}}\left[\sum_{h=0}^{t^{*}-1}r_{h}+\textbf{1}_{t^{*}=h}\right]-\expec_{\gen(\poli)}\left[\sum_{h=0}^{H}r_{h}\right]$$
$$-\gamma\expec_{\gen(\poli)}\left[\sum_{h=0}^{H}\bel{\expec}{a\sim\poli(h,s_{h})}\left[\hellsc\left((\gen(\poli)|h,s_{h},a)_{\downarrow h}\to\imdp(h,s_{h},a)\right)\right]\right]$$

Now, let $M$ be the Markov decision process which, in situation $h,s,a$, selects the distribution $\mu:\Delta([0,1]\times\states)$ from $\imdp(h,s,a)$ which minimizes $\hellsc\left(\gen(\poli)|h,s,a)_{\downarrow h},\mu\right)$. If the conditional is undefined, $M$ can select any distribution from $\imdp(h,s,a)$. This is a selection of $\imdp$, so it can only increase the expected reward. It also lets us rephrase the Hellinger error term. We then rephrase the $\gen(\poli)$-expected sum of rewards, to yield
$$\leq\bel{\expec}{M\bowtie\pi_{\imdp}}\left[\sum_{h=0}^{t^{*}-1}r_{h}+\textbf{1}_{t^{*}=h}\right]-\expec_{\gen(\poli)}\left[\sum_{h=0}^{H}r_{h}\right]$$
$$-\gamma\expec_{\gen(\poli)}\left[\sum_{h=0}^{H}\bel{\expec}{a\sim\poli(h,s_{h})}\left[\hellsc\left(\left(\gen(\poli)|h,s_{h},a\right)_{\downarrow h},M(h,s_{h},a)\right)\right]\right]$$

Then, creating some more terms, we have
\begin{equation} \label{eq:4}
=\bel{\expec}{M\bowtie\pi_{\imdp}}\left[\sum_{h=0}^{t^{*}-1}r_{h}+\textbf{1}_{t^{*}=h}\right]-\bonval^{\gen(\poli)}_{0,s_{0}}
\end{equation}
$$+\bonval^{\gen(\poli)}_{0,s_{0}}-\expec_{\gen(\poli)}\left[\sum_{h=0}^{H}r_{h}+b^{\gen(\poli)}_{h,s_{h}}\right]+\expec_{\gen(\poli)}\left[\sum_{h=0}^{H}b^{\gen(\poli)}_{h,s_{h}}\right]$$
$$-\gamma\expec_{\gen(\poli)}\left[\sum_{h=0}^{H}\bel{\expec}{a\sim\poli(h,s_{h})}\left[\hellsc\left(\left(\gen(\poli)|h,s_{h},a\right)_{\downarrow h},M(h,s_{h},a)\right)\right]\right]$$

Note that $s_{0}$ is the unique initial state. The terms in the second and third lines will be ignored until later, and we'll focus on the term in the first line. We can rewrite $\sum_{h=0}^{t^{*}-1}r_{h}$ as $\sum_{h=0}^{H}\textbf{1}_{t^{*}\geq h+1}\cdot r_{h}$, and pull the sum out of the expectation. We also have $\bonval^{\gen(\poli)}_{0,s_{0}}=\expec_{M\bowtie\pi_{\imdp}}\left[\textbf{1}_{t^{*}\geq 0}\cdot\bonval^{\gen(\poli)}_{0,s_{0}}\right]$ holding
because $t^{*}\geq 0$ always holds and expectations of constants are that same constant. $t^{*}\geq 0$ holds because 0 is always in the set that $t^{*}$ is the maximum of, due to the relevant for-all statement being vacuously true. Finally, we have
\\
$0=\expec_{M\bowtie\pi_{\imdp}}\left[\textbf{1}_{t^{*}\geq H+1}\cdot\bonval^{\gen(\poli)}_{H+1,s_{H+1}}\right]$
holding because $\bonval^{\gen(\poli)}_{H+1,s_{H+1}}$ is defined to be zero. Therefore, the first term we are focusing on can be rewritten as
$$\sum_{h=0}^{H}\left(\bel{\expec}{M\bowtie\pi_{\imdp}}\left[\textbf{1}_{t^{*}\geq h+1}\cdot r_{h}+\textbf{1}_{t^{*}=h}\right]\right)+\bel{\expec}{M\bowtie\pi_{\imdp}}\left[\textbf{1}_{t^{*}\geq H+1}\cdot\bonval^{\gen(\poli)}_{H+1,s_{H+1}}\right]-\bel{\expec}{M\bowtie\pi_{\imdp}}\left[\textbf{1}_{t^{*}\geq 0}\cdot\bonval^{\gen(\poli)}_{0,s_{0}}\right]$$

It can be further rewritten as a telescoping sum now.
$$=\sum_{h=0}^{H}\bel{\expec}{M\bowtie\pi_{\imdp}}\left[\textbf{1}_{t^{*}\geq h+1}\cdot r_{h}+\textbf{1}_{t^{*}=h}\right]$$
$$+\sum_{h=0}^{H}\left(\bel{\expec}{M\bowtie\pi_{\imdp}}\left[\textbf{1}_{t^{*}\geq h+1}\cdot\bonval^{\gen(\poli)}_{h+1,s_{h+1}}\right]-\bel{\expec}{M\bowtie\pi_{\imdp}}\left[\textbf{1}_{t^{*}\geq h}\cdot\bonval^{\gen(\poli)}_{h,s_{h}}\right]\right)$$

Combining the sums and expectations, this can be rewritten as
$$=\sum_{h=0}^{H}\bel{\expec}{M\bowtie\pi_{\imdp}}\left[\textbf{1}_{t^{*}\geq h+1}\left(r_{h}+\bonval^{\gen(\poli)}_{h+1,s_{h+1}}-\bonval^{\gen(\poli)}_{h,s_{h}}\right)+\textbf{1}_{t^{*}=h}\left(1-\bonval^{\gen(\poli)}_{h,s_{h}}\right)\right]$$

By the definition of $t^{*}$ and $h\leq H$, if $t^{*}=h$, $\bonval^{\gen(\poli)}_{h,s_{h}}=1$, so we can rewrite as
$$=\sum_{h=0}^{H}\bel{\expec}{M\bowtie\pi_{\imdp}}\left[\textbf{1}_{t^{*}\geq h+1}\left(r_{h}+\bonval^{\gen(\poli)}_{h+1,s_{h+1}}-\bonval^{\gen(\poli)}_{h,s_{h}}\right)\right]$$

The expectation can be viewed as an expectation over the history up to $s_{h}$, and then an expectation over what $r_{h}$ and $s_{h+1}$ are. Once $s_{h}$ has happened, that determines what $\bonval^{\gen(\poli)}_{h,s_{h}}$ is, and whether $t^{*}\geq h+1$, so those terms can be treated as constants, and the second expectation can be moved inside.
$$=\sum_{h=0}^{H}\bel{\expec}{M\bowtie\pi_{\imdp}}\left[\textbf{1}_{t^{*}\geq h+1}\left(\bel{\expec}{r,s\sim M(h,s_{h},\pi_{\imdp}(h,s_{h}))}\left[r+\bonval^{\gen(\poli)}_{h+1,s}\right]-\bonval^{\gen(\poli)}_{h,s_{h}}\right)\right]$$

By $t^{*}\geq h+1$ and the definition of $t^{*}$, we know that $\bonval^{\gen(\poli)}_{h,s_{h}}<1$. We can now unpack the definition of $\bonval^{\gen(\poli)}_{h,s_{h}}$ to yield
$$=\sum_{h=0}^{H}\bel{\expec}{M\bowtie\pi_{\imdp}}\left[\textbf{1}_{t^{*}\geq h+1}\left(\bel{\expec}{r,s\sim M(h,s_{h},\pi_{\imdp}(h,s_{h}))}\left[r+\bonval^{\gen(\poli)}_{h+1,s}\right]-b^{\gen(\poli)}_{h,s_{h}}-\bel{\expec}{a\sim\poli(h,s_{h})}\left[\overline{Q}^{\gen(\poli)}_{h,s_{h},a}\right]\right)\right]$$
\begin{equation} \label{eq:5}
\leq\sum_{h=0}^{H}\bel{\expec}{M\bowtie\pi_{\imdp}}\left[\textbf{1}_{t^{*}\geq h+1}\left(-b^{\gen(\poli)}_{h,s_{h}}+\bel{\max}{a'}\ \bel{\expec}{r,s\sim M(h,s_{h},a')}\left[r+\bonval^{\gen(\poli)}_{h+1,s}\right]-\bel{\expec}{a\sim\poli(h,s_{h})}\left[\overline{Q}^{\gen(\poli)}_{h,s_{h},a}\right]\right)\right]
\end{equation}

Now, because $t^{*}\geq h+1$, we have that $\bonval^{\gen(\poli)}_{h,s_{h}}<1$, and this was defined to be one if $\mathbb{P}_{\gen(\poli)}(h,s_{h})=0$, so this state has nonzero probability. By our choice of $\poli$, we may now invoke Lemma \ref{lem14}, and choose $N$ to be $\lambda a.\expec_{r,s\sim M(h,s_{h},a)}\left[r+\bonval^{\gen(\poli)}_{h+1,s}\right]$, to derive the following inequality for all $h,s_{h}$. Note that the $\eps$ term is disregarded, because $\eps$ can be set as close to zero as we desire.
$$b^{\gen(\poli)}_{h,s_{h}}\geq\bel{\max}{a'}\ \bel{\expec}{r,s\sim M(h,s_{h},a')}\left[r+\bonval^{\gen(\poli)}_{h+1,s}\right]-\bel{\expec}{a\sim\poli(h,s_{h})}\left[\overline{Q}^{\gen(\poli)}_{h,s_{h},a}\right]$$
$$-\frac{\gamma}{8}\mathbb{P}_{\gen(\poli)}(h,s_{h})\bel{\expec}{a\sim\poli(h,s_{h})}\left[\left(\bel{\expec}{r,s\sim M(h,s_{h},a)}\left[r+\bonval^{\gen(\poli)}_{h+1,s}\right]-\overline{Q}^{\gen(\poli)}_{h,s_{h},a}\right)^{2}\right]$$

This can be reshuffled into
$$b^{\gen(\poli)}_{h,s_{h}}+\frac{\gamma}{8}\mathbb{P}_{\gen(\poli)}(h,s_{h})\bel{\expec}{a\sim\poli(h,s_{h})}\left[\left(\bel{\expec}{r,s\sim M(h,s_{h},a)}\left[r+\bonval^{\gen(\poli)}_{h+1,s}\right]-\overline{Q}^{\gen(\poli)}_{h,s_{h},a}\right)^{2}\right]$$
$$\geq\bel{\max}{a'}\ \bel{\expec}{r,s\sim M(h,s_{h},a')}\left[r+\bonval^{\gen(\poli)}_{h+1,s}\right]-\bel{\expec}{a\sim\poli(h,s_{h})}\left[\overline{Q}^{\gen(\poli)}_{h,s_{h},a}\right]$$

Applying this upper bound to \ref{eq:5}, the bonus terms cancel, and we have 
$$\leq\sum_{h=0}^{H}\bel{\expec}{M\bowtie\pi_{\imdp}}\left[\textbf{1}_{t^{*}\geq h+1}\cdot\frac{\gamma}{8}\mathbb{P}_{\gen(\poli)}(h,s_{h})\cdot\bel{\expec}{a\sim\poli(h,s_{h})}\left[\left(\bel{\expec}{r,s\sim M(h,s_{h},a)}\left[r+\bonval^{\gen(\poli)}_{h+1,s}\right]-\overline{Q}^{\gen(\poli)}_{h,s_{h},a}\right)^{2}\right]\right]$$
$$\leq\sum_{h=0}^{H}\bel{\expec}{M\bowtie\pi_{\imdp}}\left[\frac{\gamma}{8}\mathbb{P}_{\gen(\poli)}(h,s_{h})\cdot\bel{\expec}{a\sim\poli(h,s_{h})}\left[\left(\bel{\expec}{r,s\sim M(h,s_{h},a)}\left[r+\bonval^{\gen(\poli)}_{h+1,s}\right]-\overline{Q}^{\gen(\poli)}_{h,s_{h},a}\right)^{2}\right]\right]$$
$$\leq\frac{\gamma}{8}\sum_{h=0}^{H}\sum_{s\in\states}\mathbb{P}_{\gen(\poli)}(h,s_{h})\cdot\bel{\expec}{a\sim\poli(h,s_{h})}\left[\left(\bel{\expec}{r,s\sim M(h,s_{h},a)}\left[r+\bonval^{\gen(\poli)}_{h+1,s}\right]-\overline{Q}^{\gen(\poli)}_{h,s_{h},a}\right)^{2}\right]$$
$$=\frac{\gamma}{8}\sum_{h=0}^{H}\expec_{\gen(\poli)}\left[\bel{\expec}{a\sim\poli(h,s_{h})}\left[\left(\bel{\expec}{r,s\sim M(h,s_{h},a)}\left[r+\bonval^{\gen(\poli)}_{h+1,s}\right]-\overline{Q}^{\gen(\poli)}_{h,s_{h},a}\right)^{2}\right]\right]$$
$$=\frac{\gamma}{8}\expec_{\gen(\poli)}\left[\sum_{h=0}^{H}\bel{\expec}{a\sim\poli(h,s_{h})}\left[\left(\bel{\expec}{r,s\sim M(h,s_{h},a)}\left[r+\bonval^{\gen(\poli)}_{h+1,s}\right]-\overline{Q}^{\gen(\poli)}_{h,s_{h},a}\right)^{2}\right]\right]$$
$$=\frac{\gamma}{8}\expec_{\gen(\poli)}\left[\sum_{h=0}^{H}\bel{\expec}{a\sim\poli(h,s_{h})}\left[\left(\bel{\expec}{r,s\sim M(h,s_{h},a)}\left[r+\bonval^{\gen(\poli)}_{h+1,s}\right]-\bel{\expec}{(\gen(\poli)|h,s_{h},a)_{\downarrow h}}\left[r+\bonval^{\gen(\poli)}_{h+1,s}\right]\right)^{2}\right]\right]$$

Passing $M(h,s_{h},a)$ and its counterpart through the convert function (to convert $\Delta([0,1]\times\states)$ into $\Delta(\{0,1\}\times\states)$), does not affect the expected rewards. Because $\bonval^{\gen(\poli)}_{h+1,s}\leq 1$, and the rewards are in $[0,1]$, these functions are in $[0,2]$, so the difference in expectations can be upper bounded by 2 times the total variation distance between the converted distributions.
$$\leq\frac{\gamma}{8}\expec_{\gen(\poli)}\left[\sum_{h=0}^{H}\bel{\expec}{a\sim\poli(h,s_{h})}\left[4D^{2}_{TV_{\{0,1\}}}\left((\gen(\poli)|h,s_{h},a)_{\downarrow h},M(h,s_{h},a)\right)\right]\right]$$

The total variation distance between the converted distributions is less than $\sqrt{2}$ times the Hellinger distance between them. Then pull the 8 out and cancel.
$$\leq\gamma\expec_{\gen(\poli)}\left[\sum_{h=0}^{H}\bel{\expec}{a\sim\poli(h,s_{h})}\left[\hellsc\left((\gen(\poli)|h,s_{h},a)_{\downarrow h},M(h,s_{h},a)\right)\right]\right]$$

Chaining these inequalities together, we have derived
$$\expec_{M\bowtie\pi_{\imdp}}\left[\sum_{h=0}^{t^{*}-1}r_{h}+\textbf{1}_{t^{*}=h}\right]-\bonval^{\gen(\poli)}_{0,s_{0}}$$
$$\leq\gamma\cdot\expec_{\gen(\poli)}\left[\sum_{h=0}^{H}\bel{\expec}{a\sim\poli(h,s_{h})}\left[\hellsc\left((\gen(\poli)|h,s_{h},a)_{\downarrow h},M(h,s_{h},a)\right)\right]\right]$$

Applying this to \ref{eq:4}, the Hellinger error terms cancel, and our overall upper bound on the offset DEC is now
\begin{equation} \label{eq:6}
\leq\bonval^{\gen(\poli)}_{0,s_{0}}-\expec_{\gen(\poli)}\left[\sum_{h=0}^{H}r_{h}+b^{\gen(\poli)}_{h,s_{h}}\right]+\expec_{\gen(\poli)}\left[\sum_{h=0}^{H}b^{\gen(\poli)}_{h,s_{h}}\right]
\end{equation}

Now, we'll prove that for all $h\in[H]_{0}$, we have
\begin{equation} \label{eq:7}
\expec_{\gen(\poli)}\left[\sum_{k=0}^{h-1}\left(r_{k}+b^{\gen(\poli)}_{k,s_{k}}\right)+\bonval^{\gen(\poli)}_{h,s_{h}}\right]\leq\expec_{\gen(\poli)}\left[\sum_{k=0}^{h}\left(r_{k}+b^{\gen(\poli)}_{k,s_{k}}\right)+\bonval^{\gen(\poli)}_{h+1,s_{h+1}}\right]
\end{equation}

Subtracting from both sides, it suffices to prove
$$\expec_{\gen(\poli)}\left[\bonval^{\gen(\poli)}_{h,s_{h}}\right]\leq\expec_{\gen(\poli)}\left[r_{h}+b^{\gen(\poli)}_{h,s_{h}}+\bonval^{\gen(\poli)}_{h+1,s_{h+1}}\right]$$

This is proven as follows. By the definition of $\bonval^{\gen(\poli)}_{h,s_{h}}$ and $\overline{Q}^{\gen(\poli)}_{h,s_{h},a}$, we have
$$\expec_{\gen(\poli)}\left[\bonval^{\gen(\poli)}_{h,s_{h}}\right]\leq\expec_{\gen(\poli)}\left[b^{\gen(\poli)}_{h,s_{h}}+\bel{\expec}{a\sim\poli(h,s_{h})}\left[\bel{\expec}{r'_{h},s'_{h+1}\sim\gen(\poli)|h,s_{h},a}\left[r'_{h}+\bonval^{\gen(\poli)}_{h+1,s'_{h+1}}\right]\right]\right]$$
$$=\expec_{\gen(\poli)}\left[b^{\gen(\poli)}_{h,s_{h}}+\bel{\expec}{r'_{h},s'_{h+1}\sim\gen(\poli)|h,s_{h}}\left[r'_{h}+\bonval^{\gen(\poli)}_{h+1,s'_{h+1}}\right]\right]=\expec_{\gen(\poli)}\left[b^{\gen(\poli)}_{h,s_{h}}+r_{h}+\bonval^{\gen(\poli)}_{h+1,s_{h+1}}\right]$$

The result in \ref{eq:7} holding for all $h$. We chain these inequalities together, and use $\bonval^{\gen(\poli)}_{H+1,s_{H+1}}=0$, to yield $\bonval^{\gen(\poli)}_{0,s_{0}}\leq \expec_{\gen(\poli)}\left[\sum_{h=0}^{H}r_{h}+b^{\gen(\poli)}_{h,s_{h}}\right]$. Plugging this into \ref{eq:6}, canceling, and pulling out the sum, we get
$$\leq\sum_{h=0}^{H}\expec_{\gen(\poli)}\left[b^{\gen(\poli)}_{h,s_{h}}\right]=\sum_{h=0}^{H}\sum_{s\in\states}\mathbb{P}_{\gen(\poli)}(h,s)\cdot b^{\gen(\poli)}_{h,s}$$

The bonus value was defined as the value of the offset DEC, with square loss, for a specific instance of the multi-armed bandit problem. The proof of Proposition 5.5 in \cite{Foster21} may be used to upper-bound this quantity, where the $\gamma$ in Foster's proof of Proposition 5.5 is chosen to be twice our value of $\gamma$. The net result is an upper bound on the offset DEC of $\frac{A}{4\gamma}$. However, the chosen value of $\gamma$ fed into the odec function varies depending on the choice of $h,s$, and there is also one instance of the one-armed bandit problem at the start. Plugging in our values, we get
$$\leq\mathbb{P}_{\gen(\poli)}(0,s_{0})\frac{1}{4\frac{\gamma}{8}\mathbb{P}_{\gen(\poli)}(0,s_{0})}+\sum_{h=1}^{H}\sum_{s\in\states}\mathbb{P}_{\gen(\poli)}(h,s)\cdot\frac{A}{4\frac{\gamma}{8}\mathbb{P}_{\gen(\poli)}(h,s)}=\frac{2(HSA+1)}{\gamma}$$

And so, we have upper-bounded the modified offset DEC by $\frac{2(HSA+1)}{\gamma}$ and the lemma follows. $\blacksquare$

\begin{theo} \label{the4}
In the episodic RMDP setting, if $\gen:\Pi_{RNS}\to\Delta([0,1]\times(\states\times\acts\times[0,1])^{[H]})$ is continuous and policy-coherent, the modified fuzzy DEC fulfills $\text{dec}^{f'}_{\eps}(\hypo,\gen)\leq 2\sqrt{2(HSA+1)}\eps$
\end{theo}

The proof of Proposition \ref{pro1} does not specifically rely on Hellinger-squared loss, and we may avoid the use of Sion's Minimax Theorem in the proof at the cost of introducing a $\leq$ instead of a strict equality, so we have
$$\text{dec}^{f'}_{\eps}(\hypo,\gen)\leq\min_{\gamma\geq 0}\left(\max\left(0,\text{dec}^{o'}_{\gamma}(\hypo,\gen)\right)+\gamma\eps^{2}\right)$$

From Lemma \ref{lem15} we have $\text{dec}^{o'}_{\gamma}(\hypo,\gen)\leq\frac{2(HSA+1)}{\gamma}$. Plugging this in and computing the minimizing value of $\gamma$ to be $\frac{\sqrt{2(HSA+1)}}{\eps}$, and plugging this in, we get
$$\text{dec}^{f'}_{\eps}(\hypo,\gen)\leq2\sqrt{2(HSA+1)}\eps$$

And the result follows. $\blacksquare$
\section{Auxiliary Definitions for RMDP Estimator:}

The estimator which achieves low estimation complexity for RMDP's is not the RUE algorithm, although it is conceptually similar. We define many functions and sets for the following lemmas, and the definition of our estimator of choice.
\\

With the state space, action space, and time horizon $H$ fixed, the space $\mathcal{RMDP}$ of compatible RMDP's (not necessarily 1-bounded), with our modifications (that the starting state and action are unique, and the ending state is unique), may be described by the transition kernel alone, so we define the type of RMDP's via
$$\mathcal{RMDP}:=\Box([0,1]\times\states)\times\left([H-1]\times\states\times\acts\to\Box([0,1]\times\states)\right)\times(\states\times\acts\to\Box[0,1])$$

The notation $[H-1]$ denotes the set of integers from 1 to $H-1$. The type $\mathcal{PRMDPR}$ of Partial RMDP's with Recommendations, is then defined as
$$\mathcal{PRMDPR}:=\Box([0,1]\times\states)\times\left([H-1]\times\states\to\acts\times\Box([0,1]\times\states)\right)\times(\states\to\acts\times\Box[0,1])$$

Intuitively, for each $h,s$, these recommend an action and predict the consequences of that action. 

Given spaces $X,Y,Z$, there is a function $\text{shift}$ of type $(X\to Y\times\Box Z)\to(X\times Y\to\Box Z)$, defined as
$$\text{shift}(f)(x,y):=\text{ if }f(x)_{Y}=y,f(x)_{\Box Z}\text{ else }\Delta Z$$

This corresponds to total uncertainty about the consequences of non-recommended $y$, while making a prediction about the consequences of the recommended $y$.

PRMDPR's can be converted to RMDP's via $(\text{id},\text{shift},\text{shift})$, which adds complete uncertainty about non-recommended actions, and RMDP's can be converted to models of type
\\
$\Pi_{RNS}\to\Box\left([0,1]\times(\states\times\acts\times[0,1])^{[H]}\right)$ via $\imdp\mapsto\Set{\sigma\bowtie\pi}{\sigma\models\imdp}$. This produces an imprecise belief about trajectories by letting an RMDP interact with a policy.
\\

Given an $f:\states\to[0,1]$ and a $c:[0,1]$, the imprecise belief $\Psi^{\{0,1\}}_{f,c}:\Box(\{0,1\}\times\states)$ is defined as
$$\Psi^{\{0,1\}}_{f,c}:=\Set{\mu\in\Delta(\{0,1\}\times\states)}{\bel{\expec}{r,s\sim\mu}[f(s)+r]\geq c}$$
The imprecise beliefs $\Psi^{\{0,1\}}_{c}:\Box\{0,1\}$, $\Psi^{[0,1]}_{f,c}:\Box([0,1]\times\states)$, and $\Psi^{[0,1]}_{c}:\Box[0,1]$ are defined similarly by adjusting the space of outcomes or probability distributions accordingly.

$\hypo^{\{0,1\}}_{\text{mid}}\subseteq\Box(\{0,1\}\times\states)$, the set of "halfspace hypotheses" for the middle of an RMDP, is
$$\hypo^{\{0,1\}}_{\text{mid}}:=\Set{\Psi^{\{0,1\}}_{f,c}}{f:\states\to[0,1],c:[0,1]}$$
The spaces $\hypo^{\{0,1\}}_{\text{end}}\subseteq\Box\{0,1\}$, $\hypo^{[0,1]}_{\text{mid}}$, and $\hypo^{[0,1]}_{\text{end}}$, are defined similarly, using their corresponding $\Psi$ sets.

Finally, the subset $\hypo_{\text{parhalf}}$ of PRMDPR's, is defined as
$$\hypo_{\text{parhalf}}:=\hypo^{[0,1]}_{\text{mid}}\times\left([H-1]\times\states\to\acts\times\hypo^{[0,1]}_{\text{mid}}\right)\times\left(\states\to\acts\times\hypo^{[0,1]}_{\text{end}}\right)$$

Because $\hypo_{\text{parhalf}}$ is a set of PRMDPR's, it can be converted to a set of RMDP's, and a set of models. By abuse of notation, these induced sets are also be denoted $\hypo_{\text{parhalf}}$. We now present some lemmas about these definitions.

\begin{lemm} \label{lem16}
Given a distribution $\nu$ supported on $\{0,1\}\times\states$, for any $f:\states\to[0,1]$, and $c:[0,1]$, $\hellsc\left(\nu\to\Psi^{[0,1]}_{f,c}\right)=\hells\left(\nu\to\Psi^{\{0,1\}}_{f,c}\right)$
\end{lemm}

We use $r'$ for rewards that are 0 or 1, and $r$ for rewards in $[0,1]$. $\bel{\expec}{r'\sim r}$ means expected value, when $r$ is rounded up to 1 with $r$ probability, and down to 0 with $1-r$ probability.
\\

To begin we have $\text{convert}\left(\Psi^{[0,1]}_{f,c}\right)=\Psi^{\{0,1\}}_{f,c}$, by the following argument. If $\mu\in\text{convert}\left(\Psi^{[0,1]}_{f,c}\right)$, there exists a $\nu\in\Psi^{[0,1]}_{f,c}$ s.t. $\mu=\text{convert}(\nu)$.
We then have
$$\bel{\expec}{r',s\sim\mu}[f(s)+r]=\bel{\expec}{r',s\sim\text{convert}(\nu)}[f(s)+r]=\bel{\expec}{r,s\sim\nu}\left[\bel{\expec}{r'\sim r}\left[f(s)+r'\right]\right]=\bel{\expec}{r,s\sim\nu}\left[f(s)+r\right]\geq c$$

This was by the definition of conversion (a distribution $\nu$ can be converted by rounding the rewards up or down appropriately), pulling the constant function out of the inner expectation, using that $\bel{\expec}{r'\sim r}[r']=r$, and using that $\nu\in\Psi^{[0,1]}_{f,c}$. The fact that $\bel{\expec}{r',s\sim\mu}[f(s)+r]\geq c$ then certifies that $\mu\in\Psi^{\{0,1\}}_{f,c}$. This establishes $\text{convert}\left(\Psi^{[0,1]}_{f,c}\right)\subseteq\Psi^{\{0,1\}}_{f,c}$. 

In the other direction, for any $\mu\in\Psi^{\{0,1\}}_{f,c}$, the type signature of $\mu$ can be straightforwardly changed to be $\Delta([0,1]\times\states)$. The expectation of $f$ plus reward exceeded $c$ because $\mu\in\Psi^{\{0,1\}}_{f,c}$, and altering its type signature doesn't change its expectation, so $\mu\in\Psi^{[0,1]}_{f,c}$. Converting $\mu$ back then involves rounding 1's up to 1 with 1 probability, and rounding 0's down to 0 with 1 probability, which has no effect, so $\mu\in\text{convert}(\Psi^{[0,1]}_{f,c})$ as well, establishing $\text{convert}(\Psi^{[0,1]}_{f,c})\supseteq\Psi^{\{0,1\}}_{f,c}$. Both subset inclusion directions have been shown, so the sets are equal.

Now, by applying the definition of the converted Hellinger distance, and our result proved above, we compute
$$\hellsc\left(\nu\to\Psi^{[0,1]}_{f,c}\right)=\hells\left(\text{convert}(\nu)\to\text{convert}(\Psi^{[0,1]}_{f,c})\right)=\hells\left(\nu\to\Psi^{\{0,1\}}_{f,c}\right)$$

We could remove the conversion from $\nu$ because it was already supported on $\{0,1\}$ by assumption. And the result follows. $\blacksquare$

\begin{lemm} \label{lem17}
Given some $\eps_{1},\eps_{2}$, $f:\states\to[0,1]$, and $c:[0,1]$, use $\lceil f\rceil$ to denote the function $\lambda s.1-\left\lfloor\frac{1-f(s)}{\eps_{1}}\right\rfloor\cdot\eps_{1}$, and $\lfloor c\rfloor$ to denote $\left\lfloor\frac{c}{\eps_{2}}\right\rfloor\cdot\eps_{2}$. We have that $\hells\left(\Psi^{\{0,1\}}_{\lceil f\rceil,\lfloor c\rfloor}\to\Psi^{\{0,1\}}_{f,c}\right)\leq\eps_{1}+\eps_{2}$.
\end{lemm}

By definition,
$$\hells\left(\Psi^{\{0,1\}}_{\lceil f\rceil,\lfloor c\rfloor}\to\Psi^{\{0,1\}}_{f,c}\right)=\bel{\max}{\mu\in\Psi^{\{0,1\}}_{\lceil f\rceil,\lfloor c\rfloor}}\bel{\min}{\mu'\in\Psi^{\{0,1\}}_{f,c}}\hells(\mu,\mu')$$

Fixing the maximizing $\mu$, we will construct a $\mu'$ which witnesses that the Hellinger distance is low. $\mu'$ is made by taking $\mu$, and moving $\eps_{1}+\eps_{2}$ probability mass from points $s,0$ to their corresponding point $s,1$. If there is less than $\eps_{1}+\eps_{2}$ probability mass on points $s,0$, move all of it up to their corresponding points $s,1$.

By Hellinger distance squared being less than the total variation distance, and us moving $\eps_{1}+\eps_{2}$ measure to construct $\mu'$ from $\mu$, we may compute 
$$\hells(\mu,\mu')\leq D_{TV}(\mu,\mu')\leq\eps_{1}+\eps_{2}$$
yielding our result. However, we still have to show that $\mu'\in\Psi^{\{0,1\}}_{f,c}$. To show this, we must show that $\bel{\expec}{r,s\sim\mu'}[f(s)+r]\geq c$. If less than $\eps_{1}+\eps_{2}$ measure was moved, this implies that $\mu'$ has all of its measure on 1 reward, which trivially implies our desired inequality. Therefore, assume that $\eps_{1}+\eps_{2}$ measure was moved to construct $\mu'$. We may then compute
$$\expec_{\mu'}[f+r]=\expec_{\mu'}[f]+\expec_{\mu'}[r]=\expec_{\mu}[f]+(\expec_{\mu}[r]+\eps_{1}+\eps_{2})=\expec_{\mu}[\lceil f\rceil+r]+\expec_{\mu}[f-\lceil f\rceil]+\eps_{1}+\eps_{2}$$
$$\geq\lfloor c\rfloor+\expec_{\mu}[f-\lceil f\rceil]+\eps_{1}+\eps_{2}\geq (c-\eps_{2})-\eps_{1}+\eps_{1}+\eps_{2}=c$$

In order, this is because $\mu'$ was made from $\mu$ by moving $\eps_{1}+\eps_{2}$ measure from 0 to 1 reward while not affecting the distribution over states. Then, we used that $\mu\in\Psi_{\lceil f\rceil,\lfloor c\rfloor}$. Finally, we used that the rounding procedure only increases the value of $f$ by $\eps_{1}$ at most, and decreases $c$ by $\eps_{2}$ at most. Because $\mu'\in\Psi^{\{0,1\}}_{f,c}$, and $\mu'$ was constructed from the distance-maximizing $\mu$, our result follows. $\blacksquare$
\section{Definition of the RMDP Estimator}

$T$ (the number of episodes), and three nonzero parameters $\eps_{[0,1]},\eps_{\states},\eps'$,are taken as input. $\eps_{\text{pess}}$ is taken to be $\sqrt{\frac{1}{T(H+1)^{2}}}$. We now define our estimator, and the associated functions, as follows.
\\

Given an MDP $M$ and an $h\geq 0$, $M^{>h}$ denotes the partial transition kernel consisting of $M(h',s,a)$ for all $s,a$ and $h'>h$. Given a $\mu:\Delta(\{0,1\}\times\states)$, $(\mu,M^{>h})$ may be considered as an MDP which starts with the reward on timestep $h$, and policies can interact with this MDP. Accordingly, given an MDP $M$ and timestep $h$, we define 
\\
$\widetilde{M}^{>h}:\Delta(\{0,1\}\times\states)\times\Pi_{RNS}\to\Delta\left(\{0,1\}\times(\states\times\acts\times\{0,1\})^{\{h+1,...,H\}}\right)$ as
$$\widetilde{M}^{>h}(\mu,\pi):=(\mu,M^{>h})\bowtie\pi$$

And we define $\widetilde{M}^{\geq h}:\states\times\acts\times\Pi_{RNS}\to\Delta\left((\states\times\acts\times\{0,1\})^\{h,...,H\}\right)$ as
$$\widetilde{M}^{\geq h}(s,a,\pi):=\delta_{s,a}\times((M(h,s,a),M^{>h})\bowtie\pi)$$

Where $\delta_{s,a}$ is the distribution which puts all measure on $s,a$. $\widetilde{M}^{>h}(\mu,\pi)$ takes the policy $\pi$ and MDP $M$ and counterfacts on $r_{h},s_{h+1}$ being distributed according to $\mu$ to derive a distribution over trajectories. $\widetilde{M}^{\geq h}(s,a,\pi)$ does the same, but counterfacts on a specific $s,a$ appearing on timestep $h$ instead. 

For the case where $h=H$, $\widetilde{M}^{>H}:\Delta\{0,1\}\times\Pi_{RNS}\to\Delta\{0,1\}$, and $\widetilde{M}^{>H}(\mu,\pi)=\mu$. For the case where $h=0$, $\widetilde{M}^{\geq 0}$ is only ever invoked on the unique initial state and action at timestep 0, and $M(0,s_{0},a_{0})$ is the initial distribution of the MDP $M$.
\\

The finite set $\hypo'_{\text{mid}}\subseteq\Box(\{0,1\}\times\states)$, is defined by
$$\Set{\Psi^{\{0,1\}}_{f,c}}{\forall s\left(\exists i\in\left\{0,1...\left\lfloor\frac{1}{\eps_{\states}}\right\rfloor\right\}:f(s)=1-i\cdot\eps_{\states}\wedge\exists j\in\left\{0,1...\left\lfloor\frac{1}{\eps_{[0,1]}}\right\rfloor\right\}:c=j\cdot\eps_{[0,1]}\right)}$$

The finite set $\hypo'_{\text{end}}\subseteq\Box\{0,1\}$, is defined by
$$\hypo'_{\text{end}}:=\Set{\Psi^{\{0,1\}}_{c}}{\exists j\in\left\{0,1...\left\lfloor\frac{1}{\eps_{[0,1]}}\right\rfloor\right\}:c=j\cdot\eps_{[0,1]}}$$

The finite sets $\mathcal{B}_{\text{frag}},\mathcal{B}_{\text{cal}},\mathcal{B}_{\text{unif}}$, of fragment, calibration, and uniform bettors, are indexed as follows. $+$ is taken to be disjoint union, in the type theory sense. $[H-1]$ denotes the set of integers from 1 to $H-1$. $\textbf{1}$ denotes the unit type.
$$\mathcal{B}_{\text{frag}}:=\hypo'_{\text{mid}}+([H-1]\times\states\times\acts\times\hypo'_{\text{mid}})+(\states\times\acts\times\hypo'_{\text{end}})$$
$$\mathcal{B}_{\text{cal}}:=([H-1]\times\states\times\acts\times\hypo'_{\text{mid}})+(\states\times\acts\times\hypo'_{\text{end}})$$
$$\mathcal{B}_{\text{unif}}:=\textbf{1}+([H-1]\times\states\times\acts)+(\states\times\acts)$$

The set $\mathcal{B}$ of all bettors consists of the fragment and calibration and uniform bettors, and one pessimism bettor.
$$\mathcal{B}:=\mathcal{B}_{\text{frag}}+\mathcal{B}_{\text{cal}}+\mathcal{B}_{\text{unif}}+\textbf{1}$$

Given a $B$ which is a fragment, calibration, or uniform bettor, $h_{B},s_{B},a_{B},\Psi_{B}$ are their corresponding timestep, state, action, and hypothesis. For fragment bettors and the uniform bettor in the first component of their disjoint union, $h_{B}=0$, and $s_{B},a_{B}$ are the unique initial state and initial action at time $0$. For fragment, calibration, and uniform bettors in the last component of their disjoint union, $h_{B}=H$. For uniform bettors, their associated hypothesis $\Psi_{B}$ is the set containing only the uniform distribution on $\states\times\{0,1\}$, or on $\{0,1\}$, respectively.

The notation $(h,s,a)_{B}$ is shorthand for the tuple $h_{B},s_{B},a_{B}$, and similar for $(h,s)_{B}$. The probability of the event $(h,s)_{B}$ is shorthand for the probability of $s_{h_{B}}=s_{B}$, that the state on the $h_{B}$'th timestep matches $s_{B}$.

Given a calibration bettor $B$, policy $\pi:\Pi_{RNS}$, and an MDP $M$, define
$$X^{B}_{M,\pi}:=\pi((h,s)_{B})(a_{B})\cdot\hells(M((h,s,a)_{B})\to\Psi_{B})$$

Given either a fragment or uniform bettor $B$, and an MDP $M$, define
$$\mu^{B}_{M}:=\bel{\text{argmin}}{\mu\in\Psi_{B}}\ \hells(M((h,s,a)_{B}),\mu)$$

If $h_{B}=0$, then $s_{B},a_{B}$ will be the unique initial state and action, and $M((h,s,a)_{B})$ is the initial distribution on rewards and states of the MDP.
\\

Now, given a bettor $B$, an MDP $M$, a policy $\pi$, and a situation $h,s,a$ with $h\geq 0$, define the $g$ functions $g^{B,h,s,a}_{M,\pi}:\Delta(\{0,1\}\times\states)\to\mathbb{R}$ and local betting functions $\text{lbet}^{B,h,s,a}_{M,\pi}:[0,1]\times\states\times\acts\to\mathbb{R}$ as follows, by case analysis on bettors.

If $B$ is a fragment or uniform bettor and $h,s,a\neq(h,s,a)_{B}$, $g^{B,h,s,a}_{M,\pi}$ is the constant-zero function and $\text{lbet}^{B,h,s,a}_{M,\pi}$ is the constant-one function. If $h,s,a=(h,s,a)_{B}$, we then have
$$g^{B,h,s,a}_{M,\pi}(\mu):=2\hells(\mu\to\Psi_{B})$$
$$\text{lbet}^{B,h,s,a}_{M,\pi}(r,s',a'):=\bel{\expec}{r'\sim r}\left[\sqrt{\frac{\mu^{B}_{M}(r',s')}{M(h,s,a)(r',s')}}\right]+\hells(M(h,s,a)\to\Psi_{B})$$

The notation $\bel{\expec}{r'\sim r}$ denotes sampling a 1 with $r$ probability, and a 0 otherwise, which is needed to convert between the continuous reward $r:[0,1]$, and the distributions being over the reward space $\{0,1\}$. As usual, if $h=0$, $s_{B},a_{B}$ will be the unique initial state and action, and $M(h,s,a)$ is the initial distribution of the MDP $M$. If $h=H$, only the probability of reward is assessed and the equation doesn't depend on $s'$.

If $B$ is a calibration bettor, and $h\geq h_{B}$, $g^{B,h,s,a}_{M,\pi}$ is the constant-zero function and $\text{lbet}^{B,h,s,a}_{B,\pi}$ is the constant-one function. If $h<h_{B}$, we then have
$$g^{B,h,s,a}_{M,\pi}(\mu):=\frac{X^{B}_{M,\pi}}{4}\mathbb{P}_{\widetilde{M}^{>h}(\mu,\pi)}((h,s)_{B})$$
$$\text{lbet}^{B,h,s,a}_{M,\pi}(r,s',a'):=1+\frac{X^{B}_{M,\pi}}{4}\left(\mathbb{P}_{\widetilde{M}^{\geq h}(s,a,\pi)}((h,s)_{B})-\mathbb{P}_{\widetilde{M}^{\geq h+1}(s',a',\pi)}((h,s)_{B})\right)$$

Rounding out our case analysis, if $B$ is the unique pessimism bettor, we have
$$g^{B,h,s,a}_{M,\pi}(\mu):=\eps_{\text{pess}}\cdot\bel{\expec}{\widetilde{M}^{>h}(\mu,\pi)}\left[\sum_{k=h}^{H}r_{k}\right]$$
$$\text{lbet}^{B,h,s,a}_{M,\pi}(r,s',a'):=1+\eps_{\text{pess}}\cdot\left(\expec_{\widetilde{M}^{\geq h}(s,a,\pi)}\left[\sum_{k=h}^{H}r_{k}\right]-\left(r-\expec_{\widetilde{M}^{\geq h+1}(s',a',\pi)}\left[\sum_{k=h+1}^{H}r_{k}\right]\right)\right)$$

This defines the $g$ functions and local betting functions $\text{lbet}$ for all bettors $B$ and situations $h,s,a$, relative to an MDP $M$ and policy $\pi$. 
\\

Now we can finally present the core of the RMDP estimator. Given a $\zeta_{t}:\Delta\mathcal{B}$, and policy $\pi$, we define the MDP $\guesspi$ by downwards induction as
$$\guesspi(h,s,a):=\bel{\text{argmin}}{\mu\in\Delta(\{0,1\}\times\states)}\ \bel{\expec}{B\sim\zeta_{t}}\left[g^{B,h,s,a}_{\guesspi,\pi}(\mu)\right]$$

Given a $\zeta_{t}:\Delta\mathcal{B}$, the final estimated model, $\widehat{M}_{t}:\Pi_{RNS}\to\Delta\left(\{0,1\}\times(\states\times\acts\times\{0,1\})^{[H]}\right)$, is then given by $\lambda\pi.\guesspi\bowtie\pi$.

All entries of the $\guesspi$ transition kernel involve solving a convex optimization problem, and so are feasible to compute. Also, the definition of $\guesspi$ is well-founded. The base case holds because the $g$ functions used at level $H$ reference $\widetilde{M}^{>H}_{t,\pi}(\mu,\pi)$, but this function is just the identity on $\mu$. The downwards induction step holds because all of the $g$ functions used in computing the transition kernel at level $h$ only depend on the transition kernel for $\guesspi$ at levels $h'>h$. 
\\

All that remains to fully define the estimator is to provide a prior distribution $\zeta_{1}$, and show how to update the distribution $\zeta_{t}$ on incoming data. For fragment or calibration bettors, their prior probability is $\frac{1}{2(|\mathcal{B}_{\text{frag}}|+|\mathcal{B}_{\text{cal}}|)}$. For the pessimism bettor, its prior probability is $\frac{1}{2}-\eps'$. For the uniform bettors, their prior probabilities are $\frac{\eps'}{|\mathcal{B}_{\text{unif}}|}$. Updating is given as follows. Given a bettor $B$, MDP $M$ and policy $\pi$, we define $\text{bet}^{B}_{M,\pi}:\left([0,1]\times(\states\times\acts\times[0,1])^{[H]}\right)\to\mathbb{R}$, the aggregate betting function, as
$$\text{bet}^{B}_{M,\pi}(r_{0},s_{1},a_{1},...r_{H}):=1+\sum_{h=0}^{H}\left(\text{lbet}^{B,h,s_{h},a_{h}}_{M,\pi}(r_{h},s_{h+1},a_{h+1})-1\right)$$

For $h=0$, as usual, $s_{0},a_{0}$ are the unique initial state and action. For $h=H$, none of the betting functions depend on the final state and action, so it doesn't matter that $s_{H+1},a_{H+1}$ are ill-defined. We may then succinctly define updating as
$$\bigstar_{t}:=\bel{\expec}{B\sim\zeta_{t}}\left[\text{bet}^{B}_{M_{t,\pi_{t}},\pi_{t}}(tr_{t})\right]$$
$$\zeta_{t+1}(B):=\frac{\zeta_{t}(B)\cdot\text{bet}^{B}_{M_{t,\pi_{t}},\pi_{t}}}{\bigstar_{t}}$$

Where $tr_{t}$ and $\pi_{t}$ are the trajectory in episode $t$ and policy in episode $t$, respectively. This concludes our specification of the estimator for RMDP's. We now turn to analyzing the estimation complexity. When $\guesspi,\pi$ shows up in a subscript, we use a subscript of $t,\pi$ instead, to abbreviate it.
\section{RMDP Estimator Lemmas}

\begin{lemm} \label{lem18}
The estimator $\widehat{M}$ is well-defined, and all $\guesspi(h,s,a)$ have full support.
\end{lemm}

The only potential issues in the definition of the estimator are showing that all $\guesspi(h,s,a)$ have full support so no division-by-zero errors arise when computing the bets, and that all $\zeta_{t}$ are indeed probability distributions. To show that all $\guesspi(h,s,a)$ have full support, we may reuse the proof from Proposition \ref{pro4}, that the uniform bettors ensure that $\guesspi(h,s,a)$ has full support if they have nonzero probability, and these bettors never run out of probability mass completely. Minor adaptations must be made to deal with the presence of the calibration bettors, but their tedium greatly outweighs their impact on the overall proof, and the result holds anyways by the same line of argument. Showing that all $\zeta_{t}$ are probability distributions is slightly more difficult. To show this, we show by induction, that \emph{if} all bets are $\geq 0$ and there is a bettor $B^{*}$ which never has a bet value of zero, all $\zeta_{t}$ are probability distributions. Then we show that these two properties indeed hold.
\\

For the induction base case, $\zeta_{1}$ is a probability distribution, and $\zeta_{1}(B^{*})>0$ holds. For the induction step, we assume that these two properties hold of $\zeta_{t}$ and prove them for $\zeta_{t+1}$ if all bets are $\geq 0$ and the bet of $B^{*}$ is $>0$, as follows.

First, because $\zeta_{t}(B^{*})>0$ by the induction assumption, and $\text{bet}^{B^{*}}_{t,\pi_{t}}(tr_{t})>0$ by assumption, and $\zeta_{t}$ is a probability distribution by the induction assumption, and all bets are $\geq 0$ by assumption, $\bigstar_{t}>0$. This ensures that no division by zero issues occur and $\zeta_{t+1}$ is well-defined. $\zeta_{t}(B^{*})>0$ and $\text{bet}^{B^{*}}_{t,\pi_{t}}(tr_{t})>0$ also implies that $\zeta_{t+1}(B^{*})>0$, proving half of our induction step. To show that $\zeta_{t+1}$ is a probability distribution, we use that $\zeta_{t}$ is a probability distribution by induction assumption and all $B$ fulfill $\text{bet}^{B}_{t,\pi_{t}}(tr_{t})\geq 0$ by assumption, which implies that all $B$ fulfill $\zeta_{t+1}(B)\geq 0$. To show that the sum of measure is 1, we compute
$$\sum_{B\in\mathcal{B}}\frac{\zeta_{t}(B)\cdot\text{bet}^{B}_{t,\pi_{t}}(tr_{t})}{\bigstar_{t}}=\sum_{B\in\mathcal{B}}\frac{\zeta_{t}(B)\cdot\text{bet}^{B}_{t,\pi_{t}}(tr_{t})}{\sum_{B'\in\mathcal{B}}\zeta_{t}(B')\cdot\text{bet}^{B'}_{t,\pi_{t}}(tr_{t})}=\frac{\sum_{B\in\mathcal{B}}\zeta_{t}(B)\cdot\text{bet}^{B}_{t,\pi_{t}}(tr_{t})}{\sum_{B'\in\mathcal{B}}\zeta_{t}(B')\cdot\text{bet}^{B'}_{t,\pi_{t}}(tr_{t})}=1$$

Therefore, the second half of the induction step goes through, proving our result, if all bets are non-negative and there's a bettor which is guaranteed to always make positive bets.
\\

We now shift to proving those two properties. We use the definition of the global betting function $\text{bet}$ and the local betting functions $\text{lbet}$ to compute the overall bets. If $B\in\mathcal{B}_{\text{frag}}\cup\mathcal{B}_{\text{unif}}$, and $s_{h_{B}},a_{h_{B}}\neq s_{B},a_{B}$, all local bets are 1, and cancel out with the $-1$'s in the sum of the global bet, leaving the global bet as 1, which is nonnegative. However, if $s_{h_{B}},a_{h_{B}}=s_{B},a_{B}$, the 1 in the global bet cancels out with the $-1$ at the nontrivial bet, yielding
$$\text{bet}_{t,\pi}^{B}(r_{0},(s,a,r)_{1:H})=\bel{\expec}{r'\sim r_{h_{B}}}\left[\sqrt{\frac{\mu^{B}_{t,\pi}(r',s_{h_{B}+1})}{\guesspi((h,s,a)_{B})(r',s_{h_{B}+1})}}\right]+\hells(\guesspi((h,s,a)_{B})\to\Psi_{B})$$

This is clearly nonnegative and well-defined if $\guesspi((h,s,a)_{B})$ has full support. If $B\in\mathcal{B}_{\text{cal}}$, local bets at and after $h_{B}$ are 1 and cancel out with the $-1$'s in the sum of the global bet. We can then compute 
$$\text{bet}_{t,\pi}^{B}(r_{0},(s,a,r)_{1:H})=1+\sum_{h=0}^{h_{B}-1}\left(\text{lbet}_{t,\pi}^{B,h,s_{h},a_{h}}(r_{h},s_{h+1},a_{h+1})-1\right)$$
$$=1+\sum_{h=0}^{h_{B}-1}\left(\left(1+\frac{X^{B}_{t,\pi}}{4}\left(\mathbb{P}_{\widetilde{M}_{t,\pi}^{\geq h}(s_{h},a_{h},\pi)}((h,s)_{B})-\mathbb{P}_{\widetilde{M}_{t,\pi}^{\geq h+1}(s_{h+1},a_{h+1},\pi)}((h,s)_{B})\right)\right)-1\right)$$

The 1's cancel and the sum is a telescoping sum, yielding
$$=1+\frac{X^{B}_{t,\pi}}{4}\left(\mathbb{P}_{\widetilde{M}_{t,\pi}^{\geq 0}(s_{0},a_{0},\pi)}((h,s)_{B})-\mathbb{P}_{\widetilde{M}_{t,\pi}^{\geq h_{B}}(s_{h_{B}},a_{h_{B}},\pi)}((h,s)_{B})\right)$$

The MDP's are always guaranteed to start with $s_{0}$ and $a_{0}$ at time 0, so $\widetilde{M}_{t,\pi}^{\geq 0}(s_{0},a_{0},\pi)$ is just $\guesspi\bowtie\pi$, ie, $\guess(\pi)$. Further, the event $(h,s)_{B}$ (reaching $s_{B}$ at time $h_{B}$) is an abbreviation for $s_{h_{B}}=s_{B}$, so this probability is always either 1 or 0 and can be replaced with an indicator function, yielding
$$=1+\frac{X^{B}_{t,\pi}}{4}\left(\mathbb{P}_{\guess(\pi)}((h,s)_{B})-\textbf{1}_{s_{h_{B}}=s_{B}}\right)$$

Because $X^{B}_{t,\pi}\in[0,1]$, this quantity is always in $[3/4,5/4]$, and we have found a family of bettors whose bets are not just nonnegative, but also bounded away from zero, providing our desired $B^{*}$.

Finally, for the pessimism bettor, we can compute 
$$\text{bet}_{t,\pi}^{B}(r_{0},(s,a,r)_{1:H})=1+\sum_{h=0}^{H}\left(\text{lbet}_{t,\pi}^{B,h,s_{h},a_{h}}(r_{h},s_{h+1},a_{h+1})-1\right)$$
$$=1+\sum_{h=0}^{H}\left(\left(1+\eps_{\text{pess}}\left(\bel{\expec}{\widetilde{M}_{t,\pi}^{\geq h}(s_{h},a_{h},\pi)}\left[\sum_{k=h}^{H}r_{k}\right]-\left(r_{h}+\bel{\expec}{\widetilde{M}_{t,\pi}^{\geq h+1}(s_{h+1},a_{h+1},\pi)}\left[\sum_{k=h+1}^{H}r_{k}\right]\right)\right)\right)-1\right)$$

The 1's cancel and the sum is a telescoping sum, yielding
$$=1+\eps_{\text{pess}}\left(\bel{\expec}{\widetilde{M}_{t,\pi}^{\geq 0}(s_{0},a_{0},\pi)}\left[\sum_{k=0}^{H}r_{k}\right]-\sum_{h=0}^{H}r_{h}-\bel{\expec}{\widetilde{M}_{t,\pi}^{\geq H+1}(s_{H+1},a_{H+1},\pi)}\left[\sum_{k=H+1}^{H}r_{k}\right]\right)$$

The latter sum is an empty sum, so it is zero, and this supersedes that $\widetilde{M}_{t,\pi}^{\geq H+1}(s_{H+1},a_{H+1},\pi)$ is ill-defined. As before, $\widetilde{M}_{t,\pi}^{\geq 0}(s_{0},a_{0},\pi)$ can be reexpressed as $\guess(\pi)$, so we are left with
$$=1+\eps_{\text{pess}}\left(\expec_{\guess(\pi)}\left[\sum_{k=0}^{H}r_{k}\right]-\sum_{h=0}^{H}r_{h}\right)$$

$\eps_{\text{pess}}=\sqrt{\frac{1}{T(H+1)^{2}}}$. Therefore, $\eps_{\text{pess}}\leq\frac{1}{H+1}$. Even in the worst possible case where the expected sum of rewards is 0, and the actual sum of rewards is $H+1$, this bet is non-negative. Because all bets are non-negative, and there are bettors whose bets are always bounded away from zero, the inductive proof goes through, and the estimator is well-defined. $\blacksquare$

\begin{lemm} \label{lem19}
For every calibration bettor $B$, with $1-\frac{\delta}{4(HS+1)}$ probability (over the true algorithm interacting with the true environment), we have
$$6\left(\ln\left(\frac{4(HS+1)}{\delta}\right)+\ln(2(|\mathcal{B}_{\text{cal}}|+|\mathcal{B}_{\text{frag}}|))+\tsum\ln\left(\bigstar_{t}\right)\right)+2\tsum\bel{\expec}{\pi,s_{h_{B}}\sim\xi_{t}}\left[X^{B}_{t,\pi}\cdot\textbf{1}_{s_{h_{B}}=s_{B}}\right]$$
$$\geq\tsum\bel{\expec}{\pi\sim p_{t}}\left[X^{B}_{t,\pi}\cdot\mathbb{P}_{\guess(\pi)}((h,s)_{B})\right]$$
Where $p_{t}$ is the distribution over policies on episode $t$ and $\xi_{t}$ is the joint distribution over policies and trajectories on episode $t$ according to the true environment.
\end{lemm}

We use $\xi_{t}$ as an abbreviation for the distribution over policies $\pi_{t}$ and trajectories $tr_{t}$, which is sampled from on the $t$'th episode. It depends on the true algorithm of the agent and the true environment it interacts with.

Let $B$ be a calibration bettor. Via Lemma A.4 of \cite{Foster21}, we have that, with $1-\frac{\delta}{4(HS+1)}$ probability according to the true algorithm interacting with the true environment, we have
$$\ln\left(\frac{4(HS+1)}{\delta}\right)+\tsum\ln(\text{bet}_{t,\pi_{t}}^{B}(tr_{t}))\geq-\tsum\ln\left(\bel{\expec}{\pi,tr\sim\xi_{t}}\left[e^{-\ln(\text{bet}_{t,\pi}^{B}(tr))}\right]\right)$$

Unpacking the overall bet for calibration bettors, which was analyzed in Lemma \ref{lem18}, yields
$$=-\tsum\ln\left(\expec_{\xi_{t}}\left[\frac{1}{\text{bet}_{t,\pi}^{B}(tr)}\right]\right)=-\tsum\ln\left(\expec_{\xi_{t}}\left[\frac{1}{1+\frac{1}{4}X^{B}_{t,\pi}\left(\mathbb{P}_{\guess(\pi)}((h,s)_{B})-\textbf{1}_{s_{h_{B}}=s_{B}}\right)}\right]\right)$$

When $x\in[-1,1]$, we have that $\frac{1}{1+\frac{1}{4}x}\leq 1-\frac{1}{4}x+\frac{1}{12}|x|$. Taking $x$ to be
\\
$X^{B}_{t,\pi}\left(\mathbb{P}_{\guess(\pi)}((h,s)_{B})-\textbf{1}_{s_{h_{B}}=s_{B}}\right)$, and upper-bounding $|x|$ by $X^{B}_{t,\pi}\left(\mathbb{P}_{\guess(\pi)}((h,s)_{B})+\textbf{1}_{s_{h_{B}}=s_{B}}\right)$, yields
$$\geq-\tsum\ln(\expec_{\xi_{t}}[1-\frac{1}{4}X^{B}_{t,\pi}\left(\mathbb{P}_{\guess(\pi)}((h,s)_{B})-\textbf{1}_{s_{h_{B}}=s_{B}}\right)$$
$$+\frac{1}{12}X^{B}_{t,\pi}(\mathbb{P}_{\guess(\pi)}((h,s)_{B})+\textbf{1}_{s_{h_{B}}=s_{B}})])$$
$$=-\tsum\ln\left(1-\expec_{\xi_{t}}\left[\frac{1}{6}X^{B}_{t,\pi}\cdot\mathbb{P}_{\guess(\pi)}((h,s)_{B})-\frac{1}{3}X^{B}_{t,\pi}\cdot\textbf{1}_{s_{h_{B}}=s_{B}}\right]\right)$$

By $-\ln(1-x)\geq x$ we have
$$\geq\tsum\expec_{\xi_{t}}\left[\frac{1}{6}X^{B}_{t,\pi}\cdot\mathbb{P}_{\guess(\pi)}((h,s)_{B})-\frac{1}{3}X^{B}_{t,\pi}\cdot\textbf{1}_{s_{h_{B}}=s_{B}}\right]$$

The expectation was over the policy and trajectory. However, $X^{B}_{t,\pi}$ and $\mathbb{P}_{\guess(\pi)}((h,s)_{B})$ don't depend on the trajectory, so we can rewrite as
$$=\sum_{t\leq T}\frac{1}{6}\bel{\expec}{\pi\sim p_{t}}\left[X^{B}_{t,\pi}\cdot\mathbb{P}_{\guess(\pi)}((h,s)_{B})\right]-\frac{1}{3}\expec_{\zeta_{t}}\left[X^{B}_{t,\pi}\cdot\textbf{1}_{s_{h_{B}}=s_{B}}\right]$$

Putting all the inequalities together, we have, for any calibration bettor $B$, with $1-\frac{\delta}{4(HS+1)}$ probability,
$$\ln\left(\frac{4(HS+1)}{\delta}\right)+\tsum\ln\left(\text{bet}_{t,\pi_{t}}^{B}(tr_{t})\right)$$
$$\geq\tsum\frac{1}{6}\bel{\expec}{\pi\sim p_{t}}\left[X^{B}_{t,\pi}\cdot\mathbb{P}_{\guess(\pi)}((h,s)_{B})\right]-\sum_{t\leq T}\frac{1}{3}\expec_{\zeta_{t}}\left[X^{B}_{t,\pi}\cdot\textbf{1}_{s_{h_{B}}=s_{B}}\right]$$

Multiply both sides by 6 and reshuffle.
\begin{equation} \label{eq:8}
6\left(\ln\left(\frac{4(HS+1)}{\delta}\right)+\tsum\ln\left(\text{bet}_{t,\pi_{t}}^{B}(tr_{t})\right)\right)+2\tsum\expec_{\zeta_{t}}\left[X^{B}_{t,\pi}\cdot\textbf{1}_{s_{h_{B}}=s_{B}}\right]
\end{equation}
$$\geq\tsum\bel{\expec}{\pi\sim p_{t}}\left[X^{B}_{t,\pi}\cdot\mathbb{P}_{\guess(\pi)}((h,s)_{B})\right]$$

Finally, by reshuffling Lemma \ref{lem6}, we have
$$\ln(\zeta_{T+1}(B))-\ln(\zeta_{1}(B))+\tsum\ln(\bigstar_{t})=\tsum\ln\left(\text{bet}_{t,\pi_{t}}^{B}(tr_{t})\right)$$

$\zeta_{T+1}(B)$ can be at most 1, and $\zeta_{1}(B)$, for calibration bettors, is known to be $\frac{1}{2(|\mathcal{B}_{\text{cal}}|+|\mathcal{B}_{\text{frag}}|)}$, yielding the inequality 
$$\ln(2(|\mathcal{B}_{\text{cal}}|+|\mathcal{B}_{\text{frag}}|))+\tsum\ln(\bigstar_{t})\geq\tsum\ln\left(\text{bet}_{t,\pi_{t}}^{B}(tr_{t})\right)$$

Which, when substituted into \ref{eq:8}, yields the desired result. $\blacksquare$

\begin{lemm} \label{lem20}
For all fragment bettors $B$ where a true RMDP $\imdp$ fulfills $\text{convert}(\imdp((h,s,a)_{B}))\subseteq\Psi_{B}$, with $1-\frac{\delta}{4(HS+1)}$ probability (over the true algorithm interacting with the true environment), we have $\ln\left(\frac{4(HS+1)}{\delta}\right)+\ln(2(|\mathcal{B}_{\text{cal}}|+|\mathcal{B}_{\text{frag}}|))+\tsum\ln\left(\bigstar_{t}\right)$
\\
$\geq\tsum\bel{\expec}{\pi,s_{h_{B'}}\sim\xi_{t}}\left[X^{B'}_{t,\pi}\cdot\textbf{1}_{s_{h_{B'}}=s_{B'}}\right]$ for $h_{B}\geq 1$ and $\geq\tsum\bel{\expec}{\pi\sim p_{t}}\left[\hells(\guesspi((h,s,a)_{B})\to\Psi_{B})\right]$ for $h_{B}=0$, where $B'$ is the calibration bettor corresponding to $B$, $\xi_{t}$ is the joint distribution over the policy and trajectory on episode $t$, and $p_{t}$ is the distribution over the policy on episode $t$.
\end{lemm}

The proof of this lemma substantially follows the analysis of the estimation complexity in the proof of Theorem \ref{the2}, so we will gloss over arguments spelled out in more detail there, and only highlight meaningful differences. The implicit uses of Lemma \ref{lem4} in the proof of Theorem \ref{the2} are permissible because the "every transition kernel has full support" argument from Theorem \ref{the2} still holds in our setting by Lemma \ref{lem18}

Fix a fragment bettor $B$ with the property that, for a true RMDP $\mathbb{M}$, $\text{convert}(\imdp((h,s,a)_{B}))\subseteq\Psi_{B}$. Via Lemma A.4 of \cite{Foster21}, we have that, with $1-\frac{\delta}{4(HS+1)}$ probability,
$$\ln\left(\frac{4(HS+1)}{\delta}\right)+\tsum\ln(\text{bet}_{t,\pi_{t}}^{B}(tr_{t}))\geq-\tsum\ln\left(\bel{\expec}{\pi,tr\sim\xi_{t}}\left[e^{-\ln(\text{bet}_{t,\pi}^{B}(tr))}\right]\right)$$

$\xi_{t}$ is, as before, the joint distribution over policies and trajectories on episode $t$. Unpacking the definition of the bet for fragment bettors, as in Lemma \ref{lem18}, yields
$$=-\tsum\ln\left(\expec_{\xi_{t}}\left[\frac{1}{\text{bet}_{t,\pi}^{B}(tr)}\right]\right)=-\tsum\ln(\expec_{\xi_{t}}$$
$$\left[\frac{1}{\textbf{1}_{(s,a)_{h_{B}}=(s,a)_{B}}\left(\bel{\expec}{r'\sim r_{h_{B}}}\left[\sqrt{\frac{\mu^{B}_{t,\pi}(r',s_{h_{B}+1})}{\guesspi((h,s,a)_{B})(r',s_{h_{B}+1})}}\right]+\hells(\guesspi((h,s,a)_{B})\to\Psi_{B})\right)+\textbf{1}_{(s,a)_{h_{B}}\neq(s,a)_{B}}}\right])$$
$$\geq-\tsum\ln\left(\expec_{\xi_{t}}\left[\frac{1}{\textbf{1}_{(s,a)_{h_{B}}=(s,a)_{B}}\left(\bel{\expec}{r'\sim r_{h_{B}}}\left[\sqrt{\frac{\mu^{B}_{t,\pi}(r',s_{h_{B}+1})}{\guesspi((h,s,a)_{B})(r',s_{h_{B}+1})}}\right]\right)+\textbf{1}_{(s,a)_{h_{B}}\neq(s,a)_{B}}}\right]\right)$$
$$=-\tsum\ln\left(\expec_{\xi_{t}}\left[\textbf{1}_{(s,a)_{h_{B}}=(s,a)_{B}}\cdot\frac{1}{\bel{\expec}{r'\sim r_{h_{B}}}\left[\sqrt{\frac{\mu^{B}_{t,\pi}(r',s_{h_{B}+1})}{\guesspi((h,s,a)_{B})(r',s_{h_{B}+1})}}\right]}+\textbf{1}_{(s,a)_{h_{B}}\neq(s,a)_{B}}\right]\right)$$

By convexity of $\frac{1}{x}$, we have
$$\geq-\tsum\ln\left(\expec_{\xi_{t}}\left[\textbf{1}_{(s,a)_{h_{B}}=(s,a)_{B}}\cdot\bel{\expec}{r'\sim r_{h}}\left[\frac{1}{\sqrt{\frac{\mu^{B}_{t,\pi}(r',s_{h_{B}+1})}{\guesspi((h,s,a)_{B})(r',s_{h_{B}+1})}}}\right]+\textbf{1}_{(s,a)_{h_{B}}\neq (s,a)_{B}}\right]\right)$$
$$=-\tsum\ln\left(\expec_{\xi_{t}}\left[\textbf{1}_{(s,a)_{h_{B}}=(s,a)_{B}}\cdot\bel{\expec}{r'\sim r_{h}}\left[\sqrt{\frac{\guesspi((h,s,a)_{B})(r',s_{h_{B}+1})}{\mu^{B}_{t,\pi}(r',s_{h_{B}+1})}}\right]+\textbf{1}_{(s,a)_{h_{B}}\neq (s,a)_{B}}\right]\right)$$

The outer expectation can be thought of as two expectations. One is over the $\pi$ and partial trajectory $tr'$ extending up to the action on turn $h_{B}$. The other is over $r_{h_{B}},s_{h_{B}+1}$. We use $\sigma^{tr'}_{t,\pi}$ to denote this latter distribution, if we play $\pi$ and have partial trajectory $tr'$ on episode $t$. This inner expectation can be moved past the indicator functions to yield
$$=-\tsum\ln\left(\bel{\expec}{\pi,tr'\sim\xi_{t}}\left[\textbf{1}_{(s,a)_{h_{B}}=(s,a)_{B}}\cdot\bel{\expec}{r,s'\sim\sigma_{t,\pi}^{tr'}}\left[\bel{\expec}{r'\sim r}\left[\sqrt{\frac{\guesspi((h,s,a)_{B})(r',s')}{\mu^{B}_{t,\pi}(r',s')}}\right]\right]+\textbf{1}_{(s,a)_{h_{B}}\neq (s,a)_{B}}\right]\right)$$

Now, sampling an $r,s'$, and then sampling either 0 or 1 with $r$ probability, is the same as sampling an $r',s'$ from $\text{convert}(\sigma^{tr'}_{t,\pi})$.
$$=-\tsum\ln\left(\bel{\expec}{\pi,tr'\sim\xi_{t}}\left[\textbf{1}_{(s,a)_{h_{B}}=(s,a)_{B}}\cdot\bel{\expec}{r',s'\sim\text{convert}(\sigma_{t,\pi}^{tr'})}\left[\sqrt{\frac{\guesspi((h,s,a)_{B})(r',s')}{\mu^{B}_{t,\pi}(r',s')}}\right]+\textbf{1}_{(s,a)_{h_{B}}\neq (s,a)_{B}}\right]\right)$$

Because $\sigma^{tr'}_{t,\pi}$, for all $t,\pi,tr'$, was chosen by the true environment, and $\imdp$ was a true RMDP (the environment respects its constraints), and this expectation is multiplied by an indicator function for being at $(h,s,a)_{B}$, when the indicator function is 1, $\sigma^{tr'}_{t,\pi}$ is guaranteed to be a selection from $\imdp((h,s,a)_{B})$. Therefore, all of the $\text{convert}(\sigma_{t,\pi}^{tr'})$ distributions lie in $\text{convert}(\imdp(h,s,a)_{B})$. By assumption, this set was a subset of $\Psi_{B}$. Because the distributions are all within our set of interest $\Psi_{B}$, we can apply the argument from the analysis of estimation complexity in Theorem \ref{the2} to yield
$$\geq-\tsum\ln\left(\expec_{\xi_{t}}\left[\textbf{1}_{(s,a)_{h_{B}}=(s,a)_{B}}\cdot(1-\hells(\guesspi((h,s,a)_{B})\to\Psi_{B}))+\textbf{1}_{(s,a)_{h_{B}}\neq(s,a)_{B}}\right]\right)$$
$$=-\tsum\ln\left(\expec_{\xi_{t}}\left[1-\textbf{1}_{(s,a)_{h_{B}}=(s,a)_{B}}\cdot\hells(\guesspi((h,s,a)_{B})\to\Psi_{B})\right]\right)$$
$$=-\tsum\ln\left(1-\bel{\expec}{\pi,s_{h_{B}},a_{h_{B}}\sim\xi_{t}}\left[\textbf{1}_{s_{h_{B}}=s_{B}}\cdot\textbf{1}_{a_{h_{B}}=a_{B}}\cdot\hells(\guesspi((h,s,a)_{B})\to\Psi_{B})\right]\right)$$

Now, the trajectory extends up to the action on turn $h_{B}$. It can be viewed as a trajectory which extends up to the state $s_{h_{B}}$, and the last action being selected via $\pi(h_{B},s_{h_{B}})$. The expectation over this last action can be pushed past the Hellinger term and the indicator function over states, to yield the probability of taking the appropriate action.
$$=-\tsum\ln\left(1-\bel{\expec}{\pi,s_{h_{B}}\sim\xi_{t}}\left[\textbf{1}_{s_{h_{B}}=s_{B}}\cdot\pi(h_{B},s_{B})(a_{B})\cdot\hells(\guesspi((h,s,a)_{B})\to\Psi_{B})\right]\right)$$

For $h_{B}=0$, we can use that the indicator function for the correct state is always 1, as is the probability of the correct action, because only one state and action are possible. Also, there is no dependence on the trajectory, so we have
$$=-\tsum\ln\left(1-\bel{\expec}{\pi\sim p_{t}}\left[\hells(\guesspi((h,s,a)_{B})\to\Psi_{B})\right]\right)$$

For $h\geq 1$, we can use that $B'$, the calibration bettor corresponding to the fragment bettor $B$, has $(h,s,a)_{B'}=(h,s,a)_{B}$ and $\Psi_{B}=\Psi_{B'}$, so we can apply the definition of $X^{B'}_{t,\pi}$ to yield
$$=-\tsum\ln\left(1-\bel{\expec}{\pi,s_{h_{B'}}\sim\xi_{t}}\left[X^{B'}_{t,\pi}\cdot\textbf{1}_{s_{h_{B'}}=s_{B'}}\right]\right)$$

By $-\ln(1-x)\geq x$ in both cases, for $h_{B}=0$ we have
$$\geq\tsum\bel{\expec}{\pi\sim p_{t}}\left[\hells(\guesspi((h,s,a)_{B})\to\Psi_{B})\right]$$

And for $h_{B}\geq 1$ we have
$$\geq\tsum\bel{\expec}{\pi,s_{h_{B'}}\sim\xi_{t}}\left[X^{B'}_{t,\pi}\cdot\textbf{1}_{s_{h_{B'}}=s_{B'}}\right]$$

Our net inequality is, for $h_{B}\geq 1$,
$$\ln\left(\frac{4(HS+1)}{\delta}\right)+\tsum\ln\left(\text{bet}_{t,\pi_{t}}^{B}(tr_{t})\right)\geq\tsum\bel{\expec}{\pi,s_{h_{B'}}\sim\xi_{t}}\left[X^{B'}_{t,\pi}\cdot\textbf{1}_{s_{h_{B'}}=s_{B'}}\right]$$

And for $h_{B}=0$,
$$\ln\left(\frac{4(HS+1)}{\delta}\right)+\tsum\ln\left(\text{bet}_{t,\pi_{t}}^{B}(tr_{t})\right)\geq\tsum\bel{\expec}{\pi\sim p_{t}}\left[\hells(\guesspi((h,s,a)_{B})\to\Psi_{B})\right]$$

To conclude in both cases, by reshuffling Lemma \ref{lem6}, we have
$$\ln(\zeta_{T+1}(B))-\ln(\zeta_{1}(B))+\tsum\ln(\bigstar_{t})=\tsum\ln\left(\text{bet}_{t,\pi_{t}}^{B}(tr_{t})\right)$$

$\zeta_{T+1}(B)$ can be at most 1, and $\zeta_{1}(B)$, for fragment bettors, is $\frac{1}{2(|\mathcal{B}_{\text{cal}}|+|\mathcal{B}_{\text{frag}}|)}$ by construction yielding the inequality 
$$\ln(2(|\mathcal{B}_{\text{cal}}|+|\mathcal{B}_{\text{frag}}|))+\tsum\ln(\bigstar_{t})\geq\tsum\ln\left(\text{bet}_{t,\pi_{t}}^{B}(tr_{t})\right)$$

Which, substituted into the previous equations, yields the desired result. $\blacksquare$

\begin{lemm} \label{lem21}
With $1-\frac{\delta}{2}$ probability, $\ln\left(\frac{2}{\delta}\right)\geq\tsum\ln(\bigstar_{t})$.
\end{lemm}

By Lemma A.4 in \cite{Foster21}, with $1-\frac{\delta}{2}$ probability, we have
$$\tsum\ln(\bigstar_{t})\leq\ln\left(\frac{2}{\delta}\right)+\tsum\ln\left(\expec_{\xi_{t}}\left[e^{\ln(\bigstar_{t})}\right]\right)=\ln\left(\frac{2}{\delta}\right)+\tsum\ln(\expec_{\xi_{t}}\left[\bigstar_{t}\right])$$

Again, $\xi_{t}$ is the distribution over the policy and the trajectory on episode $t$. Now, by applying definitions and expectation-shuffling, we have
$$\expec_{\xi_{t}}[\bigstar_{t}]=\expec_{\xi_{t}}\left[\bel{\expec}{B\sim\zeta_{t}}\left[\text{bet}^{B}_{t,\pi}(r_{0},(s,a,r)_{1:H})\right]\right]$$
$$=\expec_{\xi_{t}}\left[\bel{\expec}{B\sim\zeta_{t}}\left[1+\sum_{h=0}^{H}\left(\text{lbet}_{t,\pi}^{B,h,s_{h},a_{h}}(r_{h},s_{h+1},a_{h+1})-1\right)\right]\right]$$
\begin{equation} \label{eq:9}
=1+\sum_{h=0}^{H}\expec_{\xi_{t}}\left[\bel{\expec}{B\sim\zeta_{t}}\left[\text{lbet}_{t,\pi}^{B,h,s_{h},a_{h}}(r_{h},s_{h+1},a_{h+1})\right]-1\right]
\end{equation}

Letting $\nu_{r,s'}:\Delta(\{0,1\}\times\states)$ be the distribution which assigns $r$ measure to $1,s'$ and the rest to $0,s'$, define $\text{res}_{t,\pi}^{B,h,s,a}(r,s',a')$ (a "residual" term) as
$$\text{res}_{t,\pi}^{B,h,s,a}(r,s',a'):=\text{lbet}_{t,\pi}^{B,h,s,a}(r,s',a')-1+d(g_{t,\pi}^{B,h,s,a})_{\guesspi(h,s,a)}(\guesspi(h,s,a)-\nu_{r,s'})$$

That term is the Frechet derivative of the $g$ function at $\guesspi(h,s,a)$, in a certain direction. With this definition, we can re-express \ref{eq:9} as
$$=1+\sum_{h=0}^{H}\expec_{\xi_{t}}[\bel{\expec}{B\sim\zeta_{t}}[\text{res}_{t,\pi}^{B,h,s_{h},a_{h}}(r_{h},s_{h+1},a_{h+1})$$
$$-d(g_{t,\pi}^{B,h,s_{h},a_{h}})_{\guesspi(h,s_{h},a_{h})}(\guesspi(h,s_{h},a_{h})-\nu_{r_{h},s_{h+1}})]]$$
$$=1+\sum_{h=0}^{H}\expec_{\xi_{t}}\left[\bel{\expec}{B\sim\zeta_{t}}[\text{res}_{t,\pi}^{B,h,s_{h},a_{h}}(r_{h},s_{h+1},a_{h+1})\right]$$
$$-\bel{\expec}{B\sim\zeta_{t}}\left[d(g_{t,\pi}^{B,h,s_{h},a_{h}})_{\guesspi(h,s_{h},a_{h})}(\guesspi(h,s_{h},a_{h})-\nu_{r_{h},s_{h+1}})\right]]$$

By linearity of differentiation, we get
$$=1+\sum_{h=0}^{H}\expec_{\xi_{t}}[\bel{\expec}{B\sim\zeta_{t}}\left[\text{res}_{t,\pi}^{B,h,s_{h},a_{h}}(r_{h},s_{h+1},a_{h+1})\right]$$
$$-d\left(\bel{\expec}{B\sim\zeta_{t}}\left[g_{t,\pi}^{B,h,s_{h},a_{h}}\right]\right)_{\guesspi(h,s_{h},a_{h})}(\guesspi(h,s_{h},a_{h})-\nu_{r_{h},s_{h+1}})]$$

Now, $\guesspi(h,s_{h},a_{h})$ was picked to be the minimizer of $\bel{\expec}{B\sim\zeta_{t}}\left[g^{B,h,s_{h},a_{h}}_{t,\pi}\right]$, a Frechet-differentiable convex function, so the derivative in every direction is zero. Interchanging expectations, we have
\begin{equation} \label{eq:10}
=1+\sum_{h=0}^{H}\bel{\expec}{B\sim\zeta_{t}}\left[\expec_{\xi_{t}}\left[\text{res}_{t,\pi}^{B,h,s_{h},a_{h}}(r_{h},s_{h+1},a_{h+1})\right]\right]
\end{equation}

Now we compute the residuals. By routine but tedious calculations of derivatives similar to those we did for the robust universal estimator in Theorem \ref{the2}, the residual for the fragment bettors is zero.
For the calibration bettors, the residuals are
$$\text{res}_{t,\pi}^{B,h,s,a}(r,s',a')=\frac{X^{B}_{t,\pi}}{4}\left(\mathbb{P}_{\guess(\pi)|h+1,s'}((h,s)_{B})-\mathbb{P}_{\guess(\pi)|h+1,s',a'}((h,s)_{B})\right)$$

And for the pessimism bettor, the residual is
$$\text{res}_{t,\pi}^{B,h,s,a}(r,s',a')=\eps_{\text{pess}}\left(\bel{\expec}{\guess(\pi)|h+1,s'}\left[\sum_{k=h+1}^{H}r_{k}\right]-\bel{\expec}{\guess(\pi)|h+1,s',a'}\left[\sum_{k=h+1}^{H}r_{k}\right]\right)$$

Importantly, in both cases, the residuals are zero in expectation, and we have
$$\bel{\expec}{a'\sim\pi(h+1,s')}\left[\text{res}_{t,\pi}^{B,h,s,a}(r,s',a')\right]=0$$

We can expand \ref{eq:10} and break it up to yield
$$=1+\sum_{h=0}^{H}\bel{\expec}{B\sim\zeta_{t}}\left[\bel{\expec}{\pi,(s,a,r)_{h},s_{h+1},a_{h+1}\sim\xi_{t}}\left[\text{res}_{t,\pi}^{B,h,s_{h},a_{h}}(r_{h},s_{h+1},a_{h+1})\right]\right]$$
$$=1+\sum_{h=0}^{H}\bel{\expec}{B\sim\zeta_{t}}\left[\bel{\expec}{\pi,(s,a,r)_{h},s_{h+1}\sim\xi_{t}}\left[\bel{\expec}{a_{h+1}\sim\pi(h+1,s_{h+1})}\left[\text{res}_{t,\pi}^{B,h,s_{h},a_{h}}(r_{h},s_{h+1},a_{h+1})\right]\right]\right]=1$$

Chaining our equalities together, we have $\expec_{\xi_{t}}[\bigstar_{t}]=1$, which, substituting into the starting inequality,  yields
$$\ln\left(\frac{2}{\delta}\right)\geq\tsum\ln(\bigstar_{t})$$
$\blacksquare$

\begin{lemm} \label{lem22}
The function $\expec_{B\sim\zeta_{t}}\left[g^{B,h,s,a}_{t,\pi}(\mu)\right]$ has a unique minimum.
\end{lemm}

First, any strictly convex function has a unique minimum, because if the set of minimizers contained two distinct points, a 50/50 mix of the minimizers would attain a strictly lower value, producing a contradiction. So, we just need to show that the mixture of $g$ functions is strictly convex.

Second, a mixture of convex functions with one strictly convex function that has nonzero probability, is also strictly convex, because we get a non-strict inequality $\geq$ for the other convex functions, and a strict inequality for the strictly convex function. So, if we can verify that all of the $g$ functions are convex, and that there is a $g$ function corresponding to a bettor which retains nonzero probability, which is strictly convex, our desired result will follow.

The $g$ functions associated with the calibration bettors and pessimism bettors are linear in $\mu$, and the $g$ functions associated with the fragment bettors and uniform bettor are convex in $\mu$, by Lemma 4. To verify strict convexity for the uniform bettor, we may let $\nu,\nu'$ be unequal, and $p\in(0,1)$, and use that the set $\Psi$ contains only a single point, the uniform distribution $u$. We then have
$$\hells(p\nu+(1-p)\nu',u)=1-\sum_{o,r}\sqrt{(p\nu(o,r)+(1-p)\nu'(o,r))\cdot u(o,r)}$$
$$=1-\sum_{o,r}\sqrt{(p\nu(o,r)u(o,r)+(1-p)\nu'(o,r)u(o,r)}$$

The negative square root function is strictly convex. There is some $o,r$ where $\nu(o,r)\neq\nu'(o,r)$ because $\nu\neq\nu'$, and $u(o,r)$ is nonzero because it is the uniform distribution, so we have a strict inequality for one $o,r$ and a nonstrict inequality for the others, yielding
$$<1-p\sum_{o,r}\sqrt{\nu(o,r)u(o,r)}-(1-p)\sum_{o,r}\sqrt{\nu'(o,r)u(o,r)}=p\hells(\nu,u)+(1-p)\hells(\nu',u)$$

Because $\nu,\nu',p$ were arbitrary, strict convexity of the bet made by the uniform bettor is verified. The uniform bettors always retain nonzero probability by the inductive argument from the end of Proposition \ref{pro4}, although we must also use that all $\bigstar_{t}>0$ (proved in Lemma \ref{lem18}) to generalize the argument. Therefore, the function being minimized is a mixture of convex functions and a strictly convex function with nonzero probability, so it is strictly convex, so the minimum is unique. $\blacksquare$
\section{RMDP Estimator Theorems}
\begin{theo} \label{the5}
There is an online estimator $\estim$ where $\beta_{\estim}(T,\delta)\in\mathcal{O}\left(HS^{2}\log(T)+HS\log\left(\frac{HSAT}{\delta}\right)\right)$ and $\alpha_{\estim}(T,\delta)\in\mathcal{O}\left(H\sqrt{T}\log\left(\frac{1}{\delta}\right)\right)$, for hypotheses in $\hypo_{\text{parhalf}}$.
\end{theo}

We use the estimator defined in Appendix L, with $\eps'$ being arbitrarily small. By Lemma \ref{lem18}, it is well-defined. The parameters $\eps_{\states}$ and $\eps_{[0,1]}$ are taken as unspecified for now, and chosen later on in the proof to minimize the estimation error.

We begin by bounding $\beta_{\estim}$ for the hypothesis class $\hypo_{\text{parhalf}}$. Note that, to retain compatibility with our nonstandard notion of estimation complexity, we must use the associated definition of estimation error used in the previous theorems about the value of the modified DEC, so if $\imdp\in\hypo_{\text{parhalf}}$, we must upper-bound the following term with $1-\delta$ probability.
$$\tsum\bel{\expec}{\pi\sim p_{t}}\left[\bel{\expec}{r_{0},(s,a,r)_{1:H}\sim\guess(\pi)}\left[\sum_{h=0}^{H}\bel{\expec}{a\sim\pi(h,s_{h})}\left[\hellsc\left((\guess(\pi)|h,s_{h},a)_{\downarrow h}\to\imdp(h,s_{h},a)\right)\right]\right]\right]$$

First, observe that $(\guess(\pi)|h,s_{h},a)_{\downarrow h}$ is the distribution over the upcoming reward and state $r_{h},s_{h+1}$ if $h,s_{h},a$ occurs. Because $\guess(\pi)$ (our estimator) was defined from $\guesspi$ (our MDP), this is just $\guesspi(h,s_{h},a)$.
$$=\tsum\bel{\expec}{\pi\sim p_{t}}\left[\expec_{\guess(\pi)}\left[\sum_{h=0}^{H}\bel{\expec}{a\sim\pi(h,s_{h})}\left[\hellsc\left(\guesspi(h,s_{h},a)\to\imdp(h,s_{h},a)\right)\right]\right]\right]$$

Pulling the sum over $h$ out, we have
$$=\sum_{h=0}^{H}\tsum\bel{\expec}{\pi\sim p_{t}}\left[\expec_{\guess(\pi)}\left[\bel{\expec}{a\sim\pi(h,s_{h})}\left[\hellsc\left(\guesspi(h,s_{h},a)\to\imdp(h,s_{h},a)\right)\right]\right]\right]$$

Because $\expec_{\guess(\pi)}$ only determines $s_{h}$, we can write this as a sum over $s$ of the probability of $(h,s)$, and pull the sum out again, to yield
$$=\sum_{h=0}^{H}\sum_{s\in\states}\tsum\bel{\expec}{\pi\sim p_{t}}\left[\mathbb{P}_{\guess(\pi)}(h,s)\cdot\bel{\expec}{a\sim\pi(h,s)}\left[\hellsc\left(\guesspi(h,s,a)\to\imdp(h,s,a)\right)\right]\right]$$

Now, because $\imdp\in\hypo_{\text{parhalf}}$ (the set of RMDP's), it was created from some PRMDPR in $\hypo_{\text{parhalf}}$ (the set of PRMDPR's). For this PRMDPR, every $h,s$ is associated with some recommended action $a$, and an imprecise belief indexed by an $f:\states\to[0,1]$ and a $c:[0,1]$. We use $a_{h,s},f_{h,s},c_{h,s}$ to denote these. By how we go from PRMDPR's to RMDP's, this means that if $a\neq a_{h,s}$, 
\\
$\imdp(h,s,a)=\Delta([0,1]\times\states)$, but $\imdp(h,s,a_{h,s})=\Psi^{[0,1]}_{f_{h,s},c_{h,s}}$. This yields
$$=\sum_{h=0}^{H}\sum_{s\in\states}\tsum\bel{\expec}{\pi\sim p_{t}}\left[\mathbb{P}_{\guess(\pi)}(h,s)\cdot\pi(h,s)(a_{h,s})\cdot\hellsc\left(\guesspi(h,s,a_{h,s})\to\Psi^{[0,1]}_{f_{h,s},c_{h,s}}\right)\right]$$

$\guesspi(h,s,a_{h,s})$ was already supported over 0 or 1 reward by its construction, so we may invoke Lemma \ref{lem16} to go
$$=\sum_{h=0}^{H}\sum_{s\in\states}\tsum\bel{\expec}{\pi\sim p_{t}}\left[\mathbb{P}_{\guess(\pi)}(h,s)\cdot\pi(h,s)(a_{h,s})\cdot\hells\left(\guesspi(h,s,a_{h,s})\to\Psi^{\{0,1\}}_{f_{h,s},c_{h,s}}\right)\right]$$

Now, given some $f:\states\to[0,1]$, we use $\lceil f\rceil$ to denote the function $\lambda s.1-\left\lfloor\frac{1-f(s)}{\eps_{\states}}\right\rfloor\cdot\eps_{\states}$, and given some $c:[0,1]$, we use $\lfloor c\rfloor$ to denote $\left\lfloor\frac{c}{\eps_{[0,1]}}\right\rfloor\cdot\eps_{[0,1]}$. Intuitively, the hypothesis space $\hypo'_{\text{mid}}$ consists of the $\Psi_{f,c}$ where the values of $f$ and $c$ are discretized in this way, so we are rounding up and down as needed to find the closest discrete approximation to $f,c$. By Lemma \ref{lem5}, and linearity of expectation, we get
\begin{equation} \label{eq:11}
\leq 2\sum_{h=0}^{H}\sum_{s\in\states}\tsum\bel{\expec}{\pi\sim p_{t}}\left[\mathbb{P}_{\guess(\pi)}(h,s)\cdot\pi(h,s)(a_{h,s})\cdot\hells\left(\guesspi(h,s,a_{h,s})\to\Psi^{\{0,1\}}_{\lceil f_{h,s}\rceil,\lfloor c_{h,s}\rfloor}\right)\right]
\end{equation}
$$+2\sum_{h=0}^{H}\sum_{s\in\states}\tsum\bel{\expec}{\pi\sim p_{t}}\left[\mathbb{P}_{\guess(\pi)}(h,s)\cdot\pi(h,s)(a_{h,s})\cdot\hells\left(\Psi^{\{0,1\}}_{\lceil f_{h,s}\rceil,\lfloor c_{h,s}\rfloor}\to\Psi^{\{0,1\}}_{f_{h,s},c_{h,s}}\right)\right]$$

Focusing on that second term, we can rewrite it as
$$2\tsum\bel{\expec}{\pi\sim p_{t}}\left[\sum_{h=0}^{H}\bel{\expec}{s_{h}\sim\guess(\pi)}\left[\pi(h,s_{h})(a_{h,s_{h}})\cdot\hells\left(\Psi^{\{0,1\}}_{\lceil f_{h,s_{h}}\rceil,\lfloor c_{h,s_{h}}\rfloor}\to\Psi^{\{0,1\}}_{f_{h,s_{h}},c_{h,s_{h}}}\right)\right]\right]$$
$$\leq 2\tsum\bel{\expec}{\pi\sim p_{t}}\left[\sum_{h=0}^{H}\bel{\expec}{s_{h}\sim\guess(\pi)}\left[\hells\left(\Psi^{\{0,1\}}_{\lceil f_{h,s_{h}}\rceil,\lfloor c_{h,s_{h}}\rfloor}\to\Psi^{\{0,1\}}_{f_{h,s_{h}}, c_{h,s_{h}}}\right)\right]\right]$$

By Lemma \ref{lem17}, we may upper-bound this by
$$\leq 2T(H+1)(\eps_{\states}+\eps_{[0,1]})$$

Substituting this back in to \ref{eq:11} our upper bound on the estimation complexity is now
$$\leq 2\sum_{h=0}^{H}\sum_{s\in\states}\tsum\bel{\expec}{\pi\sim p_{t}}\left[\mathbb{P}_{\guess(\pi)}(h,s)\cdot\pi(h,s)(a_{h,s})\cdot\hells\left(\guesspi(h,s,a_{h,s})\to\Psi^{\{0,1\}}_{\lceil f_{h,s}\rceil,\lfloor c_{h,s}\rfloor}\right)\right]$$
$$+2T(H+1)(\eps_{\states}+\eps_{[0,1]})$$

We now focus on the first term. For every $h,s$, $\Psi^{\{0,1\}}_{\lceil f_{h,s}\rceil,\lfloor c_{h,s}\rfloor}\in\hypo'_{\text{mid}}$. Therefore, for every $h,s$ with $h\geq 1$, the tuple $h,s,a_{h,s},\Psi^{\{0,1\}}_{\lceil f_{h,s}\rceil,\lfloor c_{h,s}\rfloor}$ picks out a corresponding fragment bettor $B^{\text{frag}}_{h,s}$, and calibration bettor $B^{\text{cal}}_{h,s}$. If $h=0$, only a fragment bettor is picked out, so we can rewrite our upper bound as
$$=2\tsum\bel{\expec}{\pi\sim p_{t}}\left[\mathbb{P}_{\guess(\pi)}(0,s_{0})\cdot\pi(h_{0},s_{0})(a_{0})\cdot\hells\left(\guesspi(0,s_{0},a_{0})\to\Psi_{B^{\text{frag}}_{h,s}}\right)\right]$$
$$+2\sum_{h=1}^{H}\sum_{s\in\states}\tsum\bel{\expec}{\pi\sim p_{t}}\left[\mathbb{P}_{\guess(\pi)}(h,s)\cdot\pi(h,s)(a_{h,s})\cdot\hells\left(\guesspi(h,s,a_{h,s})\to\Psi_{B^{\text{cal}}_{h,s}}\right)\right]$$
$$+2T(H+1)(\eps_{\states}+\eps_{[0,1]})$$

For the first term, $\mathbb{P}_{\guess(\pi)}(0,s_{0})$ and $\pi(h_{0},s_{0})(a_{0})$ are both 1 because only one state and action is possible. For the second term, we pack up the definition of $X^{B^{\text{cal}}_{h,s}}_{t,\pi}$, and use that $(h,s)=(h,s)_{B^{\text{cal}}_{h,s}}$, to yield
$$=2\tsum\bel{\expec}{\pi\sim p_{t}}\left[\hells\left(\guesspi(0,s_{0},a_{0})\to\Psi_{B^{\text{frag}}_{h,s}}\right)\right]$$
$$+2\sum_{h=1}^{H}\sum_{s\in\states}\tsum\bel{\expec}{\pi\sim p_{t}}\left[X^{B^{\text{cal}}_{h,s}}_{t,\pi}\cdot\mathbb{P}_{\guess(\pi)}((h,s)_{B^{\text{cal}}_{h,s}})\right]+2T(H+1)(\eps_{\states}+\eps_{[0,1]})$$

By Lemma \ref{lem19}, and the union bound over $h,s$, we have that with $\geq 1-\frac{\delta}{4}$ probability, we have an upper bound of
$$\leq 2\tsum\bel{\expec}{\pi\sim p_{t}}\left[\hells\left(\guesspi(0,s_{0},a_{0})\to\Psi_{B^{\text{frag}}_{h,s}}\right)\right]$$
$$+12HS\left(\ln\left(\frac{4(HS+1)}{\delta}\right)+\ln(2(|\mathcal{B}_{\text{cal}}|+|\mathcal{B}_{\text{frag}}|))+\tsum\ln(\bigstar_{t})\right)$$
$$+4\sum_{h=1}^{H}\sum_{s\in\states}\tsum\expec_{\xi_{t}}\left[X^{B^{\text{cal}}_{h,s}}_{t,\pi}\cdot\textbf{1}_{s_{h}=s}\right]+2T(H+1)(\eps_{\states}+\eps_{[0,1]})$$

From earlier in this theorem, we had that, for the true $\imdp$ in $\hypo_{\text{parhalf}}$, for all $h,s$, 
\\
$\imdp(h,s,a_{h,s})=\Psi^{[0,1]}_{f_{h,s},c_{h,s}}$, by the definition of $a_{h,s},f_{h,s},c_{h,s}$. By the proof of Lemma \ref{lem16}, (specifically, the result that converting a $\Psi^{[0,1]}_{f,c}$ set just swaps the type signature in the superscript), this yields $\text{convert}(\imdp(h,s,a_{h,s}))=\Psi^{\{0,1\}}_{f_{h,s},c_{h,s}}$. Also, by the we rounded the function up and the constant down, along with the definitions of the $\Psi^{\{0,1\}}_{f,c}$ beliefs, and the way we defined our fragment bettors of interest, we have $\Psi^{\{0,1\}}_{f_{h,s},c_{h,s}}\subseteq\Psi^{\{0,1\}}_{\lceil f_{h,s}\rceil,\lfloor c_{h,s}\rfloor}=\Psi_{B^{\text{frag}}_{h,s}}$. Combining these, we have $\text{convert}(\imdp(h,s,a_{h,s}))\subseteq\Psi_{B^{\text{frag}}_{h,s}}$. This lets us invoke Lemma \ref{lem20}. By Lemma \ref{lem20} and the union bound over the various $h,s$, we get that with $1-\frac{\delta}{2}$ probability, our upper bound is now
\begin{equation} \label{eq:12}
\leq(16HS+2)\left(\ln\left(\frac{4(HS+1)}{\delta}\right)+\ln(2(|\mathcal{B}_{\text{cal}}|+|\mathcal{B}_{\text{frag}}|))+\tsum\ln(\bigstar_{t})\right)
+2T(H+1)(\eps_{\states}+\eps_{[0,1]})
\end{equation}

Then, by Lemma \ref{lem21}, and another application of the union bound, we get that with $1-\delta$ probability, our upper bound on the estimation complexity is now
$$\leq(16HS+2)\left(\ln\left(\frac{4(HS+1)}{\delta}\right)+\ln(2(|\mathcal{B}_{\text{cal}}|+|\mathcal{B}_{\text{frag}}|))+\ln\left(\frac{2}{\delta}\right)\right)+2T(H+1)(\eps_{\states}+\eps_{[0,1]})$$

Now we compute the size of our hypothesis space. By the definitions of the sets of traders and our space of hypothesis fragments, we have 
$$|\mathcal{B}|_{\text{cal}}<|\mathcal{B}_{\text{frag}}|\leq (HSA+1)|\hypo'_{\text{mid}}|\leq(HSA+1)\left(\frac{1}{\eps_{\states}}+1\right)^{S}\left(\frac{1}{\eps_{[0,1]}}+1\right)$$

Therefore, we have
$$\ln\left(2(|\mathcal{B}_{\text{frag}}|+|\mathcal{B}|_{\text{cal}})\right)<\ln\left(2(HSA+1)\left(\frac{1}{\eps_{\states}}+1\right)^{S}\left(\frac{1}{\eps_{[0,1]}}+1\right)\right)$$
$$=\ln(2(HSA+1))+\ln\left(\frac{1}{\eps_{[0,1]}}+1\right)+S\ln\left(\frac{1}{\eps_{\states}}+1\right)$$

Combining this with \ref{eq:12}, merging some logarithms together, and using that the $\imdp$, selection, algorithm choosing the policy, $\delta$, and $T$ were arbitrary, with $1-\delta$ probability, the estimation complexity for $\hypo_{\text{parhalf}}$ is upper-bounded by
$$\leq(16HS+2)\left(\ln\left(\frac{1}{\eps_{[0,1]}}+1\right)+S\ln\left(\frac{1}{\eps_{\states}}+1\right)+\ln\left(\frac{16(HSA+1)(HS+1)}{\delta^{2}}\right)\right)$$
$$+2T(H+1)(\eps_{\states}+\eps_{[0,1]})$$

Minimizing over the two parameters, a near-optimal value for $\eps_{\states}$ is $\frac{(8HS+1)S}{(H+1)T}$, and a near-optimal value for $\eps_{[0,1]}$ is $\frac{(8HS+1)}{(H+1)T}$. Our estimator $\estim$ run with these two parameters would make the term $2T(H+1)(\eps_{\states}+\eps_{[0,1]})$ simplify to $(16HS+2)(S+1)$. The values $\ln\left(\frac{1}{\eps_{[0,1]}}+1\right)$ and $\ln\left(\frac{1}{\eps_{\states}}+1\right)$ may both be approximated as $\mathcal{O}(\log(T))$, yielding a final estimation complexity on the order of
$$\mathcal{O}\left(HS^{2}\log(T)+HS\log\left(\frac{HSAT}{\delta}\right)\right)$$

And so, this is our value of $\beta_{\widehat{M}}(T,\delta)$. To bound $\alpha_{\widehat{M}}(T,\delta)$, we follow the corresponding segment of the proof of Theorem \ref{the2}, neglecting the arbitrarily low $\eps'$ in the same way. We have, from Lemma \ref{lem6}, rearranging, and using that around half of the probability mass was placed on the pessimism trader, in the $\eps'\to 0$ limit,
$$\tsum(\ln(\bigstar_{t}))+\ln(2)\geq\tsum\ln(\text{bet}_{t,\pi_{t}}^{B}(tr_{t}))$$

Reusing our argument in Lemma \ref{lem21} about the expected sum of $\ln(\bigstar_{t})$ being less than $\ln\left(\frac{2}{\delta}\right)$ with $1-\frac{\delta}{2}$ probability, and unpacking the bet for the pessimism trader, we get
$$\ln\left(\frac{4}{\delta}\right)\geq\tsum\ln\left(1+\eps_{\text{pess}}\left(\bel{\expec}{\guess(\pi_{t})}\left[\sum_{k=0}^{H}r_{k}\right]-\sum_{h=0}^{H}r_{h,t}\right)\right)$$

We may reuse the Azuma argument from Theorem \ref{the2} with $1-\frac{\delta}{2}$ probability (to have a net $1-\delta$ failure probability by the union bound). Reusing this argument requires making all necessary changes to account for the fact that the maximum value of the difference in reward is in $[0,H+1]$, not $[0,1]$. In order to apply our quadratic approximation to $\ln$, we need that $\sqrt{\frac{1}{T(H+1)^{2}}}(H+1)\leq\frac{1}{2}$, which holds for all $T\geq 4$. Adding 4 to the final bound will account for this. Proceeding with Theorem \ref{the2}'s Azuma argument, making necessary changes, our net upper bound ends up being
$$\leq\frac{\ln\left(\frac{4}{\delta}\right)}{\eps_{\text{pess}}}+T\cdot\eps_{\text{pess}}(H+1)^{2}+\sqrt{2\ln\left(\frac{2}{\delta}\right)T(H+1)^{2}}$$

Using that $\eps_{\text{pess}}=\sqrt{\frac{1}{T(H+1)^{2}}}$, factoring out appropriate terms, and adding the 4 onto the regret bound to account for the missing timesteps at the start, we are left with a net upper bound on $\alpha_{\estim}(T,\delta)$ of
$$(H+1)\sqrt{T}\left(\ln\left(\frac{4}{\delta}\right)+1+\sqrt{2\ln\left(\frac{2}{\delta}\right)}\right)+4$$

This bound is $\mathcal{O}\left(H\sqrt{T}\ln\left(\frac{1}{\delta}\right)\right)$, as desired, and $\alpha_{\widehat{M}}(T,\delta)$ and $\beta_{\widehat{M}}(T,\delta)$ for this estimator have been computed. $\blacksquare$

\begin{prop} \label{pro7}
For the estimator of Theorem 5, all estimates
\\
$\guess:\Pi_{RNS}\to\Delta\left([0,1]\times(\states\times\acts\times[0,1])^{[H]}\right)$ are continuous and policy-coherent.
\end{prop}

Policy-coherence holds because the estimate $\estim(\pi)$ was constructed via having $\pi$ interact with an MDP, which automatically produces policy-coherent estimates.

To show continuity of the estimator in $\pi$, we proceed as follows. The function $\lambda\pi.\estim(\pi)$ is an abbreviation for the function $\lambda\pi.\guess\bowtie\pi$, so we must show that this is a continuous function.

If $\lambda\pi.\guesspi(h,s,a)$ of type $\Pi_{RNS}\to\Delta(\{0,1\}\times\states)$ were continuous in $\pi$ for all $h,s,a$, then as $\pi_{n}$ converges to a limiting $\pi$, every trajectory $tr$ (and there are only finitely many) would have $\mathbb{P}_{M_{t,\pi_{n}}\bowtie\pi_{n}}(tr)$ limit to $\mathbb{P}_{\guesspi\bowtie\pi}(tr)$, because the probability of a trajectory can be written as a product of transition probabilities for the environment and the policy. All of the finitely many transition probabilities would converge as $\pi_{n}$ converges to $\pi$, leading to convergence of the probability of every trajectory, so the distribution $M_{t,\pi_{n}}\bowtie\pi_{n}$ would converge to the distribution $\guesspi\bowtie\pi$, certifying continuity as desired.

Therefore, to prove continuity, it suffices to prove that, for all $h,s,a$, the function $\lambda\pi.\guesspi(h,s,a)$ is continuous. We will prove this fact by downwards induction, so we may select an arbitrary $h,s,a$, and freely assume that $\lambda\pi.\guesspi(h',s',a')$ is continuous for all $s',a'$ and $h'>h$.

First, we verify continuity of the function $\lambda\pi,\mu.(\mu,M^{>h}_{t,\pi})\bowtie\pi$. Let $\pi_{n},\mu_{n}$ limit to $\pi,\mu$. The MDP $(\mu_{n},M^{>h}_{t,\pi_{n}})$ converges, in its starting distribution, and all of its transition probabilities, to $(\mu,M^{>h}_{t,\pi})$, by the induction assumption that $\guesspi(h',s',a')$ was continuous for all $s',a'$ and $h'>h$, and convergence of $\mu_{n}$. Then, by our earlier argument about there being finitely many trajectories, and convergence in the probability of each trajectory, the distributions $(\mu_{n},M^{>h}_{t,\pi_{n}})\bowtie\pi_{n}$ converge to $(\mu,M^{>h}_{t,\pi})\bowtie\pi$. Our convergent sequence was arbitrary, so continuity is established.

This function that we just established the continuity of, is the same as the function
\\
$\lambda\pi,\mu.\widetilde{M}^{>h}_{t,\pi}(\mu,\pi)$. Taking a distribution over trajectories, and asking for the probability of some $h',s'$ event, or the expected sum of future rewards, is a continuous function from probability distributions over trajectories to $\mathbb{R}$. Compositions of continuous functions are continuous, so, the functions
\\
$\lambda\pi,\mu.\expec_{\widetilde{M}^{>h}_{t,\pi}(\mu,\pi)}\left[\sum_{k=h}^{H}r_{k}\right]$ and $\lambda\pi,\mu.\mathbb{P}_{\widetilde{M}^{>h}_{t,\pi}(\mu,\pi)}(h',s')$ are continuous. 

Further, the quantities $X^{B}_{t,\pi}$, are continuous in $\pi$, because they only depend on $\pi$ through the probability of a particular action in a particular situation, and this probability converges as the policy does.

Now, for the various bettors $B$, we look at the continuity of the function $\lambda\pi,\mu.g^{B,h,s,a}_{t,\pi}(\mu)$. For fragment and uniform bettors, their $g$ functions are either constants, or constants in $\pi$ and continuous in $\mu$. For calibration bettors, their $g$ functions are either constants, or they depend on multiplying two continuous functions of $\pi,\mu$ together (the $X^{B}_{t,\pi}$ quantity, and the probability of some $h',s'$), so they are continuous. For the pessimism bettor, it is just a constant times a continuous function of $\pi,\mu$ (the expectation of future reward). Therefore, for all bettors $B$, the function $\lambda\pi,\mu.g^{B,h,s,a}_{t,\pi}(\mu)$ is continuous.

In the special case of $h=H$, the fragment and uniform $g$ functions only depend continuously on $\mu$, not $\pi$. The calibration $g$ functions are all constants, and the pessimism bettor only depends on $\mu$, so continuity of all of the $g$ functions in $\pi,\mu$ holds automatically (because it doesn't depend on $\pi$ and depends continuously on $\mu$), which provides a base case for the downwards induction that doesn't require the induction step to prove, rendering everything well-founded.

Because all of the functions $\lambda\pi,\mu.g^{B,h,s,a}_{t,\pi}(\mu)$ of type $\Pi_{RNS}\times\Delta(\{0,1\}\times\states)\to\mathbb{R}$ are continuous, their mixture, $\lambda\pi,\mu.\bel{\expec}{B\sim\zeta_{t}}[g^{B,h,s,a}_{t,\pi}(\mu)]$ is as well. $\Pi_{RNS}\times\Delta(\{0,1\}\times\states)$ is a compact space, so, by the Heine-Cantor theorem, the mixture function is uniformly continuous. Therefore, if $\pi_{n}$ converges to $\pi$, $\bel{\max}{\mu}|\bel{\expec}{B\sim\zeta_{t}}[g^{B,h,s,a}_{t,\pi_{n}}(\mu)]-\bel{\expec}{B\sim\zeta_{t}}[g^{B,h,s,a}_{t,\pi}(\mu)]|$ converges to 0 (because for all $\mu$, the low distance between $\pi_{n}$ and $\pi$ translates into a guarantee on the difference of the values, by uniform continuity), so the functions $\bel{\expec}{B\sim\zeta_{t}}[g^{B,h,s,a}_{t,\pi_{n}}]$ uniformly converge to $\bel{\expec}{B\sim\zeta_{t}}[g^{B,h,s,a}_{t,\pi}]$.

Now, we show that $M_{t,\pi_{n}}(h,s,a)$ limits to $\guesspi(h,s,a)$, as follows. Because $\Delta(\{0,1\}\times\states)$ is a compact space, the sequence $M_{t,\pi_{n}}(h,s,a)$ of distributions has limit points, call them $M$. We will show that all limit points must equal $\guesspi(h,s,a)$, so the sequence converges to the desired point. To do this, we pass to a convergent subsequence, and compute
$$\bel{\expec}{B\sim\zeta_{t}}[g^{B,h,s,a}_{t,\pi}(M)]=\lim_{m\to\infty}\bel{\expec}{B\sim\zeta_{t}}[g^{B,h,s,a}_{t,\pi_{n_{m}}}(M_{t,\pi_{n_{m}}}(h,s,a))]$$
$$=\lim_{m\to\infty}\min(\bel{\expec}{B\sim\zeta_{t}}[g^{B,h,s,a}_{t,\pi_{n_{m}}}])=\min(\bel{\expec}{B\sim\zeta_{t}}[g^{B,h,s,a}_{t,\pi}])$$

In order, this was by continuity of the mix of $g$ functions in $\pi$ and the distribution, because $M$ was a limit of a sequence of distributions. The second equality was because $M_{t,\pi_{n_{m}}}(h,s,a)$ is the minimize of the associated function. Finally, we use our result that the functions $\bel{\expec}{B\sim\zeta_{t}}[g^{B,h,s,a}_{t,\pi_{n_{m}}}]$ uniformly converge to $\bel{\expec}{B\sim\zeta_{t}}[g^{B,h,s,a}_{t,\pi}]$, so their minimum values converge as well. This certifies that any limit point of the sequence $M_{t,\pi_{n}}(h,s,a)$ minimizes $\bel{\expec}{B\sim\zeta_{t}}[g^{B,h,s,a}_{t,\pi}]$. However, by Lemma \ref{lem22}, the minimizer is unique. Therefore, the sequence $M_{t,\pi_{n}}(h,s,a)$ converges to $M_{t,\pi}(h,s,a)$. The convergent sequence $\pi_{n}$ was arbitrary, as was $s,a$, so we have established that all the functions $\lambda\pi.\guesspi(h,s,a)$ are continuous, if continuity holds for all $h'>h$. Therefore, our downwards induction proof of continuity in $\pi$ for the transition probabilities goes through (the base case at $H$ does not rely on the induction assumption), which implies continuity of the overall estimator. $\blacksquare$

\begin{coro}
There is an algorithm which attains $\Tilde{\mathcal{O}}(\sqrt{H^{2}S^{3}AT})$ regret on all 1-bounded RMDP's in the episodic RMDP setting.
\end{coro}

Take the failure probability, $\delta$, to be $T^{-1/2}$. By Theorem \ref{the5} and Proposition \ref{pro7}, there is a continuous estimator for $\hypo_{\text{parhalf}}$ with $\beta_{\estim}(T,\delta)\in\widetilde{O}(HS^{2})$, and $\alpha_{\estim}(T,\delta)\in\widetilde{O}(\sqrt{T})$. By Theorem \ref{the4}, because the estimates are continuous, the modified DEC for the 1-bounded subset of $\hypo_{\text{parhalf}}$ is on the order of $\mathcal{O}(\sqrt{HSA}\eps)$. By Theorem \ref{the1} applied to the expected sum of Hellinger losses, along with $\eps$ being $\sqrt{\frac{\beta_{\estim}(T,\delta)}{T}}$, the Theorem \ref{the5} estimator may be combined with the E2D algorithm to attain $\Tilde{\mathcal{O}}(\sqrt{H^{2}S^{3}AT})$ regret on the 1-bounded subset of $\hypo_{\text{parhalf}}$.
\\

To transfer this result to all 1-bounded RMDP's, we must show that, given any 1-bounded RMDP $\imdp$, there is a surrogate RMDP $\imdp'\in\hypo_{\text{parhalf}}$ where, if $\imdp$ is a 1-bounded true model of the environment, then the surrogate is also a 1-bounded true model with the same worst-case expected reward for optimal play. Then, if $\imdp$ was a true hypothesis (a hypothesis that the environment respects the constraints of), the same would apply to $\imdp'$. The E2D algorithm would guarantee low regret against $\imdp'$ because it is true and in $\hypo_{\text{parhalf}}$ and 1-bounded. Finally, because $\imdp,\imdp'$ ensure the same bound on expected reward, the low regret against $\imdp'$ implies low regret against $\imdp$. If this argument applies to all $\imdp$, the corollary follows.
\\

To prove that every 1-bounded RMDP $\imdp$ has a surrogate with the desired properties, we introduce the following definition. Given an RMDP $\imdp$ and a policy $\pi\in\Pi_{RNS}$, define the value of $h,s$ as follows. For $h=H+1$, it is zero. For other $h,s$, it is defined as
$$V^{\imdp,\pi}_{h,s}=\bel{\expec}{a\sim\pi(h,s)}\left[\bel{\min}{\mu\in\imdp(h,s,a)}\bel{\expec}{r,s'\sim\mu}\left[r+V^{\imdp,\pi}_{h+1,s'}\right]\right]$$
This is the expected sum of future rewards at $h,s$ if the environment will make the worst choices which are compatible with the RMDP $\imdp$.

For any RMDP, $f^{\imdp}(\pi)$ (the worst-case expected sum of rewards) equals $V^{\imdp,\pi}_{0,s_{0}}$. 1-boundedness is equivalent to $\forall h,s,\pi:V^{\imdp,\pi}_{h,s}\leq 1$. Further, given an RMDP, its optimal policy may be chosen to be deterministic, because, for any $h,s$, $V^{\imdp,\pi}_{h,s}$ may be maximized by a deterministic choice of action.

Now, given an $\imdp$, define the surrogate $\imdp'$ as follows. Let $\pi_{\imdp}$ be a deterministic optimal policy for that RMDP. Then, $\imdp'$ is the RMDP generated by the PRMDPR which maps $h,s$ to $\pi_{\imdp}(h,s),\Psi^{[0,1]}_{\lambda s'.V^{\imdp,\pi_{\imdp}}_{h+1,s'},V^{\imdp,\pi_{\imdp}}_{h,s}}$. This is in $\hypo_{\text{parhalf}}$ by construction.
\\

First, $\pi_{\imdp}$ is an optimal policy for the surrogate $\imdp'$, and any other policy attains the same score or worse in all circumstances. This holds because, when a PRMDPR is converted to an RMDP, the transition kernel for non-recommended actions is completely unconstrained, and it's possible that, for all actions which aren't recommended, the environment returns zero reward, and a transition to the worst state. Therefore, in any situation $h,s$, the recommended action produces the maximum guarantee on expected future reward. Because the optimal action in every situation coincides with the action of $\pi_{\imdp}$, this policy is optimal for the surrogate $\imdp'$.
\\

Second, we will show that, for any $h,s$, we have $V^{\imdp,\pi_{\imdp}}_{h,s}=V^{\imdp',\pi_{\imdp}}_{h,s}$. This is proved by downwards induction. It holds at the base case of $h=H+1$ because the values are zero. For the induction step, we compute
$$V^{\imdp',\pi_{\imdp}}_{h,s}=\bel{\min}{\mu\in\imdp'(h,s,\pi_{\imdp}(h,s))}\bel{\expec}{r,s'\sim\mu}\left[r+V^{\imdp,\pi}_{h+1,s'}\right]=V^{\imdp,\pi_{\imdp}}_{h,s}$$

The last equality was because $\imdp'(h,s,\pi_{\imdp})=\Psi^{[0,1]}_{\lambda s'.V^{\imdp,\pi_{\imdp}}_{h+1,s'},V^{\imdp,\pi_{\imdp}}_{h,s}}$, and this $\Psi$ set is 
\\
$\Set{\mu}{\bel{\expec}{r,s'\sim\mu}[r+V^{\imdp,\pi_{\imdp}}_{h+1,s'}]\geq V^{\imdp,\pi_{\imdp}}_{h,s}}$. By induction, we then have that $V^{\imdp,\pi_{\imdp}}_{h,s}=V^{\imdp',\pi_{\imdp}}_{h,s}$ for all $h,s$. Because $\imdp$ was 1-bounded (all the values are $\leq 1$), $\imdp'$ is as well. Because $\pi_{\imdp}$ is optimal for $\imdp'$, we also have 
$$\max(f^{\imdp})=f^{\imdp}(\pi_{\imdp})=V^{\imdp,\pi_{\imdp}}_{0,s_{0}}=V^{\imdp',\pi_{\imdp}}_{0,s_{0}}=f^{\imdp'}(\pi_{\imdp})=\max(f^{\imdp'})$$

All that remains to prove the corollary is to show that, if $\imdp$ is a true model of the environment (at every $h,s,a$, the distribution over the next reward and state is in $\imdp(h,s,a)$), the surrogate $\imdp'$ is as well. This may be proven by showing that, for all $h,s,a$ we have $\imdp(h,s,a)\subseteq\imdp'(h,s,a)$. For an $h,s,a$ where $a$ is not the recommended action, $\imdp'(h,s,a)$ is the entire space of distributions, so this case is trivial. If the action is the recommended action $\pi_{\imdp}(h,s)$, we have, for any $\mu\in\imdp(h,s,\pi_{\imdp}(h,s))$,
$$\bel{\expec}{r,s'\sim\mu}[r+V^{\imdp,\pi_{\imdp}}_{h+1,s'}]\geq\bel{\min}{\mu'\in\imdp(h,s,\pi_{\imdp}(h,s))}\bel{\expec}{r,s'\sim\mu'}[r+V^{\imdp,\pi_{\imdp}}_{h+1,s'}]=V^{\imdp,\pi_{\imdp}}_{h,s}$$

This inequality certifies $\mu\in\Psi^{[0,1]}_{\lambda s'.V^{\imdp,\pi_{\imdp}}_{h+1,s'},V^{\imdp,\pi_{\imdp}}_{h,s}}=\imdp'(h,s,\pi_{\imdp}(h,s))$. $\mu$ was arbitrary, so 
\\
$\imdp(h,s,\pi_{\imdp}(h,s))\subseteq\imdp'(h,s,\pi_{\imdp}(h,s))$. This concludes the proof. $\blacksquare$
\end{document}